%% file: neurips_2026.tex
\documentclass{article}

\PassOptionsToPackage{numbers,compress}{natbib}
\usepackage[preprint]{neurips_2026}

\usepackage[utf8]{inputenc} 
\usepackage[T1]{fontenc}    
\usepackage[breaklinks=true]{hyperref}       
\usepackage{url}            
\usepackage{booktabs}       
\usepackage{amsfonts}       
\usepackage{dsfont}         
\usepackage{nicefrac}       
\usepackage{microtype}      
\usepackage{xcolor}         
\usepackage{soul}
\usepackage[small]{caption}
\usepackage{graphicx}
\usepackage{amsmath}
\usepackage{amsthm}
\usepackage{amssymb}
\usepackage{algorithm}
\usepackage{algorithmic}
\usepackage{float}
\usepackage{xspace}
\usepackage{multirow}
\usepackage{tikz}
\usepackage{pgfplots}
\pgfplotsset{compat=1.18}
\usepgfplotslibrary{groupplots}
\usepgfplotslibrary{fillbetween}
\usetikzlibrary{arrows.meta, shapes, shapes.geometric, positioning, shadows, calc, fit, backgrounds}
\usepackage[most]{tcolorbox}
\tcbuselibrary{breakable, skins, listings}
\usepackage[framemethod=TikZ]{mdframed}
\usepackage{enumitem}
\usepackage{nameref}
\usepackage[percent]{overpic}
\usepackage{pifont}
\usepackage{nccmath} 
\usepackage{colortbl}  
\usepackage{subcaption}
\usepackage{listings}
\usepackage{wrapfig}
\usepackage{longtable}

\newcommand{\breakablett}[1]{\allowbreak\text{\texttt{#1}}}
\newcommand{\breakablesc}[1]{\allowbreak\text{\textsc{#1}}}

\newcommand{\beagle}{\textbf{\textsc{Beagle}}\xspace}

\newcommand{\assist}{\breakablesc{As\-sis\-tance}\xspace}
\newcommand{\offtopic}{\breakablesc{Off-Top\-ic}\xspace}
\newcommand{\planning}{\breakablesc{Plan\-ning}\xspace}
\newcommand{\monitoring}{\breakablesc{Mon\-i\-tor\-ing}\xspace}
\newcommand{\reflecting}{\breakablesc{Re\-flect\-ing}\xspace}
\newcommand{\enacting}{\breakablesc{En\-act\-ing}\xspace}

\newcommand{\constructing}{\breakablett{Con\-struct\-ing}\xspace}
\newcommand{\debugging}{\breakablett{De\-bug\-ging}\xspace}
\newcommand{\assessing}{\breakablett{As\-sess\-ing}\xspace}

\newcommand{\phigh}{\textsc{High}}
\newcommand{\plow}{\textsc{Low}}

\definecolor{llmjudge}{RGB}{128, 128, 128}
\definecolor{cLow}{HTML}{D94801}
\definecolor{cHigh}{HTML}{2171B5}
\definecolor{cReal}{HTML}{2CA02C}
\newcommand{\llmcol}[1]{\textcolor{llmjudge}{#1}}

\newcommand{\eqsmall}{%
}

\title{BEAGLE: Behavior-Enforced Agent for Grounded Learner Emulation}

\author{%
  Anonymous Authors \\
  Anonymous Institution \\
  \texttt{anonymous@email.com} \\
}

\author{%
  \textbf{Hanchen David Wang}, \textbf{Clayton Cohn}, \textbf{Zifan Xu}, \\
  \textbf{Siyuan Guo}, \textbf{Gautam Biswas}, \textbf{Meiyi Ma} \\
  Vanderbilt University \\
  \texttt{\{hanchen.wang.1,clayton.a.cohn,zifan.xu,} \\
  \texttt{siyuan.guo,gautam.biswas,meiyi.ma\}@vanderbilt.edu} \\
}

\begin{document}

\maketitle

\begin{abstract}
Simulating student learning behaviors in open-ended problem-solving environments holds potential for education research, from training adaptive tutoring systems to stress-testing pedagogical interventions. However, collecting authentic data is challenging due to privacy concerns and the high cost of longitudinal studies. While Large Language Models (LLMs) offer a promising path to student simulation, they suffer from \textit{competency bias}, optimizing for efficient correctness rather than the erratic, iterative struggle characteristic of novice learners. We present \beagle, a neuro-symbolic framework that addresses this bias by incorporating Self-Regulated Learning (SRL) theory into a novel architecture. \beagle integrates three key technical innovations: (1) a semi-Markov model that governs the timing and transitions of cognitive behaviors and metacognitive behaviors; (2) Bayesian Knowledge Tracing with explicit flaw injection to enforce realistic knowledge gaps and ``unknown unknowns''; and (3) a decoupled agent design that separates high-level strategy use from code generation actions to prevent the model from silently correcting its own intentional errors. In evaluations on Python programming tasks, \beagle significantly outperforms state-of-the-art baselines in reproducing authentic trajectories. In a human Turing test, participants could not reliably tell \beagle traces apart from real student data: classification accuracy was statistically equivalent to chance (52.8\%, $d'=0.15$, $N=71$).
\end{abstract}

\section{Introduction}\label{sec:introduction}

Generative AI is advancing education, especially through personalized learning and real-time feedback. However, these advances share significant challenges: they require authentic student data to train, evaluate, and refine pedagogical models. Acquiring such data is difficult because strict privacy regulations (e.g., FERPA, GDPR) limit access to collecting comprehensive classroom data~\cite{vie2022privacy}, while the prohibitive costs of data collection, such as IRB approvals, parental consent, and longitudinal designs, create substantial barriers~\cite{blom2009consent}. Moreover, intervening in how real students learn raises ethical concerns, particularly in K-12 settings where suboptimal and incorrect experimental interventions could disadvantage minors academically~\cite{holmes2022ethics}. This problem has motivated researchers to explore synthetic student generation~\cite{kaser2024simulated,xu2023leveraging,lu2024generative,leinonen2025llm}, and simulating learner behaviors computationally enables researchers to stress-test pedagogical strategies without putting real students at risk.

Current Large Language Models (LLMs) offer an appealing foundation: their capacity for nuanced generation and persona adoption makes them natural candidates for simulating rich learner behaviors. However, LLMs exhibit \textbf{\textit{competency bias}}: the inherent tendency of preference-tuned models to produce correct solutions even when prompted to simulate novice students~\cite{li2025can,mannekote2025can,aher2023using}. Competency bias persists because authentic student behavior emerges from processes rather than outcomes. 

\begin{wrapfigure}{r}{0.43\linewidth}
    \centering
    \includegraphics[width=\linewidth]{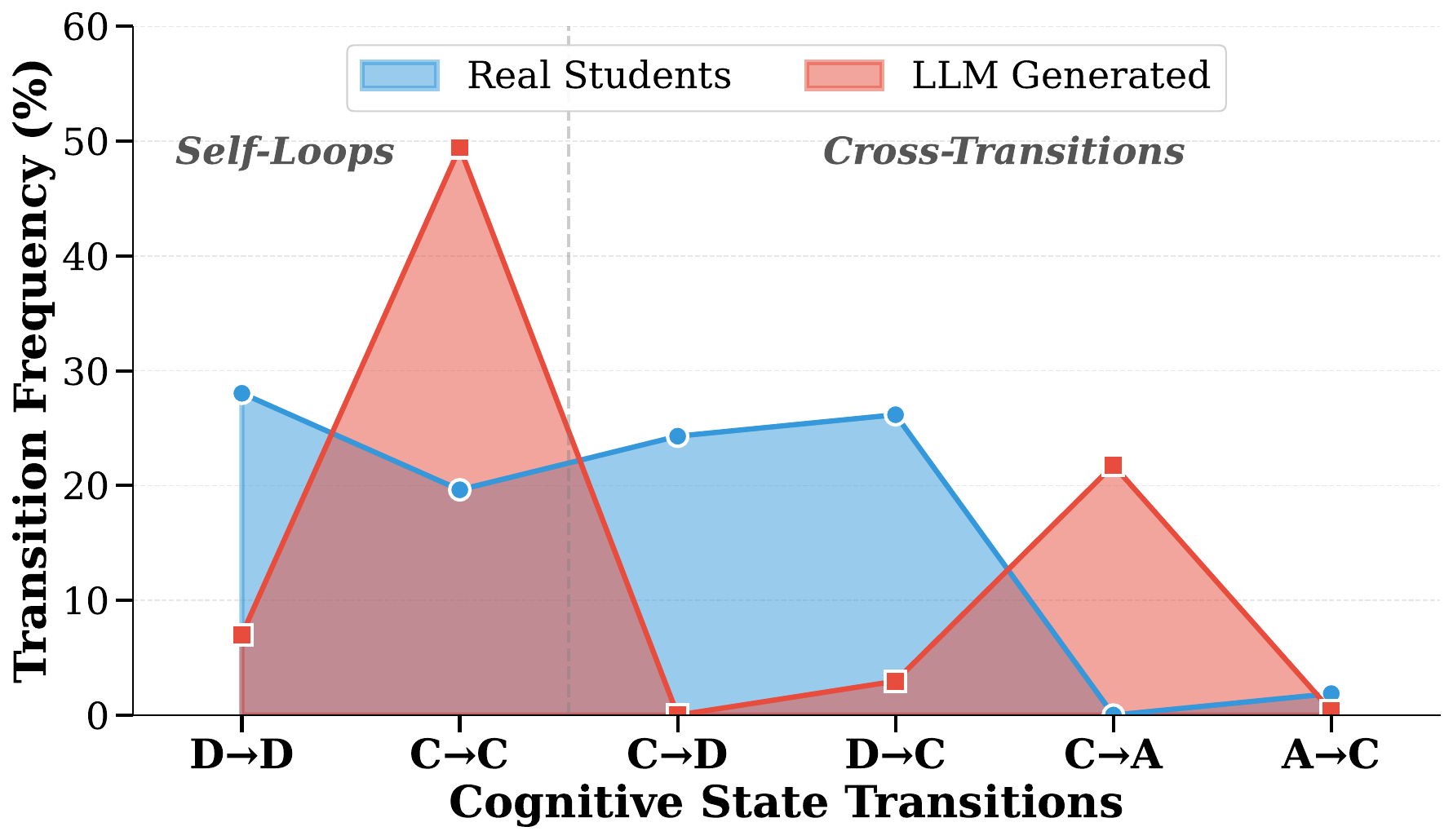}
    \caption{\textbf{\textit{Competency bias}}: real students debug persistently (D$\rightarrow$D: 28\%) while LLMs construct linearly. C (\constructing), D (\debugging), A (\assessing).}
    \label{fig:motivating}
\end{wrapfigure}

\textbf{\textit{Self-Regulated Learning (SRL)}} theory~\cite{zimmerman1990self} provides a principled framework for this view, positing that learning unfolds as a cyclic process involving cognition (attempting tasks), metacognition (monitoring progress), and motivation and affect (engagement and seeking help, or disengaging). Authentic student errors are artifacts of breakdowns in these regulatory processes: students frequently iterate between planning, enacting, monitoring, and reflecting. \textit{Prior attempts to induce novice-like outputs through persona prompting or knowledge suppression fail because preference-tuned LLMs are fundamentally optimized for outcome quality, not process fidelity.} Empirical evidence supports this claim. As shown in Fig.~\ref{fig:motivating}, student patterns observed in \cite{snyder2024analyzing} reveal a heavy loop on iterative debugging. In contrast, LLM-generated students show the opposite behavior: dominating continuous construction transitions while bypassing the back-and-forth debugging process almost entirely.

These limitations motivate our framing of the \textit{multi-dimensional fidelity problem} across three dimensions:  \textbf{\textit{(1) Behavioral Infidelity:}} LLMs optimized for efficiency exhibit a \textit{linear construction trap}, avoiding the iterative trial-and-error cycles central to SRL \cite{puech2025towards,huang2024large}. \textbf{\textit{(2) Epistemic Infidelity:}} Authentic novices suffer the \textit{curse of incompetence}~\cite{ehrlinger2008unskilled}: they lack the skill to solve problems (unknown unknowns) and the insight to recognize errors. LLMs, acting as ``masked experts,'' frequently violate these constraints by diagnosing errors before execution or leaking suppressed knowledge~\cite{shao2023character,taubenfeld2024systematic}.  \textit{\textbf{(3) Perceptual Infidelity:}} RLHF training reduces output diversity to a flattened caricature~\cite{kirk2024understanding}, failing to capture the noise of authentic student expression~\cite{cheng2023compost,hu2024quantifying}.

We present \beagle (\textbf{B}ehavior-\textbf{E}nforced \textbf{A}gent for \textbf{G}rounded \textbf{L}earner \textbf{E}mulation), a neuro-symbolic framework that overcomes competency bias to generate high-fidelity synthetic student trajectories. \beagle enables scalable AI education research to train adaptive tutors, validate pedagogical theories, and stress-test interventions without the ethical constraints of real student experimentation. Our technical contributions include:

\begin{enumerate}[nosep, leftmargin=*]
    \item A semi-Markov model trained on real student logs that governs when and how long agents plan, act, or reflect, enabling the iterative C$\leftrightarrow$D transitions that LLM baselines systematically bypass.

    \item Bayesian Knowledge Tracing that constrains what concepts the agent can utilize, with explicit flaw injection to simulate unknown unknowns and observation filtering to enforce the curse of incompetence, preventing premature diagnosis of errors.

    \item A Strategist/Executor architecture that decouples planning from code generation, using task framing to prevent silent self-correction.

    \item A comprehensive evaluation across four tasks, six frontier LLM backbones, and nine baselines, including a human Turing test ($N=71$) in which \beagle traces reach near-chance discriminability ($d'=0.15$, $p_{\text{TOST}}=0.038$).
\end{enumerate}

\section{Problem Formulation}

We formulate synthetic student generation as a sequence synthesis problem defined over two semantic domains: the universe of natural language strings $\mathcal{S}_{\text{text}}$ and the universe of executable source codes $\mathcal{S}_{\text{code}}$. The process is conditioned on a static configuration $\Phi = (\mathcal{P}, \mathcal{K}, \rho)$, where $\mathcal{P} \in \mathcal{S}_{\text{text}}$ represents the problem description and $\mathcal{K}$ denotes the set of tracked knowledge components. The student profile $\rho = (\rho_{\text{behav}}, \rho_{\text{persona}})$ consists of two dimensions: a behavioral parameter $\rho_{\text{behav}} \in \{\phigh{}, \plow{}\}$ governing transition dynamics, and a linguistic persona $\rho_{\text{persona}} \in \{\phigh{}, \plow{}\}$, which maps a proficiency level to specific stylistic constraints.

We model interaction dynamics with action space $\mathcal{A} = \mathcal{S}_{\text{code}} \times \mathcal{S}_{\text{text}}$ and a \textit{hierarchical} latent state: \textit{metacognitive behaviors} $\mathcal{M}$ (e.g., \planning) govern high-level phases, while \textit{cognitive behaviors} $\mathcal{C}$ (e.g., \constructing) are fine-grained actions \textit{within} each phase. Trajectories split into metacognitive segments indexed by $n$ (segment $n$ has behavior $M_n \in \mathcal{M}$ and duration $D_n$); segments are concatenated into cognitive steps $t \in \{1,\ldots,T\}$ with $T = \sum_n D_n$ and $n(t)$ mapping step $t$ to its segment. External interventions $\mathbf{I} = \{\mathcal{I}_t\}_{t=1}^T$ ($\mathcal{I}_t \subseteq \mathcal{S}_{\text{text}}$) carry optional tutor feedback. A synthetic trajectory is $\tau = \{(M_{n(t)}, C_t, a_t, o_t)\}_{t=1}^{T}$ with $a_t \in \mathcal{A}$ and observation $o_t$ (agent sees filtered $\tilde{o}_t$; \S\ref{sec:environment}). We write the generative mapping as $\tau \sim P_\theta(\cdot \mid \Phi, \mathbf{I})$, where $\theta$ collects the architectural mechanisms (semi-Markov dynamics, knowledge constraints, agent design).

\noindent\textbf{The Multi-Dimensional Fidelity Problem.} We formalize three orthogonal fidelity objectives. Let $P_{\text{real}}(C)$ denote the ground-truth marginal distribution over cognitive behaviors and $P_{\theta}(C)$ the simulated distribution; transition-level dynamics (e.g., debugging stickiness) are captured separately by the compound $D_{\text{debug}}$ metric (App.~\ref{app:evaluation-metrics}).

\begin{enumerate}[nosep, leftmargin=*]
    \item \textit{\textbf{Behavioral Fidelity:}} We minimize the KL divergence between the simulated and real cognitive behavior distributions:
    \[D_{\text{KL}}(P_{\theta}(C) \;||\; P_{\text{real}}(C)) \]

    \item \textit{\textbf{Epistemic Fidelity:}} We enforce the \textit{curse of incompetence} via the error recurrence rate. Real novices repeatedly encounter the same error type due to limited diagnostic expertise, while LLMs fix errors immediately:
    \[\mathbb{P}(\exists t' > t : e \in o_{t'} \mid e \in o_t) \]

    \item \textit{\textbf{Perceptual Fidelity:}} We maximize the expected realism score assigned by an evaluator function $\mathcal{J}(\tau)$ (e.g., human or LLM judge):
    \[\mathbb{E}_{\tau \sim P_{\theta}}[\mathcal{J}(\tau)] \]
\end{enumerate}

\section{Method}\label{sec:method}

\subsection{System Overview}

\beagle is a training-free framework: Neural Action does zero-shot in-context generation with frozen LLMs, emitting free-form Python rather than choosing from a fixed action vocabulary. Symbolic Control parameters (semi-Markov transitions, Gamma durations, interrupt rates) are fit once by maximum likelihood on historical student logs~\cite{snyder2024analyzing} (App.~\ref{sec:gt_data_analysis},~\ref{app:interrupt}); BKT slip/guess/learning rates use standard priors~\cite{corbett1994knowledge} (App.~\ref{app:bkt_details}). No parameter is tuned on evaluation tasks. Three coupled subsystems implement this design: \textbf{Symbolic Control} enforces SRL workflow and knowledge constraints (\S\ref{sec:symbolic_control}); \textbf{Neural Action} factors generation into a Strategist/Executor pipeline whose task framing prevents silent self-correction (\S\ref{sec:neural_action}); and the \textbf{Environment} executes code and gates feedback by metacognitive behavior to enforce novice perceptual blindness (\S\ref{sec:environment}).

\begin{figure}[t]
    \centering
    \begin{overpic}[width=1\linewidth]{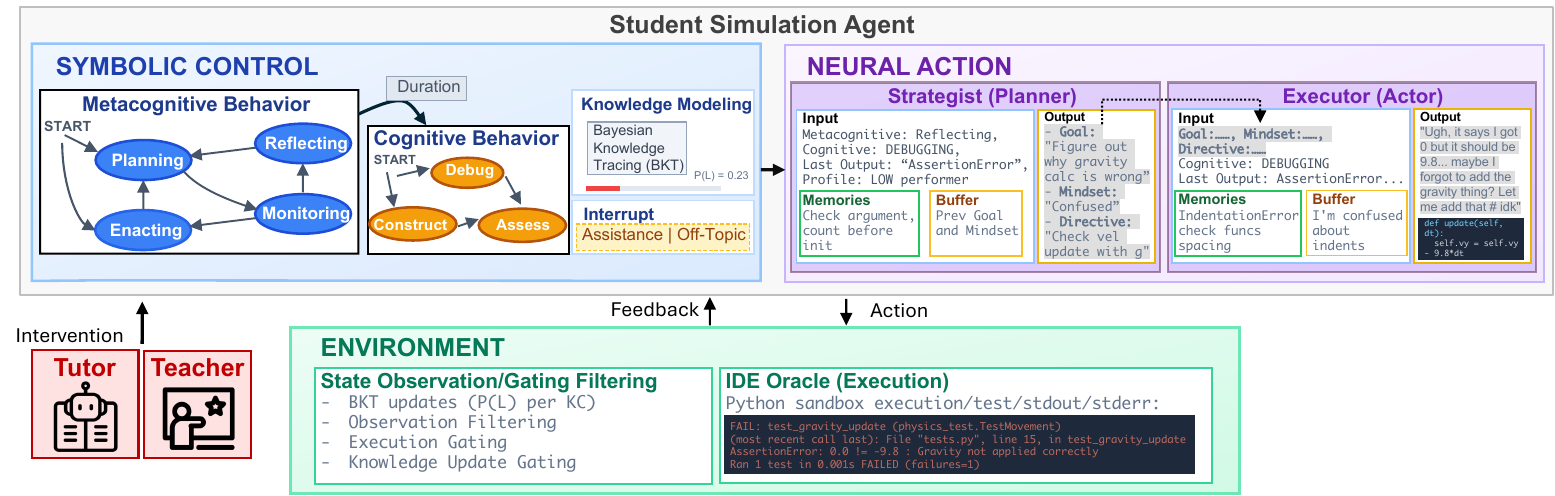}
    \end{overpic}
    \caption{Overview of the \beagle agent architecture. \textbf{Symbolic Control} (left) governs high-level behavior via semi-Markov metacognitive and cognitive behavior state machines, with BKT-based knowledge tracking. \textbf{Neural Action} (right) factors generation into a two-stage pipeline: a \emph{Strategist} that emits a goal, mindset, and directive, and an \emph{Executor} that produces code and monologue conditioned on that directive (prompts in App.~\ref{app:prompts}). \textbf{Environment} (bottom) executes code via the IDE Oracle and feeds gated observations back to both subsystems.}
    \label{fig:architecture}
\end{figure}

\noindent\textbf{Step Lifecycle.}
At each cognitive step $t$, Symbolic Control samples the current metacognitive behavior $M_n$ (semi-Markov) and cognitive behavior $C_t$, together with discretized knowledge constraints $\kappa_t$ (BKT + EFI). The Strategist consumes these and the Shared Context $\Omega_t = (c_{t-1}, \tilde{o}_{t-1}, \kappa_t, \mathcal{I}_t)$ to emit a plan $(g_n, m_n, d_n)$ (Goal, Mindset, Directive); the Executor expands the plan into action $a_t = (c_t, u_t)$ (code and monologue). The Environment runs $c_t$ in the IDE Oracle, gates the observation by $M_n$ to yield $\tilde{o}_t$, and updates BKT mastery only when $M_n \in \{\reflecting, \monitoring\}$. The remaining subsections specify each block; Fig.~\ref{fig:architecture} traces the same flow visually.

\subsection{Symbolic Control}
\label{sec:symbolic_control}

\noindent\textbf{Semi-Markov Dynamics.}\label{para:semi-markov} To enforce realistic temporal consistency, we employ a semi-Markov Controller. We adopt this architecture because standard Markov models inherently enforce geometric duration distributions that fail to capture the irregular persistence (e.g., ``getting stuck'') observed in real student logs (see App.~\ref{sec:gt_data_analysis} for empirical validation).

Each metacognitive segment $n$ is characterized by a behavior $M_n$, a sampled
duration $D_n$, and a sequence of cognitive behaviors
$\mathbf{C}_n = (C_n^{(1)}, \ldots, C_n^{(D_n)})$. Concatenating all segments
yields the global trajectory $(C_1, \ldots, C_T)$ with $T = \sum_n D_n$. We
factorize the joint distribution as:
\begin{align}
    P(M_n, D_n, \mathbf{C}_n) &= P(M_n \mid M_{n-1})
    \cdot P(D_n \mid M_n) \nonumber \cdot \prod_{i=1}^{D_n} P(C_n^{(i)} \mid M_n, C_n^{(i-1)})
    \label{eq:factorization}
\end{align}
The metacognitive layer operates as a first-order Markov chain, while durations $D_n$ follow Gamma distributions tailored to $\rho_{\text{behav}}$. Two stochastic interrupt behaviors, \assist{} and \offtopic{}, extend this chain as progress-conditioned Gaussians: \assist{} peaks mid-task (proactive help-seeking) and \offtopic{} peaks late (frustration), with parameters fit on empirical patterns (App.~\ref{app:interrupt}). Architecturally, \beagle's components are domain-agnostic, requiring only a knowledge component (KC) taxonomy and action schema to port; the semi-Markov layer can also host alternative theoretical frameworks by replacing $M$ and $C$. 

\noindent\textbf{Knowledge Modeling (BKT).}\label{para:bkt}
Furthermore, we employ a hybrid mechanism combining Bayesian Knowledge Tracing (BKT) with Explicit Flaw Injection (EFI; \S\ref{para:efi}). For each knowledge component $k \in \mathcal{K}$ at step $t$, we model the probability of mastery as a continuous variable $P(L_k^{(t)}) \in [0, 1]$. This probability is updated via Bayes' rule after every synthetic observation generated during \reflecting\ or \monitoring\ behaviors (Details in App.~\ref{app:bkt_details}).

\noindent\textbf{Explicit Flaw Injection (EFI).}\label{para:efi}
Standard BKT attributes errors solely to performance slips, failing to model the complete absence of conceptual awareness. To capture these ``unknown unknowns,'' EFI constrains selected KCs as \emph{Blocked}: they deterministically yield incorrect observations, freeze BKT updates, and are treated as nonexistent in the agent's vocabulary so the Executor cannot fall back on suppressed expert knowledge. A Blocked KC cannot be recovered through self-practice alone; the agent must receive a tutor hint targeting that KC (App.~\ref{app:tutor_details}). Mastered and Partial KCs surface as natural-language constraints $\kappa_t$, with Partial sampled stochastically Correct/Incorrect via BKT slip/guess.

\subsection{Neural Action}
\label{sec:neural_action}

We decompose the agent function into a two-stage pipeline: a \textit{Strategist} that synthesizes intent and an \textit{Executor} that realizes it. To streamline notation, let $\Omega_t = (c_{t-1}, \tilde{o}_{t-1}, \kappa_t, \mathcal{I}_t)$ denote the \textit{Shared Context}. Additionally, each stage maintains an independent memory buffer, $\Gamma_{\text{Strat}, n}$ and $\Gamma_{\text{Exec}, t}$, preventing circular dependencies between planning and execution logic.

\noindent\textbf{The Strategist.}
The Strategist bridges the gap between symbolic constraints and natural language. It maps the current metacognitive segment to an intermediate representation $(g_n, m_n, d_n)$: a durable \textit{Goal}, an affective \textit{Mindset}, and an actionable \textit{Directive}.
{\eqsmall\begin{align}
    f_{\text{Strat}} : (M_n, C_t, \Gamma_{\text{Strat}, n}, \Omega_t; \rho_{\text{persona}}) \mapsto (g_n, m_n, d_n)
\end{align}}

This architecture decouples \textit{planning} from \textit{writing}. The metacognitive behavior $M_n$ bounds the strategic depth (e.g., \enacting\ forces impulsive directives), while the persona $\rho_{\text{persona}}$ conditions the formulation. By operating solely on high-level goals, the Strategist serves as an internal monologue, separate from the details of coding.

\noindent\textbf{The Executor.}
The Executor consumes this high-level directive to generate the final action $a_t = (c_t, u_t)$ at each cognitive step. Crucially, it is constrained by the distilled directive: the symbolic state shapes vocabulary and output format (via Mandate, Profile, and Task$_{\text{Exec}}$ blocks; App.~\ref{app:prompts}), but the executable plan comes only from $(g_n, m_n, d_n)$:
{\eqsmall\begin{align}
    f_{\text{Exec}} : (M_n, C_t, \Gamma_{\text{Exec}, t}, \Omega_t, g_n, m_n, d_n; \rho_{\text{persona}}) \mapsto a_t
\end{align}}

This task framing is what enforces epistemic fidelity: the Executor implements a directive rather than diagnosing the problem, so even with full access to code and error context it follows flawed directives instead of overriding them with better fixes. Merging the two stages collapses this constraint and enables silent self-correction (validated in ablation, \S\ref{sec:ablation}). Both stages use a block-based prompt architecture (App.~\ref{app:prompts}); the Strategist's $(g_n, m_n, d_n)$ is injected into the Executor's user prompt as a hard constraint, keeping planning and writing aligned without merging them.

\subsection{Environment}
\label{sec:environment}

\noindent\textbf{Observation Filtering.} To simulate the selective attention of novices, we gate feedback based on metacognitive behavior. When the agent is in an impulsive \enacting\ state, error traces are redacted, rendering the student blind to the root cause of failure:
{\eqsmall\begin{equation}
    \tilde{o}_t = \begin{cases}
    \text{``[Error]: [\textit{output omitted...}]''} & \text{if } M_{n(t)} = \enacting \\
    o_t & \text{otherwise}
    \end{cases}
\end{equation}}

This forces the agent into realistic trial-and-error loops until the symbolic controller transitions it to any other behaviors.

\noindent\textbf{Execution and Knowledge-Update Gating.}
Code execution is just-in-time per cognitive behavior: \constructing\ writes blindly, \debugging\ runs code first to expose errors, \assessing\ runs solely for evaluation. BKT mastery updates are restricted to \monitoring\ and \reflecting\ states, mirroring how students consolidate knowledge only when consciously evaluating outcomes.

\section{Evaluation}\label{sec:experiments}
We assess whether \beagle achieves authentic student simulation across behavioral (RQ1), epistemic (RQ2), and perceptual (RQ3) dimensions, complemented by ablation studies and generalization analyses across tasks and LLM backbones. The appendix reports the full experimental setup (App.~\ref{app:setup}) and additional analyses: BKT mastery--accuracy adherence (App. Fig.~\ref{fig:bkt_adherence}), $\rho_{\text{behav}}\!\times\!\rho_{\text{persona}}$ decoupling (App.~\ref{sec:decoupling}), Turing test protocol and statistics (App.~\ref{app:turing_test}), tutor-mechanism case study (App.~\ref{app:tutor_casestudy}), and per-task / per-backbone numerics (App.~\ref{app:cross_task_full},~\ref{app:cross_model_full}).

\noindent\textbf{Problem Domain.} Our evaluation suite spans \textbf{three domains} (physics, applied math, and computing) in an established research paradigm where disciplinary knowledge and programming are studied \textit{together} \cite{hutchins2020c2stem, weller2022development}. Specifically, we target \textbf{four tasks}, three of which are in-distribution (class-based kinematics simulation matching the physics$+$code structure of the Snyder corpus and similar to our pilot and the Cohn study): kinematics via a \textit{Particle Simulator} (2D projectile motion with gravity and drag), and dynamics via a \textit{Bouncing Ball} and an \textit{Inclined Plane} (collision dynamics, friction). We additionally target one out-of-distribution task, \textit{Gradient Descent} (applied math, non-OOP), as a natural extension since it shares the same core coding KCs with the in-distribution tasks, while its math KCs (derivatives, numerical differentiation, iterative convergence) belong to the same computational-thinking-in-mathematics curriculum studied in prior work \cite{ye2023integration}. The four tasks share 24 unique knowledge components (11 coding, 8 physics, 5 math); each problem invokes a 11--14 KC subset (App.~\ref{app:problems}). KCs were elicited from the Snyder corpus (1 CS professor + 2 PhD students, grounded in SRL theory); each is an AST-parsable atomic unit, with an IDE Oracle attributing test outcomes per-KC for BKT credit assignment (full mapping in App.~\ref{app:efi_details}).

\noindent\textbf{Datasets.} We use four datasets, each with a distinct role. \textit{(i)~Snyder corpus}~\cite{snyder2024analyzing} ($N=9$ sessions, 227 segments): used once to calibrate the semi-Markov model, with \phigh{}/\plow{} labels assigned at a 0.8 final-score threshold. Snyder is never used for evaluation. \textit{(ii)~Cohn block-based study}~\cite{cohn2026dataset} ($N=10$ sessions, 411 segments): introductory physics in a block-based environment, serving as the primary behavioral ground truth. \textit{(iii)~Python pilot study} ($N=13$ sessions, 306 segments): introductory physics in a Python environment, collected under IRB-approved protocols (\#[anonymized]) with informed consent (App.~\ref{app:test_data}); used as additional behavioral ground truth and as stimuli for the human Turing test. \textit{(iv)~Bielefeld dataset}~\cite{paassen2019python} ($N=15$, keystroke-level): grounds the Gradient Descent task and used 2-second pause protocol for Turing test (App.~\ref{app:turing_test}). The Cohn and pilot datasets are aggregated into a combined evaluation set (denoted Real) for behavioral metrics ($D_{\mathrm{KL}}$, $D_{\text{debug}}$). 

\noindent\textbf{Baselines and Configuration.}
We compare against: (1)~pure prompting (Vanilla, CoT, Few-Shot); (2)~rule-based SimStudent~\cite{matsuda2007predicting}; (3)~exemplar-conditioned LLM-SS~\cite{nguyen2023large} using DCU dataset traces~\cite{azcona2020dcu}; and (4)~CoderAgent~\cite{zhan2025coderagent} with ACT-R-inspired memory (Table~\ref{tab:comparison}). The $+\mathcal{M}$ variants inject \beagle's metacognitive prompt into baselines without the architectural control, isolating prompt effects from architectural ones. We evaluate across six frontier backbones (Gemini 2.0/2.5/3 Flash, GPT-4o-mini, GPT-4.1-mini, Claude Haiku 4.5); unless noted otherwise, runs use Gemini 2.0 Flash with $N=50$ simulations and $T=30$ steps, balancing quality and cost in our multi-call-per-step setting.

\noindent\textbf{Evaluation Metrics.}
We evaluate task performance (solve rate, steps-to-solve, performance gap) and the three fidelity objectives: \textit{behavioral fidelity} via $D_{\mathrm{KL}}$, $D_{\text{debug}}$, and nonlinearity; \textit{epistemic fidelity} via $P_{\text{recur}}$ (operationalizing the error recurrence objective), reaction lag, and \llmcol{Debug} realism; and \textit{perceptual fidelity} via \llmcol{Code} and \llmcol{Lang} style scores, plus a human Turing test ($N=71$ raters; App.~\ref{app:turing_test}) on pilot stimuli, validating fidelity beyond surface realism. All \llmcol{LLM-as-judge} metrics use a 1--3 Likert scale (unfaithful/neutral/faithful) calibrated against human raters following established practice~\citep{cohn2025theory, chakma2026drawsim, burleigh2025beyond}, achieving substantial agreement on the headline \llmcol{Realism} score ($\kappa_w=0.76$~\citep{mchugh2012interrater}; full protocol in App.~\ref{app:llm_judge}). 
Uncertainty for $D_{\mathrm{KL}}$ and $D_{\text{debug}}$ is reported as bootstrap standard error~\cite{efron1992bootstrap} (App.~\ref{app:evaluation-metrics}); solve rate reports the binomial standard error $\sqrt{p(1-p)/N}$ since each run yields a single Bernoulli outcome; remaining metrics report sample standard deviation across runs. (App.~\ref{app:llm_judge}).
\begin{table}[t]
    \centering
    \caption{Main evaluation on Particle Simulator ($N{=}50$, $T{=}30$, Gemini 2.0 Flash). \llmcol{Gray} = LLM-as-Judge ($\kappa_w=0.76$; App.~\ref{app:llm_judge}). $D_{\mathrm{KL}}$, $D_{\text{debug}}$ vs.\ Real. Best \textbf{bold}, 2nd \underline{underline}; '--' = n/a (rule-based).}
    \label{tab:main_results}
    \tiny
    \setlength{\tabcolsep}{1.7pt}
    \centering
    \begin{tabular}{@{}l ccc ccc ccc cc c@{}}
        \toprule
        & \multicolumn{3}{c}{\textbf{Task Perf.}} & \multicolumn{3}{c}{\textbf{Behavioral}} & \multicolumn{3}{c}{\textbf{Epistemic}} & \multicolumn{2}{c}{\textbf{Perceptual}} & \textbf{Overall} \\
        \cmidrule(lr){2-4} \cmidrule(lr){5-7} \cmidrule(lr){8-10} \cmidrule(lr){11-12} \cmidrule(lr){13-13}
        \textbf{Method} & Solve & Steps & Gap & $D_{\mathrm{KL}}$$\downarrow$ & $D_{\text{debug}}$$\downarrow$ & Nonlin$\uparrow$ & $P_{\text{Recur}}$$\uparrow$ & Lag$\uparrow$ & \llmcol{Debug$\uparrow$} & \llmcol{Code} & \llmcol{Lang} & \llmcol{Realism$\uparrow$} \\
        \midrule
        Vanilla & 100.00$\pm$0.00\% & 6$\pm$3 & +0\% & 3.82$\pm$0.12 & 0.09$\pm$0.03 & 0.00$\pm$0.02 & 7.8$\pm$20.3\% & 1.00$\pm$0.00 & 1.20$\pm$0.53 & 2.88$\pm$0.32 & 2.00$\pm$0.49 & 1.18$\pm$0.52 \\
        Vanilla+$\mathcal{M}$ & 92.00$\pm$3.84\% & 11$\pm$7 & +8\% & 3.89$\pm$0.58 & 0.08$\pm$0.03 & 0.02$\pm$0.06 & 10.1$\pm$22.2\% & 1.02$\pm$0.14 & 1.34$\pm$0.71 & \textbf{2.92$\pm$0.27} & 2.04$\pm$0.63 & 1.32$\pm$0.71 \\
        CoT & 100.00$\pm$0.00\% & 7$\pm$4 & +0\% & \underline{0.53$\pm$0.18} & \underline{0.06$\pm$0.02} & 0.02$\pm$0.04 & 26.4$\pm$35.7\% & \underline{2.62$\pm$2.57} & 1.08$\pm$0.39 & 2.26$\pm$0.74 & 2.16$\pm$0.67 & 1.08$\pm$0.39 \\
        CoT+$\mathcal{M}$ & 90.00$\pm$4.24\% & 16$\pm$8 & +12\% & 1.54$\pm$0.14 & 0.52$\pm$0.07 & 0.00$\pm$0.01 & 33.3$\pm$47.1\% & 1.50$\pm$0.50 & 1.00$\pm$0.00 & 1.72$\pm$0.63 & 1.64$\pm$0.59 & 1.00$\pm$0.00 \\
        Few-Shot & 60.00$\pm$6.93\% & 22$\pm$8 & +16\% & 1.76$\pm$0.17 & 0.10$\pm$0.01 & 0.04$\pm$0.05 & 56.7$\pm$37.4\% & 1.00$\pm$0.00 & \underline{1.74$\pm$0.91} & 2.86$\pm$0.40 & \underline{2.36$\pm$0.74} & \underline{1.74$\pm$0.91} \\
        FewShot+$\mathcal{M}$ & 22.00$\pm$5.86\% & 28$\pm$5 & -20\% & 0.63$\pm$0.04 & 0.47$\pm$0.08 & 0.00$\pm$0.00 & 75.0$\pm$40.3\% & 1.00$\pm$0.00 & 1.42$\pm$0.72 & 2.62$\pm$0.52 & 2.30$\pm$0.67 & 1.64$\pm$0.84 \\
        SimStudent & 0.00$\pm$0.00\% & 30$\pm$0 & -- & 0.95$\pm$0.04 & 0.45$\pm$0.02 & 0.00$\pm$0.00 & \textbf{92.0$\pm$18.3\%} & -- & 1.00$\pm$0.00 & 2.20$\pm$0.45 & 1.00$\pm$0.00 & 1.00$\pm$0.00 \\
        LLMSS & 96.00$\pm$2.77\% & 8$\pm$6 & -8\% & 0.83$\pm$0.22 & 0.20$\pm$0.03 & 0.01$\pm$0.02 & 28.3$\pm$41.2\% & 1.65$\pm$1.00 & 1.06$\pm$0.24 & 2.46$\pm$0.61 & 2.30$\pm$0.73 & 1.02$\pm$0.14 \\
        CoderAgent & 56.00$\pm$7.02\% & 19$\pm$11 & +72\% & 3.81$\pm$0.49 & 0.23$\pm$0.03 & \underline{0.05$\pm$0.06} & 49.6$\pm$46.7\% & \textbf{5.47$\pm$5.38} & 1.14$\pm$0.49 & 2.02$\pm$0.58 & 1.80$\pm$0.82 & 1.14$\pm$0.45 \\
        \midrule
        \beagle & 10.00$\pm$4.24\% & 29$\pm$4 & +4\% & \textbf{0.31$\pm$0.05} & \textbf{0.02$\pm$0.01} & \textbf{0.34$\pm$0.13} & \underline{86.2$\pm$21.3\%} & \underline{2.62$\pm$1.98} & \textbf{2.46$\pm$0.85} & \underline{2.90$\pm$0.41} & \textbf{2.62$\pm$0.66} & \textbf{2.44$\pm$0.85} \\
        \bottomrule
    \end{tabular}
\end{table}

\begin{figure}[t]
    \centering
    \begin{minipage}[t]{0.48\textwidth}
        \centering
        \begin{tikzpicture}
            \begin{axis}[
                ybar=0pt,
                bar width=4pt,
                width=0.95\columnwidth,
                height=3.0cm,
                ylabel={Freq. (\%)},
                symbolic x coords={D$\rightarrow$D, C$\rightarrow$C, C$\rightarrow$D, D$\rightarrow$C, C$\rightarrow$A, A$\rightarrow$C, D$\rightarrow$A, A$\rightarrow$D},
                xtick=data,
                xticklabel style={font=\scriptsize, rotate=45, anchor=east, yshift=-3pt},
                ymin=0, ymax=75,
                legend style={
                    at={(0.5, 1.02)},
                    anchor=south,
                    font=\scriptsize,
                    legend columns=-1,
                    draw=none,
                    fill=none,
                    /tikz/every even column/.append style={column sep=0.2cm}
                },
                ymajorgrids=true,
                grid style={dashed,gray!30},
                enlarge x limits=0.10,
                ylabel style={font=\footnotesize},
                tick label style={font=\scriptsize}
            ]

            \addplot[fill=green!60, draw=black!60] coordinates {(D$\rightarrow$D, 23.3) (C$\rightarrow$C, 11.5) (C$\rightarrow$D, 32.9) (D$\rightarrow$C, 31.9) (C$\rightarrow$A, 0.0) (A$\rightarrow$C, 0.2) (D$\rightarrow$A, 0.0) (A$\rightarrow$D, 0.2)};
            \addplot[fill=blue!50, draw=black!60] coordinates {(D$\rightarrow$D, 21.8) (C$\rightarrow$C, 14.0) (C$\rightarrow$D, 32.3) (D$\rightarrow$C, 30.1) (C$\rightarrow$A, 0.0) (A$\rightarrow$C, 1.7) (D$\rightarrow$A, 0.0) (A$\rightarrow$D, 0.2)};
            \addplot[fill=red!50, draw=black!60] coordinates {(D$\rightarrow$D, 7.0) (C$\rightarrow$C, 49.4) (C$\rightarrow$D, 0.0) (D$\rightarrow$C, 3.0) (C$\rightarrow$A, 21.8) (A$\rightarrow$C, 0.4) (D$\rightarrow$A, 0.0) (A$\rightarrow$D, 18.5)};
            \addplot[fill=orange!50, draw=black!60] coordinates {(D$\rightarrow$D, 19.6) (C$\rightarrow$C, 51.4) (C$\rightarrow$D, 19.2) (D$\rightarrow$C, 6.6) (C$\rightarrow$A, 3.2) (A$\rightarrow$C, 0.0) (D$\rightarrow$A, 0.0) (A$\rightarrow$D, 0.0)};
            \addplot[fill=purple!50, draw=black!60] coordinates {(D$\rightarrow$D, 16.7) (C$\rightarrow$C, 58.8) (C$\rightarrow$D, 2.9) (D$\rightarrow$C, 5.9) (C$\rightarrow$A, 9.0) (A$\rightarrow$C, 2.0) (D$\rightarrow$A, 0.0) (A$\rightarrow$D, 4.7)};

            \legend{Real, \beagle, Vanilla, CoT, Few-Shot}
            \end{axis}
        \end{tikzpicture}
        \captionof{figure}{\textbf{Cognitive behavior transitions.} \beagle (blue) closely matches real data.}
        \label{fig:cog_trans_baselines}
    \end{minipage}
    \hfill
    \begin{minipage}[t]{0.48\textwidth}
        \centering
        \begin{tikzpicture}
            \begin{axis}[
                ybar=0pt,
                bar width=4pt,
                width=.95\columnwidth,
                height=3.0cm,
                ylabel={Freq. (\%)},
                symbolic x coords={P$\rightarrow$P, E$\rightarrow$E, P$\rightarrow$E, E$\rightarrow$P, P$\rightarrow$M, M$\rightarrow$P, P$\rightarrow$R, R$\rightarrow$P},
                xtick=data,
                xticklabel style={font=\scriptsize, rotate=45, anchor=east, yshift=-3pt},
                ymin=0, ymax=75,
                ytick={0, 25, 50, 58, 72},
                yticklabels={0, 25, 50, 90, 100},
                legend style={
                    at={(0.5, 1.02)},
                    anchor=south,
                    font=\scriptsize,
                    legend columns=-1,
                    draw=none,
                    fill=none,
                    /tikz/every even column/.append style={column sep=0.2cm}
                },
                ymajorgrids=true,
                grid style={dashed,gray!30},
                enlarge x limits=0.10,
                ylabel style={font=\footnotesize},
                tick label style={font=\scriptsize},
                clip=false
            ]

            \fill[gray!20] (rel axis cs:0, 0.68) rectangle (rel axis cs:1, 0.76);

            \addplot[fill=green!60, draw=black!60] coordinates {(P$\rightarrow$P, 0.0) (E$\rightarrow$E, 0.0) (P$\rightarrow$E, 15.8) (E$\rightarrow$P, 16.7) (P$\rightarrow$M, 10.0) (M$\rightarrow$P, 5.8) (P$\rightarrow$R, 5.0) (R$\rightarrow$P, 3.3)};
            \addplot[fill=blue!50, draw=black!60] coordinates {(P$\rightarrow$P, 0.0) (E$\rightarrow$E, 0.0) (P$\rightarrow$E, 23.5) (E$\rightarrow$P, 12.8) (P$\rightarrow$M, 12.3) (M$\rightarrow$P, 8.4) (P$\rightarrow$R, 3.4) (R$\rightarrow$P, 2.2)};
            \addplot[fill=red!50, draw=black!60] coordinates {(P$\rightarrow$P, 0.0) (E$\rightarrow$E, 0.0) (P$\rightarrow$E, 2.0) (E$\rightarrow$P, 0.0) (P$\rightarrow$M, 2.0) (M$\rightarrow$P, 0.0) (P$\rightarrow$R, 64.7) (R$\rightarrow$P, 0.0)};
            \addplot[fill=orange!50, draw=black!60] coordinates {(P$\rightarrow$P, 0.0) (E$\rightarrow$E, 0.0) (P$\rightarrow$E, 0.0) (E$\rightarrow$P, 0.0) (P$\rightarrow$M, 0.0) (M$\rightarrow$P, 0.0) (P$\rightarrow$R, 72.0) (R$\rightarrow$P, 0.0)};
            \addplot[fill=purple!50, draw=black!60] coordinates {(P$\rightarrow$P, 0.0) (E$\rightarrow$E, 0.0) (P$\rightarrow$E, 46.2) (E$\rightarrow$P, 0.0) (P$\rightarrow$M, 0.0) (M$\rightarrow$P, 0.0) (P$\rightarrow$R, 19.2) (R$\rightarrow$P, 0.0)};

            \legend{Real, \beagle, Vanilla+$\mathcal{M}$, CoT+$\mathcal{M}$, FS+$\mathcal{M}$}
            \end{axis}
        \end{tikzpicture}
        \captionof{figure}{\textbf{Metacognitive behavior transitions.} \beagle closely matches Real. Baselines show extreme P$\rightarrow$R bias (axis break at 50\%).}
        \label{fig:metacog_trans_baselines}
    \end{minipage}
    \vspace{-4mm}
\end{figure}

\subsection{RQ1: Behavioral Fidelity}
\label{sec:rq1}


\begin{wrapfigure}[15]{r}{0.40\linewidth}
    \vspace{-12mm}
    \centering
    \includegraphics[width=\linewidth]{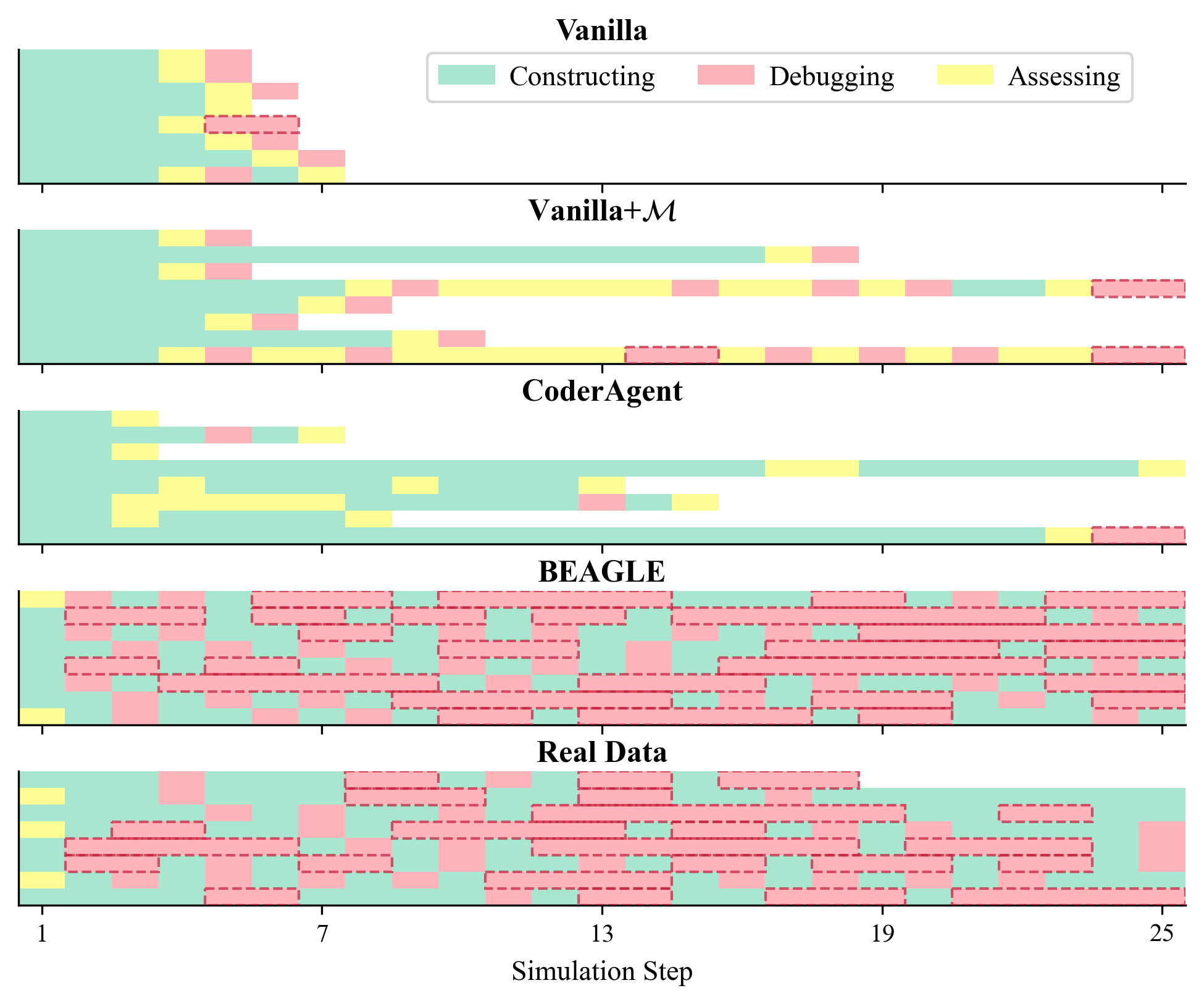}
    \caption{\textbf{Behavioral DNA of cognitive behaviors.} Longer contiguous blocks $=$ closer to authentic behavior; bolded segments are debugging loops.}
    \label{fig:beh-dna}
    \vspace{-1mm}
\end{wrapfigure}

The \textit{linear construction trap} manifests clearly in our baselines. Vanilla solves problems in $\approx$6 steps at 100\% accuracy (Table~\ref{tab:main_results}), while \beagle requires $\approx$29 steps, mirroring real student temporal density. Fig.~\ref{fig:cog_trans_baselines} reveals why: vanilla exhibits dominant constructing self-loops (49.4\%) with zero C$\to$D transitions; \beagle shows balanced bidirectional flow (C$\to$D 32.9\%, D$\to$C 31.9\%), capturing iterative attempt-fail-debug cycles. At the metacognitive level (Fig.~\ref{fig:metacog_trans_baselines}), CoT+$\mathcal{M}$ shows extreme P$\to$R bias ($\sim$99\%), bypassing enacting entirely; \beagle maintains diverse P$\to$E (23.5\%) and E$\to$P (12.8\%) flows. \beagle achieves $D_{\mathrm{KL}}=0.31$ (vs.\ baselines $\geq 0.53$) and $D_{\text{debug}}=0.02$ (vs.\ 0.06--0.52). Fig.~\ref{fig:beh-dna} visualizes this as behavioral DNA: vanilla terminates quickly in \constructing\ segments, while \beagle shows extended interleaving with characteristic \textit{debugging loops} (Details in App. Fig.~\ref{fig:beh-dna-complete}). The +$\mathcal{M}$ variants stress-test context engineering by injecting the same metacognitive vocabulary \beagle uses, yet none beat $D_{\mathrm{KL}}=0.63$ or Realism $=1.64$, and removing the semi-Markov controller while preserving prompts pushes $D_{\mathrm{KL}}$ to 6.81 (\S\ref{sec:ablation}):  \textit{sequence-level student simulation cannot be recovered through context engineering alone; architectural enforcement closes the gap.}

\begin{wrapfigure}{r}{0.39\linewidth}
    \vspace{-3mm}
    \centering
    \begin{tikzpicture}
        \begin{axis}[
            ybar,
            width=\linewidth,
            height=3.3cm,
            ylabel={State (\%)},
            symbolic x coords={Plan, Mon., Refl., Enact},
            xtick=data,
            ymin=0, ymax=62,
            bar width=8pt,
            enlarge x limits=0.15,
            legend style={at={(0.55,0.98)}, anchor=north, font=\scriptsize, draw=none, fill=none},
            tick label style={font=\scriptsize},
            label style={font=\scriptsize},
            grid style={dashed,gray!30},
            nodes near coords,
            every node near coord/.append style={font=\tiny},
        ]
        \addplot[fill=blue!60] coordinates {(Plan, 47.5) (Mon., 24.4) (Refl., 14.6) (Enact, 13.4)};
        \addplot[fill=red!50] coordinates {(Plan, 28.6) (Mon., 14.9) (Refl., 5.6) (Enact, 50.8)};
        \legend{\phigh, \plow}
        \end{axis}
    \end{tikzpicture}
    \caption{\textbf{Metacognitive distribution.} \phigh{} succeed via \planning{}; \plow{} trap in \enacting{}.}
    \label{fig:metacog_distribution}
    \vspace{-3mm}
\end{wrapfigure}
\noindent\textbf{Performance Differentiation.}
Beyond aggregate fidelity, \beagle separates \phigh{} and \plow{} profiles by strategy: \phigh{} performers allocate 72\% of steps to \planning{} and \monitoring{}, while \plow{} performers spend 50.8\% trapped in \enacting{} (Fig.~\ref{fig:metacog_distribution}), reproducing the ``wheel-spinning'' pattern documented in SRL literature~\cite{beck2013wheel}. This profile-driven differentiation translates to solve-rate gaps that scale with backbone capability (e.g., +40\% on Gemini 2.5 Flash; App.~\ref{app:cross_model_full}), whereas vanilla LLMs show no profile sensitivity (+0\%, Table~\ref{tab:main_results}).

\subsection{RQ2: Epistemic Fidelity}
\label{sec:rq2}

\begin{figure}[t]
    \centering
    \hspace*{-0.07\linewidth}%
    \begin{tikzpicture}
        \begin{groupplot}[
            group style={group size=5 by 1, horizontal sep=5pt,
                         ylabels at=edge left, yticklabels at=edge left},
            width=0.275\textwidth, height=3.6cm,
            xlabel={Step $t$}, ylabel={$\overline{P(L_k^{(t)})}$},
            xmin=0, xmax=30, ymin=0.15, ymax=1.0,
            xtick={0,10,20,30}, ytick={0.2,0.4,0.6,0.8,1.0},
            grid=both, grid style={line width=.1pt, draw=gray!18},
            tick label style={font=\scriptsize}, label style={font=\scriptsize},
            title style={font=\small\bfseries, yshift=-3pt, text depth=0.25ex, text height=1ex},
        ]
        \nextgroupplot[title={Reflective}]
            \addplot+[only marks, forget plot, mark=*, mark size=0.9pt, mark options={fill=cReal!90, draw=none},
                      error bars/.cd, y dir=both, y explicit,
                      error bar style={line cap=round, line width=4pt, color=cReal!35, opacity=0.7},
                      error mark=none]
                coordinates {
                    (3, 0.250) +- (0,0.053) (0,0.150)
                    (6, 0.250) +- (0,0.075) (0,0.039)
                    (10,0.350) +- (0,0.060) (0,0.025)
                    (14,0.400) +- (0,0.063) (0,0.050)
                    (18,0.428) +- (0,0.072) (0,0.053)
                    (22,0.450) +- (0,0.050) (0,0.075)
                    (26,0.450) +- (0,0.100) (0,0.056)
                    (30,0.500) +- (0,0.050) (0,0.075)};
            \addplot[name path=Rl_lo, draw=none, forget plot] coordinates {(2,0.236)(3,0.228)(4,0.214)(5,0.212)(6,0.239)(7,0.257)(8,0.279)(9,0.316)(10,0.328)(11,0.347)(12,0.371)(13,0.392)(14,0.423)(15,0.457)(16,0.486)(17,0.492)(18,0.514)(19,0.537)(20,0.555)(21,0.565)(22,0.594)(23,0.605)(24,0.614)(25,0.630)(26,0.655)(27,0.665)(28,0.683)(29,0.694)(30,0.697)};
            \addplot[name path=Rl_hi, draw=none, forget plot] coordinates {(2,0.362)(3,0.434)(4,0.512)(5,0.561)(6,0.638)(7,0.683)(8,0.719)(9,0.758)(10,0.788)(11,0.819)(12,0.833)(13,0.855)(14,0.884)(15,0.900)(16,0.905)(17,0.917)(18,0.934)(19,0.940)(20,0.944)(21,0.958)(22,0.960)(23,0.969)(24,0.976)(25,0.974)(26,0.966)(27,0.974)(28,0.975)(29,0.985)(30,0.990)};
            \addplot[fill=cLow, fill opacity=0.18, draw=none, forget plot] fill between[of=Rl_lo and Rl_hi];
            \addplot[thick, color=cLow] coordinates {(2,0.299)(3,0.331)(4,0.363)(5,0.387)(6,0.439)(7,0.470)(8,0.499)(9,0.537)(10,0.558)(11,0.583)(12,0.602)(13,0.623)(14,0.654)(15,0.679)(16,0.696)(17,0.705)(18,0.724)(19,0.738)(20,0.749)(21,0.761)(22,0.777)(23,0.787)(24,0.795)(25,0.802)(26,0.811)(27,0.820)(28,0.829)(29,0.839)(30,0.844)};
            \addplot[name path=Rh_lo, draw=none, forget plot] coordinates {(2,0.219)(3,0.228)(4,0.227)(5,0.228)(6,0.239)(7,0.252)(8,0.267)(9,0.281)(10,0.301)(11,0.319)(12,0.340)(13,0.370)(14,0.398)(15,0.446)(16,0.487)(17,0.504)(18,0.521)(19,0.557)(20,0.610)(21,0.616)(22,0.625)(23,0.631)(24,0.648)(25,0.659)(26,0.668)(27,0.675)(28,0.679)(29,0.684)(30,0.692)};
            \addplot[name path=Rh_hi, draw=none, forget plot] coordinates {(2,0.417)(3,0.456)(4,0.494)(5,0.527)(6,0.575)(7,0.614)(8,0.666)(9,0.707)(10,0.733)(11,0.765)(12,0.802)(13,0.829)(14,0.849)(15,0.874)(16,0.895)(17,0.907)(18,0.921)(19,0.926)(20,0.931)(21,0.936)(22,0.942)(23,0.947)(24,0.956)(25,0.967)(26,0.972)(27,0.979)(28,0.982)(29,0.989)(30,0.994)};
            \addplot[fill=cHigh, fill opacity=0.18, draw=none, forget plot] fill between[of=Rh_lo and Rh_hi];
            \addplot[thick, color=cHigh] coordinates {(2,0.318)(3,0.342)(4,0.360)(5,0.377)(6,0.407)(7,0.433)(8,0.467)(9,0.494)(10,0.517)(11,0.542)(12,0.571)(13,0.599)(14,0.623)(15,0.660)(16,0.691)(17,0.705)(18,0.721)(19,0.741)(20,0.770)(21,0.776)(22,0.784)(23,0.789)(24,0.802)(25,0.813)(26,0.820)(27,0.827)(28,0.831)(29,0.837)(30,0.843)};
        \nextgroupplot[title={Plan-first}]
            \addplot+[only marks, forget plot, mark=*, mark size=0.9pt, mark options={fill=cReal!90, draw=none},
                      error bars/.cd, y dir=both, y explicit,
                      error bar style={line cap=round, line width=4pt, color=cReal!35, opacity=0.7},
                      error mark=none]
                coordinates {
                    (3, 0.250) +- (0,0.053) (0,0.150)
                    (6, 0.250) +- (0,0.075) (0,0.039)
                    (10,0.350) +- (0,0.060) (0,0.025)
                    (14,0.400) +- (0,0.063) (0,0.050)
                    (18,0.428) +- (0,0.072) (0,0.053)
                    (22,0.450) +- (0,0.050) (0,0.075)
                    (26,0.450) +- (0,0.100) (0,0.056)
                    (30,0.500) +- (0,0.050) (0,0.075)};
            \addplot[name path=Pl_lo, draw=none, forget plot] coordinates {(2,0.218)(3,0.220)(4,0.204)(5,0.203)(6,0.192)(7,0.190)(8,0.196)(9,0.196)(10,0.222)(11,0.227)(12,0.227)(15,0.227)(20,0.249)(25,0.280)(30,0.296)};
            \addplot[name path=Pl_hi, draw=none, forget plot] coordinates {(2,0.326)(3,0.324)(4,0.406)(5,0.454)(6,0.531)(7,0.563)(8,0.602)(9,0.612)(10,0.674)(11,0.694)(12,0.695)(15,0.695)(20,0.762)(25,0.832)(30,0.911)};
            \addplot[fill=cLow, fill opacity=0.18, draw=none, forget plot] fill between[of=Pl_lo and Pl_hi];
            \addplot[thick, color=cLow] coordinates {(2,0.272)(3,0.272)(4,0.305)(5,0.328)(6,0.361)(7,0.377)(8,0.399)(9,0.404)(10,0.448)(11,0.461)(12,0.461)(15,0.461)(20,0.506)(25,0.556)(30,0.603)};
            \addplot[name path=Ph_lo, draw=none, forget plot] coordinates {(2,0.213)(3,0.223)(4,0.218)(5,0.215)(6,0.210)(7,0.236)(8,0.265)(9,0.310)(10,0.316)(11,0.324)(12,0.357)(15,0.359)(20,0.359)(25,0.398)(30,0.400)};
            \addplot[name path=Ph_hi, draw=none, forget plot] coordinates {(2,0.282)(3,0.371)(4,0.469)(5,0.473)(6,0.562)(7,0.578)(8,0.675)(9,0.716)(10,0.752)(11,0.776)(12,0.814)(15,0.846)(20,0.891)(25,0.927)(30,0.955)};
            \addplot[fill=cHigh, fill opacity=0.18, draw=none, forget plot] fill between[of=Ph_lo and Ph_hi];
            \addplot[thick, color=cHigh] coordinates {(2,0.248)(3,0.297)(4,0.344)(5,0.344)(6,0.386)(7,0.407)(8,0.470)(9,0.513)(10,0.534)(11,0.550)(12,0.586)(15,0.602)(20,0.625)(25,0.662)(30,0.678)};
        \nextgroupplot[title={Test-driven}]
            \addplot+[only marks, forget plot, mark=*, mark size=0.9pt, mark options={fill=cReal!90, draw=none},
                      error bars/.cd, y dir=both, y explicit,
                      error bar style={line cap=round, line width=4pt, color=cReal!35, opacity=0.7},
                      error mark=none]
                coordinates {
                    (3, 0.250) +- (0,0.053) (0,0.150)
                    (6, 0.250) +- (0,0.075) (0,0.039)
                    (10,0.350) +- (0,0.060) (0,0.025)
                    (14,0.400) +- (0,0.063) (0,0.050)
                    (18,0.428) +- (0,0.072) (0,0.053)
                    (22,0.450) +- (0,0.050) (0,0.075)
                    (26,0.450) +- (0,0.100) (0,0.056)
                    (30,0.500) +- (0,0.050) (0,0.075)};
            \addplot[name path=Tl_lo, draw=none, forget plot] coordinates {(2,0.232)(5,0.204)(10,0.166)(15,0.184)(20,0.205)(25,0.201)(30,0.202)};
            \addplot[name path=Tl_hi, draw=none, forget plot] coordinates {(2,0.317)(5,0.376)(10,0.441)(15,0.482)(20,0.576)(25,0.590)(30,0.632)};
            \addplot[fill=cLow, fill opacity=0.18, draw=none, forget plot] fill between[of=Tl_lo and Tl_hi];
            \addplot[thick, color=cLow] coordinates {(2,0.274)(5,0.290)(10,0.303)(15,0.333)(20,0.390)(25,0.396)(30,0.417)};
            \addplot[name path=Th_lo, draw=none, forget plot] coordinates {(2,0.318)(5,0.267)(10,0.263)(15,0.285)(20,0.278)(25,0.260)(30,0.289)};
            \addplot[name path=Th_hi, draw=none, forget plot] coordinates {(2,0.501)(5,0.417)(10,0.495)(15,0.631)(20,0.657)(25,0.712)(30,0.719)};
            \addplot[fill=cHigh, fill opacity=0.18, draw=none, forget plot] fill between[of=Th_lo and Th_hi];
            \addplot[thick, color=cHigh] coordinates {(2,0.410)(5,0.342)(10,0.379)(15,0.458)(20,0.468)(25,0.486)(30,0.504)};
        \nextgroupplot[title={Depth-first}]
            \addplot+[only marks, forget plot, mark=*, mark size=0.9pt, mark options={fill=cReal!90, draw=none},
                      error bars/.cd, y dir=both, y explicit,
                      error bar style={line cap=round, line width=4pt, color=cReal!35, opacity=0.7},
                      error mark=none]
                coordinates {
                    (3, 0.250) +- (0,0.053) (0,0.150)
                    (6, 0.250) +- (0,0.075) (0,0.039)
                    (10,0.350) +- (0,0.060) (0,0.025)
                    (14,0.400) +- (0,0.063) (0,0.050)
                    (18,0.428) +- (0,0.072) (0,0.053)
                    (22,0.450) +- (0,0.050) (0,0.075)
                    (26,0.450) +- (0,0.100) (0,0.056)
                    (30,0.500) +- (0,0.050) (0,0.075)};
            \addplot[name path=Dl_lo, draw=none, forget plot] coordinates {(2,0.241)(5,0.215)(10,0.263)(15,0.259)(20,0.267)(25,0.287)(30,0.305)};
            \addplot[name path=Dl_hi, draw=none, forget plot] coordinates {(2,0.364)(5,0.570)(10,0.685)(15,0.762)(20,0.777)(25,0.802)(30,0.826)};
            \addplot[fill=cLow, fill opacity=0.18, draw=none, forget plot] fill between[of=Dl_lo and Dl_hi];
            \addplot[thick, color=cLow] coordinates {(2,0.303)(5,0.392)(10,0.474)(15,0.510)(20,0.522)(25,0.545)(30,0.566)};
            \addplot[name path=Dh_lo, draw=none, forget plot] coordinates {(2,0.234)(5,0.225)(10,0.202)(15,0.195)(20,0.214)(25,0.234)(30,0.256)};
            \addplot[name path=Dh_hi, draw=none, forget plot] coordinates {(2,0.394)(5,0.395)(10,0.475)(15,0.534)(20,0.610)(25,0.695)(30,0.756)};
            \addplot[fill=cHigh, fill opacity=0.18, draw=none, forget plot] fill between[of=Dh_lo and Dh_hi];
            \addplot[thick, color=cHigh] coordinates {(2,0.314)(5,0.310)(10,0.339)(15,0.365)(20,0.412)(25,0.464)(30,0.506)};
        \nextgroupplot[title={Tinker}]
            \addplot+[only marks, forget plot, mark=*, mark size=0.9pt, mark options={fill=cReal!90, draw=none},
                      error bars/.cd, y dir=both, y explicit,
                      error bar style={line cap=round, line width=4pt, color=cReal!35, opacity=0.7},
                      error mark=none]
                coordinates {
                    (3, 0.250) +- (0,0.053) (0,0.150)
                    (6, 0.250) +- (0,0.075) (0,0.039)
                    (10,0.350) +- (0,0.060) (0,0.025)
                    (14,0.400) +- (0,0.063) (0,0.050)
                    (18,0.428) +- (0,0.072) (0,0.053)
                    (22,0.450) +- (0,0.050) (0,0.075)
                    (26,0.450) +- (0,0.100) (0,0.056)
                    (30,0.500) +- (0,0.050) (0,0.075)};
            \addplot[name path=Kl_lo, draw=none, forget plot] coordinates {(2,0.223)(5,0.199)(10,0.232)(15,0.275)(20,0.300)(25,0.310)(30,0.339)};
            \addplot[name path=Kl_hi, draw=none, forget plot] coordinates {(2,0.342)(5,0.466)(10,0.596)(15,0.735)(20,0.814)(25,0.835)(30,0.864)};
            \addplot[fill=cLow, fill opacity=0.18, draw=none, forget plot] fill between[of=Kl_lo and Kl_hi];
            \addplot[thick, color=cLow] coordinates {(2,0.283)(5,0.332)(10,0.414)(15,0.505)(20,0.557)(25,0.573)(30,0.601)};
            \addplot[name path=Kh_lo, draw=none, forget plot] coordinates {(2,0.252)(5,0.210)(10,0.165)(15,0.208)(20,0.252)(25,0.314)(30,0.313)};
            \addplot[name path=Kh_hi, draw=none, forget plot] coordinates {(2,0.346)(5,0.425)(10,0.564)(15,0.686)(20,0.705)(25,0.775)(30,0.802)};
            \addplot[fill=cHigh, fill opacity=0.18, draw=none, forget plot] fill between[of=Kh_lo and Kh_hi];
            \addplot[thick, color=cHigh] coordinates {(2,0.299)(5,0.318)(10,0.365)(15,0.447)(20,0.478)(25,0.545)(30,0.557)};
        \end{groupplot}
        \node[anchor=south, rotate=-90, font=\scriptsize, color=cReal!80!black]
            at ($(group c5r1.east) + (4pt, 0)$)
            {Study mastery rubrics};
    \end{tikzpicture}
    \caption{\textbf{BKT mastery growth conditioned on dominant strategy} (Particle Sim., pooled across six backbones). \textcolor{cLow}{Orange} = low performer, \textcolor{cHigh}{blue} = high performer; shaded halos show $\pm 1\sigma$ across runs. \textcolor{cReal!80!black}{Green} capsules: rubric mastery from the Cohn block-based study (Inter-Quartile Range with median dot).}
    \label{fig:bkt_strategy_5panel_realref}
\end{figure}

Authentic simulation requires enforcing the \textit{curse of incompetence}~\cite{ehrlinger2008unskilled}: novices lack both the skill to solve a problem and the metacognition to recognize their errors. Real novices struggle repeatedly with the same errors due to limited diagnostic expertise, whereas LLMs exhibit expert-like error recognition, a fundamental ``gap between LLM's underlying reasoning process and human cognitive processes''~\cite{liu2025llmsmistakes}. We measure this through the error recurrence rate, $P_{\text{recur}}$ (App.~\ref{app:evaluation-metrics}): the proportion of runs in which the same error type recurs. Vanilla LLMs achieve only 7.8\% recurrence (Table~\ref{tab:main_results}), quickly resolving diverse errors; \beagle sits at 86.2\%, within the realistic novice envelope bracketed by Vanilla (7.8\%, far below) and rule-based SimStudent (92.0\%, which saturates the upper anchor through rigid production rules that mechanically repeat errors rather than higher novice authenticity). Two mechanisms enable \beagle's epistemic fidelity: observation filtering withholds diagnostic feedback during impulsive \enacting{} states, preventing premature error recognition; and the Strategist/Executor split forces the Executor to implement flawed plans without silent correction (ablation confirms merging them reduces recurrence by 21\%). Beyond aggregate recurrence, \textit{strategy-conditional} BKT mastery growth (Fig.~\ref{fig:bkt_strategy_5panel_realref}) shows that \textit{Reflective} trajectories saturate near 0.85 prior-independently, whereas \textit{Plan-first}, \textit{Test-driven}, and \textit{Depth-first} show clear high > low gaps, confirming strategy choice shapes mastery trajectories, and BKT meaningfully tracks this in simulated students.

\subsection{RQ3: Perceptual Fidelity}
\label{sec:rq3}

\begin{wrapfigure}{r}{0.34\linewidth}
    \vspace{-14mm}
    \centering
    \begin{tikzpicture}
        \begin{axis}[
            width=1\linewidth,
            height=3.13cm,
            domain=-3:3,
            samples=121,
            xlabel={Evidence},
            ylabel={Density},
            ymin=0, ymax=0.55,
            axis y line=left,
            axis x line=bottom,
            ytick=\empty,
            tick label style={font=\footnotesize},
            label style={font=\footnotesize},
            xtick={-2,0,2},
            xlabel style={yshift=2pt},
            ylabel style={yshift=-6pt, font=\footnotesize},
            legend style={at={(0.02,0.98)}, anchor=north west,
                          font=\scriptsize, draw=none, fill=none,
                          inner sep=1pt, row sep=-3pt},
            legend cell align={left},
            no markers,
            clip=false,
        ]
        \addplot[fill=red,  fill opacity=0.28, draw=red!75!black,  line width=0.9pt]
            {1/sqrt(2*pi)*exp(-x^2/2)} \closedcycle;
        \addplot[fill=blue, fill opacity=0.28, draw=blue!75!black, line width=0.9pt]
            {1/sqrt(2*pi)*exp(-(x-0.15)^2/2)} \closedcycle;
        \legend{AI, Real}

        \draw[fill=red!80!black,  draw=white, line width=0.3pt]
            (axis cs:0,    0.399) circle (1.6pt);
        \draw[fill=blue!80!black, draw=white, line width=0.3pt]
            (axis cs:0.15, 0.399) circle (1.6pt);

        \draw[black!75, line width=0.6pt]
            (axis cs:0,    0.45) -- (axis cs:0.15, 0.45);
        \draw[black!75, line width=0.6pt]
            (axis cs:0,    0.44) -- (axis cs:0,    0.46);
        \draw[black!75, line width=0.6pt]
            (axis cs:0.15, 0.44) -- (axis cs:0.15, 0.46);

        \draw[black!60, line width=0.4pt]
            (axis cs:0.075,0.45) -- (axis cs:1.0,0.50);
        \node[font=\scriptsize, anchor=west, black!75]
            at (axis cs:1.0,0.50) {$d'{=}0.15$};
        \end{axis}
    \end{tikzpicture}

    \vspace{-4mm}
    \begin{tikzpicture}[x=1pt, y=1pt]
        \def\W{120}\def\H{5}
        \fill[blue!60]   (0,0)         rectangle (0.36*\W,\H);
        \fill[blue!30]   (0.36*\W,0)   rectangle (0.68*\W,\H);
        \fill[orange!60] (0.68*\W,0)   rectangle (0.79*\W,\H);
        \fill[green!50]  (0.79*\W,0)   rectangle (0.88*\W,\H);
        \fill[red!40]    (0.88*\W,0)   rectangle (0.95*\W,\H);
        \fill[gray!50]   (0.95*\W,0)   rectangle (\W,\H);
        \draw[black!50, line width=0.3pt] (0,0) rectangle (\W,\H);
        \node[font=\tiny, anchor=north] at (0.18*\W,-0.5) {Grad 36};
        \node[font=\tiny, anchor=north] at (0.52*\W,-0.5) {UG 32};
        \draw[black!55, line width=0.3pt] (0.735*\W,\H) -- (1.05*\W,-9);
        \node[font=\tiny, anchor=west, black!75] at (1.05*\W,-9) {Prof 11};
        \draw[black!55, line width=0.3pt] (0.835*\W,\H) -- (1.05*\W,-3);
        \node[font=\tiny, anchor=west, black!75] at (1.05*\W,-3) {Hob 9};
        \draw[black!55, line width=0.3pt] (0.915*\W,\H) -- (1.05*\W,3);
        \node[font=\tiny, anchor=west, black!75] at (1.05*\W,3) {Beg 7};
        \draw[black!55, line width=0.3pt] (0.975*\W,\H) -- (1.05*\W,9);
        \node[font=\tiny, anchor=west, black!75] at (1.05*\W,9) {Edu 5};
    \end{tikzpicture}
    \caption{Turing test SDT and participant composition (\%); $N{=}71$.}
    \label{fig:sdt_main}
    \vspace{-10mm}
\end{wrapfigure}
In the LLM-as-Judge evaluation, \beagle traces achieve the highest realism score (2.44/3.00, Table~\ref{tab:main_results}), far exceeding baselines (1.00--1.74). We further conducted a human Turing test with $N=71$ participants (852 total classifications; each rated 12 traces across 3 problem scenarios, 2 real and 2 simulated per scenario) spanning graduate students (Grad), undergraduates (UG), professionals (Prof), hobbyists (Hob), beginners (Beg), and educators (Edu); composition shown in Fig.~\ref{fig:sdt_main}. Participants were shown real student traces and \beagle-generated traces and were asked to view each trace as a full temporal progression of roughly 25+ snapshots in an animated viewer before classifying each as real or AI-generated. Overall accuracy was 52.8\%, not statistically distinguishable from chance ($p = 0.053$).
Signal Detection Theory (SDT) analysis (Fig.~\ref{fig:sdt_main}), which separates true perceptual ability from response tendencies, reveals a bias toward labeling any trace as real ($c = -0.24$). This bias manifests asymmetrically: participants correctly identified 62.2\% of real traces but only 43.4\% of AI traces, worse than random guessing. To formally test whether \beagle traces are perceptually equivalent to real data (rather than merely failing to reject a difference), we applied a Two One-Sided Tests (TOST) equivalence procedure. Results confirm that the discriminability index ($d' = 0.15$, where $d' = 0$ indicates perfect indistinguishability) falls significantly within the equivalence bound of $\pm 0.3$ ($p_{\text{TOST}} = 0.038$). Notably, evaluator expertise offered no advantage: graduate students and professionals performed no better than beginners (App.~\ref{app:turing_test}), suggesting the discriminability ceiling is set by the trace itself rather than by reader experience. Together, these analyses establish that \beagle traces are perceptually indistinguishable from the real data (see App.~\ref{app:turing_test} for details). 

\subsection{Ablations and Generalization}
\label{sec:ablation}

\begin{table}[t]
\centering
\caption{\textbf{Ablation Study} ($N=50$, $T=30$,Gemini 2.0 Flash). $D_{\mathrm{KL}}$ = behavioral divergence vs.\ combined test data. Combined Agent merges Strategist/Executor into a single LLM.}
\label{tab:ablation}
\footnotesize
\setlength{\tabcolsep}{5pt}
\begin{tabular}{@{}l c c c c c c@{}}
    \toprule
    \textbf{Variant} & Steps & $D_{\mathrm{KL}}$$\downarrow$ & $P_{\text{recur}}$$\uparrow$ & \llmcol{Debug$\uparrow$} & \llmcol{Lang$\uparrow$} & \llmcol{Realism$\uparrow$} \\
    \midrule
    Full \beagle               & 29$\pm$4 & 0.31$\pm$0.05 & 86.2$\pm$21.3\% & 2.46$\pm$0.85 & 2.62$\pm$0.66 & 2.44$\pm$0.85 \\
    No BKT                    & 29$\pm$4 & 0.26$\pm$0.05 & 90.2$\pm$16.1\% & 2.33$\pm$0.91 & 2.49$\pm$0.73 & 2.33$\pm$0.84 \\
    No semi-Markov            & 27$\pm$4 & 6.81$\pm$0.29 & 83.2$\pm$17.5\% & 2.38$\pm$0.87 & 2.62$\pm$0.60 & 2.38$\pm$0.85 \\
    No Interrupts             & 29$\pm$4 & 0.27$\pm$0.05 & 91.0$\pm$17.4\% & 2.68$\pm$0.65 & 2.60$\pm$0.66 & 2.54$\pm$0.78 \\
    No $\Gamma_{\text{exec}}$ & 30$\pm$1 & 0.27$\pm$0.05 & 83.2$\pm$18.2\% & 2.36$\pm$0.82 & 2.70$\pm$0.54 & 2.38$\pm$0.77 \\
    No $\Gamma_{\text{strat}}$ & 29$\pm$3 & 0.29$\pm$0.05 & 86.3$\pm$19.9\% & 2.42$\pm$0.87 & 2.70$\pm$0.54 & 2.42$\pm$0.87 \\
    Combined Agent            & 26$\pm$7 & 0.32$\pm$0.05 & 65.3$\pm$20.3\% & 1.76$\pm$0.95 & 2.62$\pm$0.56 & 1.88$\pm$0.93 \\
    \bottomrule
\end{tabular}
\end{table}

\begin{figure}[t]
\centering
\hspace*{-0.07\linewidth}%
\begin{tikzpicture}
\begin{groupplot}[
    group style={group size=5 by 1, horizontal sep=5pt},
    width=0.275\textwidth,
    height=3.2cm,
    enlarge y limits=0.18,
    title style={font=\footnotesize, yshift=-6pt},
    tick label style={font=\scriptsize},
    label style={font=\scriptsize},
    symbolic y coords={GradDesc,InclinedPlane,BouncingBall,ParticleSim},
    ytick=data,
    yticklabel style={font=\scriptsize},
    xmajorgrids, grid style={dashed, gray!20},
    ytick style={draw=none},
]

\nextgroupplot[title={$D_{\mathrm{KL}}\downarrow$}, xmin=0, xmax=8.3,
    yticklabels={Gradient Desc., Inclined Plane, Bouncing Ball, Particle Sim.}]
\draw[gray!55, line cap=round, line width=4.5pt, opacity=0.4] (axis cs:1.22,ParticleSim) -- (axis cs:1.54,ParticleSim);
\draw[gray!55, line cap=round, line width=4.5pt, opacity=0.4] (axis cs:3.24,BouncingBall) -- (axis cs:3.72,BouncingBall);
\draw[gray!55, line cap=round, line width=4.5pt, opacity=0.4] (axis cs:2.60,InclinedPlane) -- (axis cs:3.00,InclinedPlane);
\draw[gray!55, line cap=round, line width=4.5pt, opacity=0.4] (axis cs:5.76,GradDesc) -- (axis cs:7.06,GradDesc);
\draw[blue!50, line cap=round, line width=4.5pt, opacity=0.4] (axis cs:0.34,ParticleSim) -- (axis cs:0.52,ParticleSim);
\draw[blue!50, line cap=round, line width=4.5pt, opacity=0.4] (axis cs:0.15,BouncingBall) -- (axis cs:0.25,BouncingBall);
\draw[blue!50, line cap=round, line width=4.5pt, opacity=0.4] (axis cs:0.26,InclinedPlane) -- (axis cs:0.40,InclinedPlane);
\draw[blue!50, line cap=round, line width=4.5pt, opacity=0.4] (axis cs:0.26,GradDesc) -- (axis cs:0.40,GradDesc);
\draw[gray!50, line width=1.4pt] (axis cs:1.38,ParticleSim) -- (axis cs:0.43,ParticleSim);
\draw[gray!50, line width=1.4pt] (axis cs:3.48,BouncingBall) -- (axis cs:0.20,BouncingBall);
\draw[gray!50, line width=1.4pt] (axis cs:2.80,InclinedPlane) -- (axis cs:0.33,InclinedPlane);
\draw[gray!50, line width=1.4pt] (axis cs:6.41,GradDesc) -- (axis cs:0.33,GradDesc);
\addplot[only marks, mark=*, mark options={fill=gray!75, draw=gray!90!black}, mark size=2pt] coordinates {
  (1.38,ParticleSim) (3.48,BouncingBall) (2.80,InclinedPlane) (6.41,GradDesc)
};
\addplot[only marks, mark=*, mark options={fill=blue!70!black, draw=blue!90!black}, mark size=2.6pt] coordinates {
  (0.43,ParticleSim) (0.20,BouncingBall) (0.33,InclinedPlane) (0.33,GradDesc)
};

\nextgroupplot[title={$P_{\mathrm{recur}}\uparrow$ (\%)}, xmin=20, xmax=120, yticklabel=\empty]
\draw[gray!55, line cap=round, line width=4.5pt, opacity=0.4] (axis cs:25,ParticleSim) -- (axis cs:96.3,ParticleSim);
\draw[gray!55, line cap=round, line width=4.5pt, opacity=0.4] (axis cs:38.6,BouncingBall) -- (axis cs:103.6,BouncingBall);
\draw[gray!55, line cap=round, line width=4.5pt, opacity=0.4] (axis cs:41.1,InclinedPlane) -- (axis cs:109.5,InclinedPlane);
\draw[gray!55, line cap=round, line width=4.5pt, opacity=0.4] (axis cs:33.1,GradDesc) -- (axis cs:118.7,GradDesc);
\draw[blue!50, line cap=round, line width=4.5pt, opacity=0.4] (axis cs:56.4,ParticleSim) -- (axis cs:102.4,ParticleSim);
\draw[blue!50, line cap=round, line width=4.5pt, opacity=0.4] (axis cs:67.1,BouncingBall) -- (axis cs:104.1,BouncingBall);
\draw[blue!50, line cap=round, line width=4.5pt, opacity=0.4] (axis cs:60.0,InclinedPlane) -- (axis cs:103.4,InclinedPlane);
\draw[blue!50, line cap=round, line width=4.5pt, opacity=0.4] (axis cs:69.4,GradDesc) -- (axis cs:103.4,GradDesc);
\draw[gray!50, line width=1.4pt] (axis cs:49.3,ParticleSim) -- (axis cs:79.4,ParticleSim);
\draw[gray!50, line width=1.4pt] (axis cs:71.1,BouncingBall) -- (axis cs:85.6,BouncingBall);
\draw[gray!50, line width=1.4pt] (axis cs:75.3,InclinedPlane) -- (axis cs:81.7,InclinedPlane);
\draw[gray!50, line width=1.4pt] (axis cs:75.9,GradDesc) -- (axis cs:86.4,GradDesc);
\addplot[only marks, mark=*, mark options={fill=gray!75, draw=gray!90!black}, mark size=2pt] coordinates {
  (49.3,ParticleSim) (71.1,BouncingBall) (75.3,InclinedPlane) (75.9,GradDesc)
};
\addplot[only marks, mark=*, mark options={fill=blue!70!black, draw=blue!90!black}, mark size=2.6pt] coordinates {
  (79.4,ParticleSim) (85.6,BouncingBall) (81.7,InclinedPlane) (86.4,GradDesc)
};

\nextgroupplot[title={Debug$\uparrow$}, xmin=0.5, xmax=3.4, yticklabel=\empty]
\draw[gray!55, line cap=round, line width=4.5pt, opacity=0.4] (axis cs:0.72,ParticleSim) -- (axis cs:1.62,ParticleSim);
\draw[gray!55, line cap=round, line width=4.5pt, opacity=0.4] (axis cs:0.61,BouncingBall) -- (axis cs:1.99,BouncingBall);
\draw[gray!55, line cap=round, line width=4.5pt, opacity=0.4] (axis cs:0.60,InclinedPlane) -- (axis cs:1.80,InclinedPlane);
\draw[gray!55, line cap=round, line width=4.5pt, opacity=0.4] (axis cs:0.70,GradDesc) -- (axis cs:1.50,GradDesc);
\draw[blue!50, line cap=round, line width=4.5pt, opacity=0.4] (axis cs:2.15,ParticleSim) -- (axis cs:3.39,ParticleSim);
\draw[blue!50, line cap=round, line width=4.5pt, opacity=0.4] (axis cs:0.93,BouncingBall) -- (axis cs:2.67,BouncingBall);
\draw[blue!50, line cap=round, line width=4.5pt, opacity=0.4] (axis cs:1.02,InclinedPlane) -- (axis cs:2.92,InclinedPlane);
\draw[blue!50, line cap=round, line width=4.5pt, opacity=0.4] (axis cs:1.07,GradDesc) -- (axis cs:2.93,GradDesc);
\draw[gray!50, line width=1.4pt] (axis cs:1.17,ParticleSim) -- (axis cs:2.77,ParticleSim);
\draw[gray!50, line width=1.4pt] (axis cs:1.30,BouncingBall) -- (axis cs:1.80,BouncingBall);
\draw[gray!50, line width=1.4pt] (axis cs:1.20,InclinedPlane) -- (axis cs:1.97,InclinedPlane);
\draw[gray!50, line width=1.4pt] (axis cs:1.10,GradDesc) -- (axis cs:2.00,GradDesc);
\addplot[only marks, mark=*, mark options={fill=gray!75, draw=gray!90!black}, mark size=2pt] coordinates {
  (1.17,ParticleSim) (1.30,BouncingBall) (1.20,InclinedPlane) (1.10,GradDesc)
};
\addplot[only marks, mark=*, mark options={fill=blue!70!black, draw=blue!90!black}, mark size=2.6pt] coordinates {
  (2.77,ParticleSim) (1.80,BouncingBall) (1.97,InclinedPlane) (2.00,GradDesc)
};

\nextgroupplot[title={Lang$\uparrow$}, xmin=0.8, xmax=3.4, yticklabel=\empty]
\draw[gray!55, line cap=round, line width=4.5pt, opacity=0.4] (axis cs:1.55,ParticleSim) -- (axis cs:2.45,ParticleSim);
\draw[gray!55, line cap=round, line width=4.5pt, opacity=0.4] (axis cs:1.28,BouncingBall) -- (axis cs:2.78,BouncingBall);
\draw[gray!55, line cap=round, line width=4.5pt, opacity=0.4] (axis cs:1.68,InclinedPlane) -- (axis cs:2.98,InclinedPlane);
\draw[gray!55, line cap=round, line width=4.5pt, opacity=0.4] (axis cs:0.89,GradDesc) -- (axis cs:1.85,GradDesc);
\draw[blue!50, line cap=round, line width=4.5pt, opacity=0.4] (axis cs:2.40,ParticleSim) -- (axis cs:3.20,ParticleSim);
\draw[blue!50, line cap=round, line width=4.5pt, opacity=0.4] (axis cs:2.13,BouncingBall) -- (axis cs:3.21,BouncingBall);
\draw[blue!50, line cap=round, line width=4.5pt, opacity=0.4] (axis cs:1.88,InclinedPlane) -- (axis cs:3.12,InclinedPlane);
\draw[blue!50, line cap=round, line width=4.5pt, opacity=0.4] (axis cs:1.75,GradDesc) -- (axis cs:3.19,GradDesc);
\draw[gray!50, line width=1.4pt] (axis cs:2.00,ParticleSim) -- (axis cs:2.80,ParticleSim);
\draw[gray!50, line width=1.4pt] (axis cs:2.03,BouncingBall) -- (axis cs:2.67,BouncingBall);
\draw[gray!50, line width=1.4pt] (axis cs:2.33,InclinedPlane) -- (axis cs:2.50,InclinedPlane);
\draw[gray!50, line width=1.4pt] (axis cs:1.37,GradDesc) -- (axis cs:2.47,GradDesc);
\addplot[only marks, mark=*, mark options={fill=gray!75, draw=gray!90!black}, mark size=2pt] coordinates {
  (2.00,ParticleSim) (2.03,BouncingBall) (2.33,InclinedPlane) (1.37,GradDesc)
};
\addplot[only marks, mark=*, mark options={fill=blue!70!black, draw=blue!90!black}, mark size=2.6pt] coordinates {
  (2.80,ParticleSim) (2.67,BouncingBall) (2.50,InclinedPlane) (2.47,GradDesc)
};

\nextgroupplot[title={Realism$\uparrow$}, xmin=0.5, xmax=3.4, yticklabel=\empty]
\draw[gray!55, line cap=round, line width=4.5pt, opacity=0.4] (axis cs:0.72,ParticleSim) -- (axis cs:1.62,ParticleSim);
\draw[gray!55, line cap=round, line width=4.5pt, opacity=0.4] (axis cs:0.65,BouncingBall) -- (axis cs:1.69,BouncingBall);
\draw[gray!55, line cap=round, line width=4.5pt, opacity=0.4] (axis cs:0.65,InclinedPlane) -- (axis cs:1.69,InclinedPlane);
\draw[gray!55, line cap=round, line width=4.5pt, opacity=0.4] (axis cs:0.71,GradDesc) -- (axis cs:1.43,GradDesc);
\draw[blue!50, line cap=round, line width=4.5pt, opacity=0.4] (axis cs:2.15,ParticleSim) -- (axis cs:3.39,ParticleSim);
\draw[blue!50, line cap=round, line width=4.5pt, opacity=0.4] (axis cs:1.14,BouncingBall) -- (axis cs:2.86,BouncingBall);
\draw[blue!50, line cap=round, line width=4.5pt, opacity=0.4] (axis cs:1.07,InclinedPlane) -- (axis cs:2.93,InclinedPlane);
\draw[blue!50, line cap=round, line width=4.5pt, opacity=0.4] (axis cs:1.16,GradDesc) -- (axis cs:2.90,GradDesc);
\draw[gray!50, line width=1.4pt] (axis cs:1.17,ParticleSim) -- (axis cs:2.77,ParticleSim);
\draw[gray!50, line width=1.4pt] (axis cs:1.17,BouncingBall) -- (axis cs:2.00,BouncingBall);
\draw[gray!50, line width=1.4pt] (axis cs:1.17,InclinedPlane) -- (axis cs:2.00,InclinedPlane);
\draw[gray!50, line width=1.4pt] (axis cs:1.07,GradDesc) -- (axis cs:2.03,GradDesc);
\addplot[only marks, mark=*, mark options={fill=gray!75, draw=gray!90!black}, mark size=2pt] coordinates {
  (1.17,ParticleSim) (1.17,BouncingBall) (1.17,InclinedPlane) (1.07,GradDesc)
};
\addplot[only marks, mark=*, mark options={fill=blue!70!black, draw=blue!90!black}, mark size=2.6pt] coordinates {
  (2.77,ParticleSim) (2.00,BouncingBall) (2.00,InclinedPlane) (2.03,GradDesc)
};

\end{groupplot}
\end{tikzpicture}
\caption{\textbf{Cross-task generalization (\beagle vs.\ best non-\beagle baseline, per task).} \textcolor{gray!75!black}{$\bullet$}~Best baseline (CoT+$\mathcal{M}$ or CoderAgent, whichever is stronger per cell); \textcolor{blue!70!black}{$\bullet$}~\beagle. Connecting lines show the gap; translucent halos show $\pm 1$ std. All simulations are generated using Gemini 2.5 Flash. Across the four tasks, \beagle shrinks behavioral divergence by $\geq 3\times$, raises error recurrence to $\geq 79\%$, and improves all three perceptual scores (Debug, Lang, Realism). Full numerics in App.~\ref{app:cross_task_full}.}
\label{fig:cross_task_viz}
\end{figure}

\noindent\textbf{Ablation Study.} We further evaluate individual modules (Table~\ref{tab:ablation}). Three key findings emerge: (1) \textit{The semi-Markov controller is the primary driver of behavioral fidelity.} Its removal causes catastrophic divergence from real student distributions ($D_{\mathrm{KL}}$ rises to 6.81, compared to $\leq$0.32 for all other ablations), confirming that standard LLMs inherently default to linear construction. (2) \textit{Architectural separation is essential for epistemic fidelity.} Merging Strategist/Executor into a Combined Agent produces the largest realism drop ($2.44 \to 1.88$) and reduces error recurrence by 21\% ($86.2\% \to 65.3\%$). Without this task framing, the LLM reverts to superficial, skilled error resolution, silently correcting mistakes rather than exhibiting diagnostic incompetence. (3) \textit{BKT contributes to perceptual realism.} Removing BKT degrades Debug realism ($2.46 \to 2.33$) and overall Realism ($2.44 \to 2.33$), though interestingly increases error recurrence ($86.2\% \to 90.2\%$), suggesting BKT's role is shaping \emph{how} errors manifest rather than \emph{whether} they persist (App.~\ref{app:tutor_casestudy}).

\noindent\textbf{Generalization.} We test two transfer settings. \textit{Cross-task} (Fig.~\ref{fig:cross_task_viz}, App.~\ref{app:cross_task_full}): \beagle runs on the three remaining problems: two in-distribution physics tasks (Bouncing Ball, Inclined Plane) and one out-of-distribution applied-math task (Gradient Descent, non-OOP). Even on Gradient Descent, which lies outside the Snyder corpus, \beagle still shrinks $D_{\mathrm{KL}}$ by $\geq 3\times$, raises error recurrence to $\geq 79\%$, and improves all three perceptual scores. Because the semi-Markov is applied as-is, this argues the transferred signal is general novice dynamics, not task-specific physics structure. \textit{Cross-backbone} (Fig.~\ref{fig:cross_model_viz}, App.~\ref{app:cross_model_full}): every one of six frontier backbones clears the strongest baseline on $D_{\mathrm{KL}}$, Debug, Lang, and Realism; on $P_{\mathrm{recur}}$, two exceed and four sit within $\sim 1.5\%$.

\begin{figure}[t]
\centering
\hspace*{-0.07\linewidth}%
\begin{tikzpicture}
\begin{groupplot}[
    group style={group size=5 by 1, horizontal sep=5pt},
    xbar,
    /pgf/bar width=6pt,
    width=0.275\textwidth,
    height=3.0cm,
    enlarge y limits=0.10,
    xmin=0,
    title style={font=\footnotesize, yshift=-6pt},
    tick label style={font=\scriptsize},
    label style={font=\scriptsize},
    symbolic y coords={Claude,GPT41m,GPT4om,Gem3,Gem25,Gem20},
    yticklabels={Claude H4.5,GPT-4.1m,GPT-4om,Gemini 3F,Gemini 2.5F,Gemini 2.0F},
    ytick=data,
    yticklabel style={font=\scriptsize},
    xmajorgrids, grid style={dashed, gray!20},
    ytick style={draw=none},
]

\nextgroupplot[title={$D_{\mathrm{KL}}\downarrow$}, xmax=0.85]
\addplot[fill=gray!55, draw=black!60, line width=0.4pt,
    error bars/.cd, x dir=both, x explicit,
    error bar style={black!55, line width=0.4pt},
    error mark options={rotate=90, mark size=1.2pt, line width=0.4pt}]
coordinates {
  (0.41,Claude) +- (0.09,0)
  (0.36,GPT41m) +- (0.07,0)
  (0.18,GPT4om) +- (0.04,0)
  (0.37,Gem3) +- (0.07,0)
  (0.35,Gem25) +- (0.06,0)
  (0.31,Gem20) +- (0.05,0)
};
\draw[red!60!black, dashed, line width=0.6pt] (axis cs:0.63,Gem20) -- (axis cs:0.63,Claude);

\nextgroupplot[title={$P_{\mathrm{recur}}\uparrow$ (\%)}, xmax=120, yticklabel=\empty]
\addplot[fill=gray!55, draw=black!60, line width=0.4pt,
    error bars/.cd, x dir=both, x explicit,
    error bar style={black!55, line width=0.4pt},
    error mark options={rotate=90, mark size=1.2pt, line width=0.4pt}]
coordinates {
  (73.8,Claude) +- (34.0,0)
  (73.5,GPT41m) +- (22.9,0)
  (73.0,GPT4om) +- (23.0,0)
  (65.8,Gem3) +- (23.9,0)
  (79.4,Gem25) +- (23.0,0)
  (86.2,Gem20) +- (21.3,0)
};
\draw[red!60!black, dashed, line width=0.6pt] (axis cs:75,Gem20) -- (axis cs:75,Claude);

\nextgroupplot[title={Debug$\uparrow$}, xmin=0.5, xmax=3.6, yticklabel=\empty]
\addplot[fill=gray!55, draw=black!60, line width=0.4pt,
    error bars/.cd, x dir=both, x explicit,
    error bar style={black!55, line width=0.4pt},
    error mark options={rotate=90, mark size=1.2pt, line width=0.4pt}]
coordinates {
  (1.70,Claude) +- (0.90,0)
  (2.32,GPT41m) +- (0.86,0)
  (1.82,GPT4om) +- (0.97,0)
  (2.46,Gem3) +- (0.85,0)
  (2.66,Gem25) +- (0.71,0)
  (2.46,Gem20) +- (0.85,0)
};
\draw[red!60!black, dashed, line width=0.6pt] (axis cs:1.42,Gem20) -- (axis cs:1.42,Claude);

\nextgroupplot[title={Lang$\uparrow$}, xmin=0.5, xmax=3.6, yticklabel=\empty]
\addplot[fill=gray!55, draw=black!60, line width=0.4pt,
    error bars/.cd, x dir=both, x explicit,
    error bar style={black!55, line width=0.4pt},
    error mark options={rotate=90, mark size=1.2pt, line width=0.4pt}]
coordinates {
  (2.10,Claude) +- (0.64,0)
  (2.26,GPT41m) +- (0.80,0)
  (2.52,GPT4om) +- (0.61,0)
  (2.74,Gem3) +- (0.48,0)
  (2.82,Gem25) +- (0.43,0)
  (2.62,Gem20) +- (0.66,0)
};
\draw[red!60!black, dashed, line width=0.6pt] (axis cs:2.30,Gem20) -- (axis cs:2.30,Claude);

\nextgroupplot[title={Realism$\uparrow$}, xmin=0.5, xmax=3.6, yticklabel=\empty]
\addplot[fill=gray!55, draw=black!60, line width=0.4pt,
    error bars/.cd, x dir=both, x explicit,
    error bar style={black!55, line width=0.4pt},
    error mark options={rotate=90, mark size=1.2pt, line width=0.4pt}]
coordinates {
  (1.95,Claude) +- (0.92,0)
  (2.20,GPT41m) +- (0.94,0)
  (1.90,GPT4om) +- (0.94,0)
  (2.48,Gem3) +- (0.85,0)
  (2.77,Gem25) +- (0.62,0)
  (2.44,Gem20) +- (0.85,0)
};
\draw[red!60!black, dashed, line width=0.6pt] (axis cs:1.64,Gem20) -- (axis cs:1.64,Claude);

\end{groupplot}
\end{tikzpicture}
\caption{\textbf{Cross-model generalization on Particle Simulator} ($N=50$, $T=30$). Bars show \beagle's fidelity scores for each backbone across the same five metrics as Fig.~\ref{fig:cross_task_viz}. \textcolor{red!60!black}{\itshape Red dashed line}: strongest non-\beagle LLM baseline (FewShot+$\mathcal{M}$ on Gemini 2.0 Flash). On behavioral and perceptual metrics ($D_{\mathrm{KL}}$, Debug, Lang, Realism), every backbone clears baseline; on $P_{\mathrm{recur}}$, two exceed and the remaining four sit within $\sim$1.5\% (the baseline itself has $\pm$40\% variance). Complete results in App.~\ref{app:cross_model_full}.}
\label{fig:cross_model_viz}
\end{figure}

\begin{figure}[t]

\centering
\begin{minipage}[t]{0.465\linewidth}
    \begin{tcolorbox}[
        title=\textbf{Without EFI} (math library available),
        colback=green!5!white,
        colframe=green!50!black,
        fonttitle=\scriptsize,
        width=\linewidth,
        boxsep=1pt,           
        left=8pt, right=4pt,  
        top=-2mm, bottom=-2mm,
    ]
    \begin{lstlisting}[language=Python, basicstyle=\scriptsize\fontfamily{pcr}\selectfont, breaklines=true, xleftmargin=-3mm]
theta_rad = math.radians(self.angle)
a = g * (math.sin(theta_rad) - self.mu * math.cos(theta_rad))
    \end{lstlisting}
    \end{tcolorbox}
\end{minipage}
\begin{minipage}[t]{0.465\linewidth}
    \begin{tcolorbox}[
        title=\textbf{With EFI} (math library blocked $\rightarrow$ manual approx),
        colback=red!5!white,
        colframe=red!50!black,
        fonttitle=\scriptsize,
        width=\linewidth,
        boxsep=1pt,           
        left=8pt, right=4pt,
        top=-2mm, bottom=-2mm,
    ]
    \begin{lstlisting}[language=Python, basicstyle=\scriptsize\fontfamily{pcr}\selectfont, breaklines=true, xleftmargin=-3mm]
def sin_approx(a): return a / 90.0          # improvised
def cos_approx(a): return 1.0 - a / 90.0
a = g * (sin_approx(self.angle) - self.mu * cos_approx(self.angle))
    \end{lstlisting}
    \end{tcolorbox}
\end{minipage}
\caption{\textbf{EFI forces authentic knowledge gaps.} Without EFI, the agent trivially uses \texttt{math.radians()}. With EFI, the agent improvises (incorrect) manual approximations. Full comparison in App. Fig.~\ref{fig:efi_qualitative_complete}.}
\label{fig:efi_qualitative}

\end{figure}

\noindent\textbf{EFI Ablation (Qualitative).}
We also show the design feature of blocking the knowledge using EFI. As shown in Fig.~\ref{fig:efi_qualitative}, \beagle forces agents to navigate conceptual gaps when enabling EFI. When the \texttt{math} library is blocked via EFI, \phigh{} performers attempt to implement manual trigonometric approximations, while \plow{} performers guess based on expected outputs. This shows that BKT+EFI effectively forced the agent to operate in genuine ignorance, improvising alternative solutions. 

\section{Related Work}
\label{sec:related}

Prior educational data generation has focused on static artifacts~\cite{wang2017deep,chen2021evaluating} rather than longitudinal trajectories. Computational student models span symbolic (e.g., SimStudent~\cite{matsuda2007predicting}, grounded in ACT-R), neural (Deep Knowledge Tracing~\cite{liu2019ekt}, which predicts correctness but not granular actions), and hybrid RL-LLM~\cite{radmehr2025pharmasimtext} approaches; recent LLM-based simulators include Generative Agents~\cite{park2023generative}, Generative Students~\cite{lu2024generative}, LLM-SS~\cite{nguyen2023large}, EduAgent~\cite{xu2024eduagent}, and CoderAgent~\cite{zhan2025coderagent}, targeting competency bias~\cite{qu2024recursive} via memory modules~\cite{gao2025agent4edu} or iterative reflection~\cite{xu2025classroom} (theoretical groundings and feature comparison in App.~Table~\ref{tab:comparison}). All these rely on the LLM's internal reasoning to respect knowledge boundaries, leaving them vulnerable to leakage that interpretable knowledge tracing~\cite{li2025priority,berthon2025language} does not resolve. Only \beagle jointly enforces behavioral dynamics and knowledge constraints. We ground our approach in Winne and Hadwin's SRL framework~\cite{winne1998studying}, with transitions quantified via recent work~\cite{li2025turning} and operationalized through data-driven semi-Markov models~\cite{faucon2016semi,geigle2017modeling} that extend previous work on Markov modeling~\cite{snyder2024analyzing} and Hidden Markov modeling~\cite{kinnebrew2013investigating}.

\section{Conclusion}
\label{sec:conclusion}

We presented \beagle, a neuro-symbolic framework that overcomes \textit{competency bias} via architectural constraints (semi-Markov control, BKT with EFI, and a decoupled Strategist/Executor) rather than prompting. To our knowledge, it is the first system to simultaneously achieve behavioral, epistemic, and perceptual fidelity: $D_{\mathrm{KL}}{=}0.31$ (best baseline 0.53), error recurrence 86.2\%, and traces statistically indistinguishable from real students ($N{=}71$, $d'{=}0.15$, $p_{\text{TOST}}{=}0.038$). Limitations point directly to future work: residual competency bias on the strongest backbones (Fig.~\ref{fig:bkt_strategy_5panel_realref}) motivates capability-adaptive symbolic-control gain; the knowledge--fluency trade-off (Table~\ref{tab:ablation}) calls for richer mastery representations; and \beagle's single-agent scope opens the door to multi-agent classrooms and adaptive tutoring policies trained on synthetic trajectories.


\bibliographystyle{plainnat}
\bibliography{reference}

\appendix
\input{_appendix}




\end{document}

%% file: _appendix.tex
\clearpage
\newpage

{
\footnotesize
\setlength{\tabcolsep}{4pt}
\renewcommand{\arraystretch}{0.95}
\begin{longtable}{@{}l l l@{}}
\caption{Complete notation reference.}\label{tab:notation}\\
        \toprule
        Symbol & Domain & Description \\
        \midrule
        \endfirsthead
        \multicolumn{3}{@{}l}{\textit{(Table~\ref{tab:notation} continued)}}\\
        \toprule
        Symbol & Domain & Description \\
        \midrule
        \endhead
        \midrule
        \multicolumn{3}{r@{}}{\textit{Continued on next page}}\\
        \endfoot
        \bottomrule
        \endlastfoot
        \multicolumn{3}{@{}l}{\textit{Semantic Domains}} \\
        $\mathcal{S}_{\text{text}}$ & Universe & Set of all possible natural language strings \\
        $\mathcal{S}_{\text{code}}$ & Universe & Set of all possible executable source codes \\
        $\mathcal{A}$ & $\mathcal{S}_{\text{code}} \times \mathcal{S}_{\text{text}}$ & Action space (code, utterance) \\
        \midrule
        \multicolumn{3}{@{}l}{\textit{Static Configuration $\Phi$}} \\
        $\Phi$ & $(\mathcal{P}, \mathcal{K}, \rho)$ & Full static configuration tuple \\
        $\mathcal{P}$ & $\in \mathcal{S}_{\text{text}}$ & Problem description \\
        $\mathcal{K}$ & Finite set & Set of tracked knowledge components \\
        $k$ & $\in \mathcal{K}$ & Individual knowledge component \\
        \midrule
        \multicolumn{3}{@{}l}{\textit{Student Profile $\rho$}} \\
        $\rho$ & $(\rho_{\text{behav}}, \rho_{\text{persona}})$ & Student profile tuple \\
        $\rho_{\text{behav}}$ & $\in \{\phigh, \plow\}$ & Behavioral parameter (governs dynamics) \\
        $\rho_{\text{persona}}$ & $\in \{\phigh, \plow\}$ & Linguistic persona (governs style) \\
        \midrule
        \multicolumn{3}{@{}l}{\textit{Behavior Spaces}} \\
        $\mathcal{M}$ & Finite set & Metacognitive: \planning, \enacting, \monitoring, \reflecting \\
        $\mathcal{C}$ & Finite set & Cognitive: \constructing, \debugging, \assessing \\
        \midrule
        \multicolumn{3}{@{}l}{\textit{Indexing \& Timing}} \\
        $t$ & $\in \{1, \ldots, T\}$ & Cognitive behavior step index \\
        $T$ & $\in \mathbb{N}^+$ & Total trajectory length (steps) \\
        $n$ & $\in \mathbb{N}^+$ & Metacognitive segment index \\
        $n(t)$ & $\mathbb{N}^+ \to \mathbb{N}^+$ & Maps step $t$ to its metacognitive segment \\
        $i$ & $\in \{1, \ldots, D_n\}$ & Index within metacognitive segment \\
        \midrule
        \multicolumn{3}{@{}l}{\textit{Latent State Variables}} \\
        $S_t$ & $\in \mathcal{M} \times \mathcal{C}$ & Latent behavioral state $(M_{n(t)}, C_t)$ \\
        $M_n$ & $\in \mathcal{M}$ & Metacognitive behavior for segment $n$ \\
        $M_{n(t)}$ & $\in \mathcal{M}$ & Metacognitive behavior governing step $t$ \\
        $C_t$ & $\in \mathcal{C}$ & Cognitive behavior at step $t$ \\
        $C_n^{(i)}$ & $\in \mathcal{C}$ & $i$-th cognitive behavior in segment $n$ \\
        $\mathbf{C}_n$ & Sequence & Cognitive sequence $(C_n^{(1)}, \ldots, C_n^{(D_n)})$ \\
        $D_n$ & $\in \mathbb{N}^+$ & Duration of metacognitive segment $n$ \\
        $x_t$ & $\in [0, 1]$ & Task progress at step $t$ \\
        \midrule
        \multicolumn{3}{@{}l}{\textit{Knowledge State (BKT)}} \\
        $P(L_k^{(t)})$ & $\in [0, 1]$ & Mastery probability for KC $k$ at step $t$ \\
        $\mathbf{L}_t$ & $\in [0,1]^{|\mathcal{K}|}$ & Vector of all mastery probabilities at $t$ \\
        $\kappa_t$ & $\in \mathcal{S}_\text{text}$ & Natural language constraints at $t$ \\
        $P(S)$ & $\in [0, 1]$ & BKT slip probability \\
        $P(G)$ & $\in [0, 1]$ & BKT guess probability \\
        $P(T)$ & $\in [0, 1]$ & BKT learning/transition probability \\
        \midrule
        \multicolumn{3}{@{}l}{\textit{Actions \& Observations}} \\
        $a_t$ & $= (c_t, u_t) \in \mathcal{A}$ & Action at step $t$ \\
        $c_t$ & $\in \mathcal{S}_{\text{code}}$ & Code snapshot at step $t$ \\
        $u_t$ & $\in \mathcal{S}_{\text{text}}$ & Think-aloud utterance at step $t$ \\
        $o_t$ & -- & Raw environment observation at step $t$ \\
        $\tilde{o}_t$ & -- & Filtered observation (after $\mathcal{F}$) \\
        \midrule
        \multicolumn{3}{@{}l}{\textit{Intervention}} \\
        $\mathbf{I}$ & $\{\mathcal{I}_t\}_{t=1}^T$ & Full intervention sequence \\
        $\mathcal{I}_t$ & $\subseteq \mathcal{S}_{\text{text}}$ & Optional tutor feedback at step $t$ \\
        \midrule
        \multicolumn{3}{@{}l}{\textit{Neural Action Module}} \\
        $\Omega_t$ & Tuple & Shared context $(c_{t-1}, \tilde{o}_{t-1}, \kappa_t, \mathcal{I}_t)$ \\
        $\Gamma_{\text{Strat}, n}$ & Buffer & Strategist memory at segment $n$ \\
        $\Gamma_{\text{Exec}, t}$ & Buffer & Executor memory at step $t$ \\
        $g_n$ & $\in \mathcal{S}_{\text{text}}$ & Goal for segment $n$ (Strategist output) \\
        $m_n$ & $\in \mathcal{S}_{\text{text}}$ & Mindset for segment $n$ (Strategist output) \\
        $d_n$ & $\in \mathcal{S}_{\text{text}}$ & Directive for segment $n$ (Strategist output) \\
        $f_{\text{Strat}}$ & Function & Strategist mapping to $(g_n, m_n, d_n)$ \\
        $f_{\text{Exec}}$ & Function & Executor mapping to $a_t$ \\
        \midrule
        \multicolumn{3}{@{}l}{\textit{Semi-Markov Dynamics}} \\
        $P(M_n | M_{n-1})$ & $\in [0,1]$ & Metacognitive transition probability \\
        $P(D_n | M_n)$ & Distribution & Duration distribution (Gamma) \\
        $P(C_n^{(i)} | M_n, C_n^{(i-1)})$ & $\in [0,1]$ & Cognitive transition within segment \\
        $\mathcal{F}$ & Function & Observation filter: $(o_t, M_{n(t)}) \mapsto \tilde{o}_t$ \\
        \midrule
        \multicolumn{3}{@{}l}{\textit{Stochastic Interrupts}} \\
        $P_i(x_t)$ & $\in [0,1]$ & Interrupt probability at progress $x$ \\
        $r_{\text{peak}}$ & $\in [0,1]$ & Peak interrupt probability \\
        $\mu_i$ & $\in [0,1]$ & Gaussian mean for interrupt $i$ \\
        $\sigma_i$ & $\in \mathbb{R}^+$ & Gaussian std.\ dev.\ for interrupt $i$ \\
        \midrule
        \multicolumn{3}{@{}l}{\textit{Fidelity Objectives \& Trajectory Output}} \\
        $P_{\text{real}}(C'|C)$ & Distribution & Ground-truth cognitive transitions \\
        $P_{\theta}(C'|C)$ & Distribution & Simulated cognitive transitions \\
        $D_{\text{KL}}(\cdot \| \cdot)$ & $\mathbb{R}^+$ & KL divergence \\
        $\mathcal{J}(\tau)$ & $\in \mathbb{R}$ & Evaluator/realism scoring function \\
        $\tau$ & Sequence & Trajectory $\{(M_{n(t)}, C_t, a_t, o_t)\}_{t=1}^{T}$ \\
        $e$ & -- & Error instance \\
\end{longtable}
}

\clearpage
\newpage

\section{Ground Truth Data Analysis}
\label{sec:gt_data_analysis}

Before training the semi-Markov model, we conducted systematic analyses of the Snyder corpus~\cite{snyder2024analyzing} to address two methodological challenges: (1) inflated self-transition probabilities from fine-grained logging, and (2) duration distributions that violate standard Markov assumptions.

\subsection{Temporal Aggregation}

The dataset records cognitive actions approximately every 6 seconds. When a student debugs for 2 minutes, this generates $\approx$20 consecutive \debugging entries, artificially inflating $P(\debugging \to \debugging)$ to $>$89\%. To recover meaningful transition dynamics, we analyzed temporal gaps between consecutive actions ($N_{\text{same}}=1095$ same-action pairs, $N_{\text{diff}}=139$ different-action pairs).

\begin{figure}[t]
    \centering
    \begin{tikzpicture}
        \begin{axis}[
            ybar,
            width=0.65\columnwidth,
            height=4.5cm,
            ylabel={Count},
            xlabel={Gap (sec)},
            xtick={0,1,2,3,4},
            xticklabels={0--5, 5--10, 10--30, 30--60, 60+},
            xticklabel style={font=\small},
            ymin=0,
            ymax=700, 
            bar width=12pt,
            legend style={at={(0.98,0.98)}, anchor=north east, font=\footnotesize},
            ymajorgrids=true,
            grid style={dashed,gray!30},
            enlarge x limits=0.15,
            nodes near coords,
            every node near coord/.append style={font=\tiny},
        ]
        
        \addplot[fill=blue!60] coordinates {
            (0, 562) (1, 220) (2, 230) (3, 54) (4, 27)
        };
        
        \addplot[fill=red!50] coordinates {
            (0, 62) (1, 25) (2, 32) (3, 12) (4, 8)
        };

        \draw[dashed, orange!80!black, very thick] (axis cs:2.5, 0) -- (axis cs:2.5, 650) 
            node[at end, left, font=\footnotesize, xshift=-2pt] {$\lambda=30$s};


        \legend{Same action, Diff action}
        \end{axis}
    \end{tikzpicture}
    \caption{Distribution of temporal gaps between consecutive actions. The 30-second threshold (dashed) captures 93\% of same-action sequences as single episodes.}
    \label{fig:time_gap_dist}
\end{figure}

Figure~\ref{fig:time_gap_dist} shows the gap distribution by time bucket. The ratio of same-action to different-action gaps remains $\approx$7--9:1 for short intervals, confirming that logging frequency is independent of action type. The 90th percentile of same-action gaps is 23 seconds; we select $\lambda = 30$s as a conservative threshold for episode aggregation.

Table~\ref{tab:aggregated_dynamics} presents the resulting cognitive transition probabilities after time-based aggregation, stratified by metacognitive behavior and performance level.

\begin{table}[ht]
    \centering
    \caption{Key cognitive transitions $P(C_{t+1} \mid M_t, C_t)$ after 30-second aggregation. Bold indicates notable HIGH/LOW differences.}
    \label{tab:aggregated_dynamics}
    \footnotesize
    \begin{tabular}{@{}l l c c@{}}
        \toprule
        $M_t$ & Transition & LOW & HIGH \\
        \midrule
        \multirow{2}{*}{\enacting}
            & start $\to$ \debugging & 67.6\% & 55.0\% \\
            & start $\to$ \constructing & 32.4\% & 40.0\% \\
        \midrule
        \multirow{2}{*}{\monitoring}
            & $\debugging \to \debugging$ & \textbf{86.7\%} & \textbf{40.0\%} \\
            & $\debugging \to \constructing$ & 13.3\% & 60.0\% \\
        \midrule
        \multirow{2}{*}{\planning}
            & start $\to$ \constructing & 34.4\% & \textbf{58.3\%} \\
            & start $\to$ \debugging & \textbf{62.5\%} & 36.1\% \\
        \midrule
        \reflecting
            & $\debugging \to \debugging$ & 57.1\% & 36.4\% \\
        \bottomrule
    \end{tabular}
\end{table}

\begin{figure}[ht]
    \centering
        \begin{tikzpicture}[
            node distance=2.2cm,
            state/.style={circle, draw, thick, minimum size=1.4cm, font=\scriptsize, align=center},
            every edge/.style={draw, ->, >=stealth, thick},
        ]
            \node[font=\small\bfseries] at (-2.2, 2.2) {LOW Performer};
            
            \node[state, fill=blue!25] (D1) at (-3.2, 0.5) {Debug};
            \node[state, fill=green!25] (C1) at (-1.2, 0.5) {Construct};
            
            \draw[->] (D1) to[out=150, in=210, looseness=4] node[left, font=\scriptsize] {87\%} (D1);
            
            \draw[->] (D1) to[bend left=25] node[above, font=\scriptsize] {13\%} (C1);
            \draw[->] (C1) to[bend left=25] node[below, font=\scriptsize] {100\%} (D1);
            
            \node[font=\scriptsize\itshape, text=red!60!black] at (-2.2, -0.8) {Trapped in debugging};
            
            \draw[dashed, gray!60, thick] (0.5, -1.2) -- (0.5, 2.5);
            
            \node[font=\small\bfseries] at (3.2, 2.2) {HIGH Performer};
            
            \node[state, fill=blue!25] (D2) at (2.2, 0.5) {Debug};
            \node[state, fill=green!25] (C2) at (4.2, 0.5) {Construct};
            
            \draw[->] (D2) to[out=150, in=210, looseness=4] node[left, font=\scriptsize] {40\%} (D2);
            
            \draw[->] (D2) to[bend left=25] node[above, font=\scriptsize] {60\%} (C2);
            \draw[->] (C2) to[bend left=25] node[below, font=\scriptsize] {71\%} (D2);
            
            \node[font=\scriptsize\itshape, text=green!50!black] at (3.2, -0.8) {Balanced pivoting};
            
            \node[font=\footnotesize, gray] at (0.5, -1.5) {\monitoring state};
        \end{tikzpicture}%
    \caption{Cognitive transition dynamics within \monitoring. LOW performers exhibit ``debugging loops'' (87\% self-loop), while HIGH performers pivot effectively (60\% escape rate).}
    \label{fig:monitoring_dynamics}
\end{figure}

\subsection{Duration Distribution Analysis}

A standard Markov chain implies \textit{\textbf{geometric duration distributions}}, which impose a fixed relationship between mean and variance. For a state with stay probability $p$ (probability of remaining in the state at the next step), the coefficient of variation (CV) is constrained to $\text{CV} = \sqrt{p} \approx \sqrt{1 - 1/\mu}$ where $\mu$ is the mean duration. This constraint prevents independent fitting of mean and variance.

\begin{figure}[t]
    \centering
    \begin{tikzpicture}
        \begin{axis}[
            ybar,
            width=0.65\columnwidth,
            height=4.5cm,
            ylabel={CV},
            xtick={0,1,2,3},
            xticklabels={P, E, M, R}, 
            xticklabel style={font=\small},
            ymin=0,
            ymax=1.8, 
            bar width=12pt,
            legend style={at={(0.98,0.98)}, anchor=north east, font=\footnotesize},
            ymajorgrids=true,
            grid style={dashed,gray!30},
            enlarge x limits=0.15,
            nodes near coords,
            every node near coord/.append style={font=\tiny, /pgf/number format/fixed, /pgf/number format/precision=2},
        ]

        \fill[gray!15] (axis cs:-0.5, 0.9) rectangle (axis cs:3.5, 1.0);
        \draw[dashed, gray!70, thick] (axis cs:-0.5, 0.95) -- (axis cs:3.5, 0.95)
            node[at end, right, font=\tiny, gray!50!black, align=left, xshift=-2pt] {Geometric\\($\approx$0.95)};

        \addplot[fill=blue!60] coordinates {
            (0, 1.21) (1, 0.69) (2, 0.79) (3, 0.77)
        };

        \addplot[fill=red!60] coordinates {
            (0, 0.80) (1, 1.35) (2, 0.58) (3, 0.63)
        };

        
        \node[above, font=\scriptsize\bfseries, red!70!black, xshift=6pt] 
            at (axis cs:1, 1.45) {+42\%};
            
        \node[below, font=\scriptsize\bfseries, red!70!black, xshift=10pt, yshift=-12pt] 
            at (axis cs:2, 1.0) {-39\%};

        \legend{High, Low}
        \end{axis}
    \end{tikzpicture}
    \caption{Actual CV of segment durations compared to geometric prediction (gray band, $\approx$0.95). Bars above the band indicate overdispersion; below indicates underdispersion. LOW \enacting is notably overdispersed (+42\%), while LOW \monitoring is underdispersed (-39\%). P (\planning), E (\enacting), M (\monitoring), R (\reflecting)}
    \label{fig:cv_deviation}
\end{figure}
Figure~\ref{fig:cv_deviation} compares actual CV values against the geometric constraint. The data reveals substantial deviations: LOW performers in \enacting show CV$=$1.35 (42\% above geometric prediction), indicating high variance in segment lengths, as students either exit quickly or become deeply stuck. Conversely, LOW \monitoring shows CV$=$0.58 (39\% below prediction), suggesting more regular, predictable durations. These patterns cannot be captured by a standard Markov chain.

We therefore adopt a \textit{\textbf{two-level semi-Markov}} architecture: (1) a Markov chain for state transitions $P(M_{t+1} \mid M_t)$, and (2) independent Gamma distributions for durations $P(D_t \mid M_t)$ as shown in Eq.~\ref{eq:factorization}. The Gamma distribution provides flexibility via its shape parameter $\alpha$: $\text{CV} = 1/\sqrt{\alpha}$, allowing independent fitting of each state's variance.

\begin{table}[ht]
    \centering
    \caption{Moment-matched Gamma duration parameters by profile and state. Shape $\alpha$ and scale are chosen so the fitted mean and CV exactly match the empirical mean and CV.}
    \label{tab:gamma_params}
    \footnotesize
    \begin{tabular}{@{}l l c c c c@{}}
        \toprule
        Profile & $M_t$ & Shape ($\alpha$) & Scale & Mean & Fitted CV \\
        \midrule
        \multirow{4}{*}{LOW}
            & \planning    & 1.56 & 4.92  & 7.69  & 0.80 \\
            & \enacting    & 0.55 & 17.81 & 9.84  & 1.35 \\
            & \monitoring  & 3.02 & 2.69  & 8.12  & 0.58 \\
            & \reflecting  & 2.53 & 1.74  & 4.40  & 0.63 \\
        \midrule
        \multirow{4}{*}{HIGH}
            & \planning    & 0.68 & 14.48 & 9.92  & 1.21 \\
            & \enacting    & 2.08 & 3.58  & 7.45  & 0.69 \\
            & \monitoring  & 1.61 & 7.82  & 12.56 & 0.79 \\
            & \reflecting  & 1.67 & 5.20  & 8.69  & 0.77 \\
        \bottomrule
    \end{tabular}
\end{table}

Table~\ref{tab:gamma_params} reports the moment-matched parameters. LOW \enacting has low shape ($\alpha=0.55$), capturing the heavy variance of students who either exit quickly or become deeply stuck; LOW \monitoring has high shape ($\alpha=3.02$), reflecting regular durations. HIGH \planning has the lowest shape ($\alpha=0.68$), modeling highly variable exploration time. These profile-specific duration dynamics enable authentic temporal patterns without artificial constraints.


\subsection{Interrupt Modeling}\label{app:interrupt}

Beyond the core metacognitive behaviors (Planning, Monitoring, Reflecting, Enacting), student behavior includes two \textbf{interrupt states} that temporally punctuate the learning flow:

\begin{enumerate}
    \item \textbf{Assistance} -- Student requests help from external resources or tutor
    \item \textbf{Off-Topic} -- Student disengages from the task temporarily
\end{enumerate}

These interrupts are \textit{not} part of the semi-Markov transition matrix. Instead, they are modeled as stochastic events conditioned on session progress, with probabilities derived from empirical data analysis \cite{snyder2024analyzing}.

\noindent\textbf{Empirical Patterns.}
Table~\ref{tab:interrupt_patterns} summarizes the key characteristics of each interrupt state.

\begin{table}[ht]
    \centering
    \caption{Interrupt state characteristics from data analysis.}
    \label{tab:interrupt_patterns}
    \footnotesize
    \begin{tabular}{@{}l c c l c l@{}}
        \toprule
        \textbf{State} & \textbf{\plow} & \textbf{\phigh} & \textbf{Peak Position} & \textbf{Self-Loop} & \textbf{Primary Trigger} \\
        \midrule
        Assistance & 11.7\% & 15.0\% & $\mu=0.5$ (mid)  & 33\% & Post-\enacting failure \\
        Off-Topic  & 9.2\%  & 3.7\%  & $\mu=0.73$ (late) & 40\% & Fatigue/disengagement \\
        \bottomrule
    \end{tabular}
\end{table}

Two counterintuitive findings emerge:
\begin{itemize}
    \item \textbf{High performers seek more help}: 15\% vs.\ 11.7\% assistance rate. This suggests proactive metacognitive regulation, recognizing confusion early and seeking clarification before becoming deeply stuck.
    \item \textbf{Low performers disengage more}: 9.2\% vs.\ 3.7\% off-topic rate. This reflects the ``giving up'' pattern where struggling students mentally exit the task, particularly after extended \enacting periods.
\end{itemize}

\noindent\textbf{Probability Model.}
We model interrupt probabilities using session-progress-conditioned Gaussian distributions:

\begin{equation}
    P(\text{interrupt} \mid x_t, \rho) = r_{\text{peak}}(\rho) \cdot \exp\left(-\frac{(x_t - \mu)^2}{2\sigma^2}\right)
    \label{eq:interrupt_prob}
\end{equation}

\noindent where $x_t \in [0,1]$ is session progress at $t$ step, $\mu$ is the peak position, $\sigma$ controls spread, and $r_{\text{peak}}(\rho)$ is the profile-specific peak rate.

\begin{figure}[t]
    \centering
    \begin{tikzpicture}
        \begin{axis}[
            width=0.95\columnwidth,
            height=5cm,
            xlabel={Session Progress $x$},
            ylabel={Probability},
            xmin=0, xmax=1,
            ymin=0, ymax=0.18,
            legend style={at={(0.5,1.05)}, anchor=south, font=\footnotesize},
            legend columns=2,
            grid=both,
            grid style={dashed,gray!30},
        ]
        \addplot[blue!70, thick, domain=0:1, samples=50] 
            {0.15 * exp(-((x-0.5)^2)/(2*0.25^2))};
        \addlegendentry{Assistance (\phigh)}
        
        \addplot[blue!70, thick, dashed, domain=0:1, samples=50] 
            {0.117 * exp(-((x-0.5)^2)/(2*0.25^2))};
        \addlegendentry{Assistance (\plow)}
        
        \addplot[red!70, thick, domain=0:1, samples=50] 
            {0.092 * exp(-((x-0.73)^2)/(2*0.20^2))};
        \addlegendentry{Off-Topic (\plow)}
        
        \addplot[red!70, thick, dashed, domain=0:1, samples=50] 
            {0.037 * exp(-((x-0.73)^2)/(2*0.20^2))};
        \addlegendentry{Off-Topic (\phigh)}
        
        \end{axis}
    \end{tikzpicture}
    \caption{Interrupt probability distributions. \assist peaks mid-session ($\mu=0.5$) while \offtopic peaks late ($\mu=0.73$). High performers (solid) request assistance more; low performers (dashed) disengage more.}
    \label{fig:interrupt_probs}
\end{figure}

Figure~\ref{fig:interrupt_probs} visualizes these distributions. The parameters are:

\begin{table}[ht]
    \small
    \centering
    \caption{Interrupt probability parameters.}
    \label{tab:interrupt_params}
    \begin{tabular}{@{}lccc@{}}
        \toprule
        \textbf{State} & $\mu$ & $\sigma$ & $r_{\text{peak}}$ \\
        \midrule
        Assistance (\phigh) & 0.50 & 0.25 & 0.150 \\
        Assistance (\plow) & 0.50 & 0.25 & 0.117 \\
        Off-Topic (\plow) & 0.73 & 0.20 & 0.092 \\
        Off-Topic (\phigh) & 0.73 & 0.20 & 0.037 \\
        \bottomrule
    \end{tabular}
\end{table}

\noindent\textbf{Architectural Integration.}
Interrupts are checked at each simulation step \textit{before} the semi-Markov state transition:

\begin{enumerate}
    \item Check termination conditions (success, max steps)
    \item Check \texttt{just\_received\_help} flag $\rightarrow$ skip interrupts
    \item Sample $P(\text{Off-Topic})$ $\rightarrow$ enter Off-Topic if triggered
    \item Sample $P(\text{Assistance})$ $\rightarrow$ enter Assistance if triggered
    \item Normal semi-Markov transition
\end{enumerate}

\offtopic is checked \textit{first} because it represents complete disengagement as a student who is truly distracted will not think to ask for help. Assistance represents partial engagement where the student is stuck but still cognitively active.

\noindent\textbf{Assistance Flow (Two-Turn Protocol).}
The Assistance state spans two simulation turns to capture realistic help-seeking behavior:

\textbf{Turn $t$ (Asking):}
\begin{itemize}
    \item Student formulates question based on current confusion
    \item Tutor generates hint (informed by BKT knowledge state)
    \item Sets \texttt{just\_received\_help = True}
    \item No code execution this turn
\end{itemize}

\textbf{Turn $t+1$ (Applying):}
\begin{itemize}
    \item Interrupt checks are skipped (flag is set)
    \item Tutor response is injected into Strategist/Executor prompts
    \item Student applies the help in their code
    \item Flag and response are cleared
\end{itemize}

This two-turn protocol prevents the unrealistic pattern of immediately utilizing help in the same cognitive step as requesting it. The tutor response influences BKT updates: successful application of a hint reinforces the associated knowledge component.

\noindent\textbf{Off-Topic Behavior.}
When Off-Topic is triggered:
\begin{itemize}
    \item An idle step is recorded with a distracted monologue
    \item No code changes or execution occur
    \item Step counter increments (wasting a turn)
    \item Control returns to normal semi-Markov flow
\end{itemize}

Extended Off-Topic periods (self-loop rate 40\%) model the ``rabbit hole'' phenomenon where distracted students have difficulty re-engaging. This contributes to the performance gap: low performers lose more productive steps to disengagement.

\noindent\textbf{Impact on Evaluation Metrics.}
Interrupt states directly affect several fidelity metrics:
\begin{itemize}
    \item \textbf{Step count}: Off-Topic increases steps without progress
    \item \textbf{Performance gap}: More Off-Topic for \plow{} widens the gap
    \item \textbf{Assistance rate}: Observable in behavioral traces
    \item \textbf{Realism}: LLM-as-Judge explicitly rates help-seeking patterns
\end{itemize}

\subsection{Test Data Analysis}
\label{app:test_data}

To provide an independent benchmark for behavioral evaluation, we aggregate student interaction logs from two cohorts solving physics simulation problems. The first cohort ($N=10$ sessions, 411 aggregated episodes) completed block-based programming tasks involving truck motion, package delivery, and collision detection. The second cohort ($N=13$ sessions, 306 aggregated episodes) completed Python-based coding tasks spanning bouncing ball, inclined plane, particle simulator, boat crossing, and spring-mass systems. Both datasets target middle-school physics concepts, including kinematics, forces, and energy conservation.

We apply a contextual inheritance mapping strategy to recover cognitive behaviors from low-level action logs. Code-editing actions inherit their cognitive context from prior execution events: edits following a test execution within 30 seconds are classified as \debugging{}, while edits after longer gaps are classified as \constructing{}. This approach captures the insight that running code typically triggers a debugging episode, consistent with patterns observed in prior work~\cite{snyder2024analyzing}. Consecutive same-state actions within 30 seconds are aggregated into single episodes to avoid artificial inflation of transition probabilities.

The combined test dataset yields a cognitive distribution of 54.4\% \constructing{} and 45.6\% \debugging{} with 54.1\% debug-to-debug stickiness. This balanced distribution provides a realistic benchmark for evaluating whether simulated students exhibit human-like cognitive behavior dynamics, rather than the competence-biased behavior typical of LLM baselines (Table~\ref{tab:main_results}).

\section{Method - Cont.}\label{appendx:method-cont}

\subsection{Bayesian Knowledge Tracing Details}\label{app:bkt_details}

BKT models latent mastery $L_k$ for each knowledge component $k \in \mathcal{K}$ as a hidden Markov model. The observation model relates latent mastery to observed performance through slip probability $P(S)$ (probability of incorrect response despite mastery) and guess probability $P(G)$ (probability of correct response without mastery):
\begin{equation}
    P(\text{correct}_k) = P(L_k)(1 - P(S)) + (1 - P(L_k))P(G)
\end{equation}

At each \reflecting\ or \monitoring\ step, the IDE Oracle (App.~\ref{app:problems}) returns a per-KC verdict $o_k \in \{\text{correct}, \text{incorrect}\}$ derived from AST-level grading of the agent's submitted code. Mastery beliefs for each relevant KC are then updated via Bayes' rule:
\begin{align}
    P(L_k \mid \text{correct}) &= \frac{P(L_k)(1 - P(S))}{P(L_k)(1 - P(S)) + (1 - P(L_k))P(G)} \\[1ex]
    P(L_k \mid \text{incorrect}) &= \frac{P(L_k) P(S)}{P(L_k) P(S) + (1 - P(L_k))(1 - P(G))}
\end{align}

Learning occurs after each update with transition probability $P(T)$, representing the probability of acquiring mastery after practice:
\begin{equation}
    P(L_k^{(t+1)}) = P(L_k^{(t)} \mid o_k) + (1 - P(L_k^{(t)} \mid o_k)) \cdot P(T)
\end{equation}

We use standard BKT parameters from prior work~\cite{corbett1994knowledge}: initial mastery $P(L_k^{(0)})=0.10$, learning rate $P(T)=0.25$, slip $P(S)=0.05$, and guess $P(G)=0.20$. Mastery is discretized into three levels for prompt injection: \textsc{Unknown} ($P(L_k) < 0.3$), \textsc{Partial} ($0.3 \leq P(L_k) < 0.7$), and \textsc{Mastered} ($P(L_k) \geq 0.7$).

\noindent\textbf{Interaction with EFI.}
When EFI is active on a KC (App.~\ref{app:efi_details}), BKT updates for that KC are \emph{frozen}: the agent has no perceptual access to the concept, so no informative observation can be generated and $P(L_k)$ remains at its prior. This freeze is what makes EFI-blocked KCs absorbing under self-practice alone; without it, deterministic-incorrect observations would drive $P(L_k)$ toward an interior fixed point ($\approx 0.27$) just below the \textsc{Unknown}/\textsc{Partial} boundary. Recovery is instead mediated by the ZPD tutor (App.~\ref{app:tutor_details}) rather than by stochastic accumulation, mirroring the pedagogical view that wholly unfamiliar concepts require scaffolded instruction.

\subsection{Explicit Flaw Injection}\label{app:efi_details}

While BKT models the probability of mastery, it does not prevent the LLM from accessing its latent knowledge of blocked concepts. Explicit Flaw Injection (EFI) enforces ``unknown unknowns'' by injecting hard constraints into prompts when a knowledge component is sampled as unmastered.

\noindent\textbf{Mechanism.}
Each knowledge component $k \in \mathcal{K}$ is associated with a natural language description identifying the concept to block. When BKT determines that a KC should be unavailable (through low mastery probability or explicit sampling), the following constraint is injected into both Strategist and Executor prompts:

\begin{quote}
\small
\texttt{CRITICAL CONSTRAINT -- You have NEVER heard of and CANNOT use: [\textit{concept}]. This concept does not exist in your knowledge. You must solve the problem WITHOUT using it. If the code requires `[\textit{concept}]', you will be stuck and confused.}
\end{quote}

\noindent\textbf{KC-to-Constraint Mapping.}
Table~\ref{tab:efi_mapping} provides representative mappings from knowledge components to their EFI constraint descriptions.

\begin{table}[ht]
    \centering
    \caption{Example EFI constraint mappings.}
    \label{tab:efi_mapping}
    \small
    \begin{tabular}{@{}ll@{}}
        \toprule
        Knowledge Component & Injected Constraint \\
        \midrule
        KC\_C2 (Math Library) & ``the math library and its functions'' \\
        KC\_P5 (Radian Conversion) & ``degree-to-radian conversion'' \\
        KC\_P9 (Euler Integration) & ``numerical integration methods'' \\
        KC\_C14 (Conditionals) & ``if/else conditional statements'' \\
        \bottomrule
    \end{tabular}
\end{table}

This mechanism forces the agent to improvise around genuine knowledge gaps rather than accessing the LLM's latent capabilities. As shown in Figure~\ref{fig:efi_qualitative_complete}, blocking the math library forces students to invent (incorrect) manual approximations, which is a behavior characteristic of authentic novices who lack awareness that standard solutions exist.

\noindent\textbf{Tutor-mediated EFI release.}
EFI models a student's prior-knowledge gap: while EFI is active on a KC, observations are deterministic-incorrect by construction and BKT updates for that KC are frozen, so self-practice alone cannot recover the skill. This captures the \textit{curse of incompetence}~\cite{ehrlinger2008unskilled}: a novice who lacks awareness of a missing concept cannot diagnose or repair it through repeated trial. Recovery is therefore tutor-mediated. When the help-seeking interrupt fires (metacognitive interrupt \assist{}) and the ZPD tutor (App.~\ref{app:tutor_details}) delivers a hint targeting the EFI-active KC, the simulator deactivates EFI on that KC and credits one positive Bayesian observation, lifting $P(L_k)$ from its frozen prior ($\approx 0.10$) to a single-update posterior ($\approx 0.51$). The student then returns to \constructing{} with the new concept available, and subsequent practice follows the standard BKT trajectory. This realises the intuition that genuinely unfamiliar concepts (e.g., the \texttt{math} library) are acquired through scaffolded interaction rather than solo iteration, and ensures EFI-blocked KCs are pedagogically rather than mechanically recoverable.

\subsection{Memory Buffer Architecture}\label{app:memory_buffers}

Both Strategist and Executor maintain independent memory buffers to ensure coherent, non-repetitive behavior.

\noindent\textbf{Sliding Window Buffers.}
Each agent maintains a buffer of size 3 storing recent outputs: $\Gamma_{\text{Strat}, n}$ holds recent $(g_n, m_n, d_n)$ tuples, while $\Gamma_{\text{Exec}, t}$ holds recent monologues $u_t$. These are injected into prompts with instructions to avoid verbatim repetition.

\noindent\textbf{Episodic Memory.}
Beyond sliding windows, the agent maintains episodic records of error-learning episodes, each containing the error pattern encountered, the realization learned, and whether a fix was applied. This anti-amnesia mechanism prevents the agent from repeatedly making identical mistakes without recognition, a key differentiator from vanilla LLMs, which treat each turn independently. When a familiar error recurs, the associated realization is surfaced in context, enabling the ``Oh right, I made this mistake before'' pattern enforced by the Rules block (App.~\ref{app:rules}).

\subsection{Tutor Intervention Protocol}\label{app:tutor_details}

When the \assist{} interrupt is triggered, \beagle invokes a tutor agent to generate contextually appropriate hints. The framework supports pluggable tutor implementations; we adopt a ZPD-calibrated tutor as our default, following established principles from intelligent tutoring systems~\cite{cohn2025theory}.

\noindent\textbf{Default Tutor: ZPD-Calibrated Scaffolding.}
The default tutor operationalizes Vygotsky's Zone of Proximal Development by calibrating the specificity of hints to the student's current knowledge state. Scaffold intensity is determined by the BKT mastery probability for the struggling knowledge component:

\begin{table}[ht]
    \centering
    \caption{ZPD-based scaffold level mapping.}
    \label{tab:scaffold_levels}
    \small
    \begin{tabular}{@{}ccl@{}}
        \toprule
        Mastery $P(L_k)$ & ZPD Zone & Scaffold Level \\
        \midrule
        $> 0.7$ & Comfort & \textsc{None} \\
        $0.5 - 0.7$ & Upper ZPD & \textsc{Minimal} \\
        $0.3 - 0.5$ & Mid ZPD & \textsc{Guiding} \\
        $< 0.3$ & Lower ZPD & \textsc{Explicit} \\
        \bottomrule
    \end{tabular}
\end{table}

\noindent\textbf{Escalation Protocol.}
The scaffold level increases by $+1$ for each consecutive failure on the same KC, up to a maximum of $+2$. This model of realistic tutoring provides increasingly direct support when students remain stuck despite initial hints.

\noindent\textbf{Knowledge State Context.}
The tutor receives a summary identifying the KC with the lowest current mastery (the primary intervention target), recent error patterns from the trajectory, and the student's current metacognitive behavior (to calibrate tone). This BKT-informed approach ensures hints target actual knowledge gaps rather than providing generic assistance.

\noindent\textbf{EFI Release as a Tutor Effect.}
When the targeted KC is currently EFI-blocked (App.~\ref{app:efi_details}), the intervention deactivates EFI on that KC and credits one positive BKT observation. The student leaves the interaction with both new conceptual access (EFI off) and a measurable belief update ($P(L_k){:}\;0.10 \to 0.51$). Releasing additional blocks requires further help-seeking events on the corresponding KCs; tutor turns are therefore consequential state changes rather than purely textual exchanges.

\input{prompts}

\section{Experimental Setup - Cont.}\label{app:setup}

\noindent\textbf{BKT Parameters.}
BKT uses standard parameters from prior work~\cite{corbett1994knowledge}: initial mastery $P(L_k^{(0)})=0.10$, learning rate $P(T)=0.25$, slip $P(S)=0.05$, guess $P(G)=0.20$. Mastery thresholds are \textsc{Unknown} ($<0.3$), \textsc{Partial} ($0.3$--$0.7$), and \textsc{Mastered} ($\geq 0.7$). These values are consistent with empirical estimates across educational domains and were not tuned to our specific task.

\subsection{Evaluation Problem Suite}\label{app:problems}
\beagle's evaluation suite consists of four Python implementation problems spanning two domains: physics (Particle Simulator, Bouncing Ball, Inclined Plane) and applied math (Gradient Descent). The three simulation tasks require a Python class with specific methods and physically accurate behavior using Euler integration; the math task is purely functional. Each problem includes 24--28 progressive unit tests.
\noindent\textbf{Particle Simulator} (24 tests, 12 KCs). Students implement a \texttt{Particle} class simulating 2D projectile motion. The constructor accepts initial position $(x, y)$, velocity $(v_x, v_y)$, and mass. The physics model combines constant gravitational acceleration ($g = 9.8$ m/s$^2$ downward) with linear drag force ($F_{drag} = -kv$, where $k = 0.1$). The \texttt{update(dt)} method must compute total force, derive acceleration via $a = F/m$, then update velocity and position using Euler integration:
\begin{align*}
v_{new} &= v_{old} + a \cdot dt \\
x_{new} &= x_{old} + v_{new} \cdot dt
\end{align*}
Additional methods return position, velocity, and kinetic energy ($KE = \frac{1}{2}mv^2$). This problem exercises vector decomposition (KC\_P1), numerical integration (KC\_P9), force-acceleration relationships (KC\_P10), and class structure (KC\_C9--C12).

\noindent\textbf{Inclined Plane Slider} (24 tests, 14 KCs). Students implement a \texttt{Slider} class modeling motion on a frictionful ramp. The constructor accepts angle (in degrees), ramp length, and friction coefficient $\mu$. Students must decompose gravity into components: the parallel component $g\sin\theta$ accelerates the object down the ramp, while the normal component $g\cos\theta$ determines the friction force. Net acceleration is:
$$a = g(\sin\theta - \mu\cos\theta)$$
If friction dominates ($a < 0$), the object does not slide. The class requires degree-to-radian conversion via \texttt{math.radians()}, making this problem suitable for explicit flaw-injection experiments; blocking the math library forces students to derive the conversion manually. Required methods include position, velocity, acceleration, and a \texttt{has\_reached\_bottom()} check.

\noindent\textbf{Bouncing Ball} (24 tests, 12 KCs). Students implement a \texttt{BouncingBall} class simulating vertical motion with energy-dissipating collisions. The constructor accepts initial height, velocity, and coefficient of restitution $e$ (default 0.8). On each floor collision (height $\leq 0$), velocity reverses and scales: $v_{new} = -e \cdot v_{old}$. Students must track cumulative bounce count and implement an \texttt{is\_at\_rest()} method that returns \texttt{True} when post-bounce maximum height falls below threshold (0.01m). The problem emphasizes conditional state transitions and tests both kinetic ($KE = \frac{1}{2}mv^2$) and potential ($PE = mgh$) energy calculations.

\noindent\textbf{Gradient Descent} (28 tests, 11 KCs). Students implement three functions in dependency order: (i) the objective $f(x) = x^4 - x^2 + 0.25x$ (which has a global minimum near $x \approx -0.76$); (ii) a numerical gradient $f'(x)$ computed via the central-difference formula
$$f'(x) \approx \frac{f(x + \delta) - f(x - \delta)}{2\delta}, \quad \delta = 10^{-4};$$
and (iii) the optimizer \texttt{gradient\_descent(f\_grad, x0, eta)}, which iterates $x \leftarrow x - \eta f'(x)$ until $|f'(x)| < 10^{-3}$ or a maximum-iteration cap is reached. Unlike the three physics simulators, this problem is purely functional (no class scaffold), making it the only task that exercises the full math-domain KC bundle — derivative concept (KC\_M1), numerical differentiation (KC\_M2), iterative algorithms (KC\_M3), convergence criteria (KC\_M4), and learning-rate semantics (KC\_M5) — together with the looping/conditional/function-call subset of coding KCs (KC\_C13--C16). Its functional-only structure also makes it the cleanest test of whether \beagle's class-centric scaffolding generalizes to non-OOP problems. The objective $f(x){=}x^4{-}x^2{+}x/4$ matches the function used in Paassen's Bielefeld student dataset~\cite{paassen2019python} ($N{=}15$ undergraduates, keystroke-level traces), grounding this OOD evaluation in published prior work on real-student behavior on the same problem.

\subsection{Knowledge Components}

Across the suite, BKT tracks 24 unique Knowledge Components (KCs) drawn from three domains; each problem invokes a 11--14 KC subset.

\noindent\textbf{Coding KCs (11 total)} cover Python programming skills shared across all five problems: function definition with return (C1), math library import (C2), arithmetic implementation (C4), class definition (C9), \texttt{\_\_init\_\_} method (C10), instance variables (C11), method definition (C12), loop construct (C13), conditional logic (C14), variable update patterns (C15), and function calls (C16).

\noindent\textbf{Physics KCs (8 total)} cover domain concepts shared by the three physics simulators: vector decomposition (P1), trigonometric application (P2), radian conversion (P5), numerical (Euler) integration (P9), force-acceleration relationship (P10), kinetic energy (P11), collision reflection (P14), and friction (P16).

\noindent\textbf{Math KCs (5 total)} are exercised by the Gradient Descent task: derivative concept (M1), numerical differentiation (M2), iterative algorithms (M3), convergence criteria (M4), and learning-rate semantics (M5).

An IDE Oracle provides fine-grained, KC-level feedback by analyzing student code via AST parsing. Rather than binary pass/fail, the oracle determines which specific KCs were correctly applied, enabling BKT to credit partial progress. For example, a student who correctly imports \texttt{math} but fails the overall test still receives positive updates for KC\_C2 (math library import), preventing the infinite loop where BKT penalizes correct partial work.
\subsection{Evaluation Metrics}\label{app:evaluation-metrics}
\noindent\textbf{Solve Rate.}
The Solve Rate measures the percentage of simulations that pass the full test suite before reaching the maximum step limit. This metric captures task completion capability.

Let $\mathcal{T}_{\text{pass}}(t) = \{t : \text{all unit tests pass at step } t\}$ denote the set of steps where the submitted code passes all tests. For trajectory $\tau_i$ with terminal step $T_i$, define the success indicator:
\begin{equation}
    \mathds{1}_{\text{solve}}(\tau_i) = \begin{cases}
        1 & \text{if } \mathcal{T}_{\text{pass}}(\tau_i) \neq \emptyset \\
        0 & \text{otherwise}
    \end{cases}
\end{equation}
The Solve Rate across $N$ simulations is:
\begin{equation}
    \text{Solve} = \frac{\sum_{i=1}^{N} \mathds{1}_{\text{solve}}(\tau_i)}{N} \times 100\%
\end{equation}
Excessively high solve rates (e.g., 100\%) suggest the simulation is too competent for realistic student modeling, as authentic novices frequently fail to complete tasks within time limits.

\noindent\textbf{Steps to Solve.}
The Steps to Solve metric captures the average number of simulation steps required to reach a passing solution. Real students exhibit variable completion times reflecting individual differences in problem-solving efficiency.

For trajectories that achieve success, let $t^*_i = \min\{t : t \in \mathcal{T}_{\text{pass}}(\tau_i)\}$ denote the first step at which all tests pass. The mean steps to solve is:
\begin{equation}
    \text{Steps} = \frac{\sum_{i : \mathds{1}_{\text{solve}}(\tau_i) = 1} t^*_i}{|\{i : \mathds{1}_{\text{solve}}(\tau_i) = 1\}|}
\end{equation}
We report both mean and standard deviation. When no trajectory solves (e.g., SimStudent at 0\% solve rate), we report $\text{Steps} = T_{\max}$ with std $0$ to indicate the simulation exhausted the step budget. Models with very low step counts are unrealistically efficient; real students require many iterations of trial and error.

\noindent\textbf{Performance Gap.}
The Performance Gap quantifies the model's ability to differentiate between skill levels by measuring the difference in solve rates between high-performer ($\rho_{\text{behav}} = \phigh$) and low-performer ($\rho_{\text{behav}} = \plow$) configurations:
\begin{equation}
    \text{Gap} = \text{Solve}_{\phigh} - \text{Solve}_{\plow}
\end{equation}
A positive gap indicates the model successfully produces distinct behavioral profiles. Prior literature on novice programming reports approximately 20--40\% performance differentiation between top and bottom quartile students~\cite{snyder2024analyzing}. A gap near 0\% suggests the model ignores the behavioral profile parameter.

\noindent\textbf{Cognitive Behavior KL Divergence.}
The Cognitive Behavior KL Divergence measures how closely the simulated distribution over cognitive behaviors matches the empirical distribution from the combined Cohn + Pilot evaluation cohorts (App.~\ref{app:test_data}).

Let $P_{\text{sim}}(c)$ denote the fraction of simulation steps in cognitive behavior $c \in \mathcal{C} = \{\constructing, \debugging, \assessing\}$, computed as:
\begin{equation}
    P_{\text{sim}}(c) = \frac{\sum_{\tau} |\{t : C_t = c\}|}{\sum_{\tau} T_\tau}
\end{equation}
Similarly, let $P_{\text{real}}(c)$ denote the corresponding distribution from real student data. The KL divergence is:
\begin{equation}
    D_{\mathrm{KL}} = \sum_{c \in \mathcal{C}} P_{\text{sim}}(c) \log \frac{P_{\text{sim}}(c)}{P_{\text{real}}(c)}
\end{equation}
Lower values indicate greater distributional fidelity. A value of 0 indicates perfect match; values above 0.5 suggest significant divergence.

\noindent\textbf{Compound Debugging KL.} The Compound Debugging KL ($D_{\text{debug}}$) combines two well-defined Bernoulli KL divergences over the \debugging\ state: \emph{stickiness} (the probability of remaining in \debugging) and \emph{prevalence} (the marginal fraction of steps spent in \debugging). Real students exhibit characteristic patterns: once in \debugging, they tend to remain there for multiple consecutive steps rather than flickering across states.

Let $p^{\text{sticky}} = P(C_{t+1}{=}\debugging \mid C_t{=}\debugging)$ denote the conditional stay-probability in \debugging\ and $p^{\text{ratio}} = P(C_t{=}\debugging)$ the marginal prevalence, computed separately for sim and real trajectories. We define:
\begin{equation}
    D_{\text{debug}} \;=\; \tfrac{1}{2}\, D_{\mathrm{KL}}\!\bigl(p^{\text{sticky}}_{\text{sim}} \,\big\|\, p^{\text{sticky}}_{\text{real}}\bigr) \;+\; \tfrac{1}{2}\, D_{\mathrm{KL}}\!\bigl(p^{\text{ratio}}_{\text{sim}} \,\big\|\, p^{\text{ratio}}_{\text{real}}\bigr)
\end{equation}
where $D_{\mathrm{KL}}(p \,\|\, q) = p \log(p/q) + (1-p)\log\bigl((1-p)/(1-q)\bigr)$ is the standard 2-bin Bernoulli KL. Both summands are well-defined; the $\tfrac{1}{2}$ weighting gives equal importance to the two \debugging\ properties (how persistent it is when entered, and how frequently it occurs). Lower values indicate the simulation produces debugging patterns closer to real students. Pure LLM baselines typically exhibit low stickiness (rapid state flickering), inflating both terms. $P_{\text{sim}}(C_t{=}\debugging)$ is floored at $0.01$ to avoid $\log 0$ for simulators that produce zero \debugging\ steps, and the stickiness term defaults to $1.0$ when fewer than five \debugging\ transitions are observed (insufficient-data penalty).

\noindent\textbf{Uncertainty Quantification for $D_{\mathrm{KL}}$ and $D_{\text{debug}}$.}
Both KL divergences are reported as point estimate $\pm$ bootstrap standard error. Following Efron's non-parametric bootstrap~\cite{efron1992bootstrap}, we resample \emph{trajectories} (not individual steps) with replacement, since trajectories are clusters of correlated steps and step-level resampling would underestimate variance. For each cell, we draw $n_{\text{boot}} = 2000$ resamples (seed 42 for reproducibility); each resample of $N$ trajectories is pooled into one empirical cognitive-behavior distribution and KL is recomputed against the reference (zero-floor of $10^{-10}$ for any unobserved \assessing\ behavior mass). The point estimate is the KL on the full trajectory set; the reported $\pm$ value is the standard deviation of the 2000 bootstrap statistics (a bootstrap standard error, not a 95\% CI). The same recipe applies to $D_{\text{debug}}$: pool transitions across resampled trajectories, recompute the compound stickiness/debug-ratio KL on each resample. Solve rate reports the binomial standard error $\sqrt{p(1-p)/N}$ since each run yields a single Bernoulli outcome; all other metrics (e.g., $P_{\text{recur}}$, LLM-judge scores) report sample standard deviation across runs.

\noindent\textbf{Nonlinearity.}
The Nonlinearity metric captures non-monotonic problem-solving patterns where students backtrack, delete code, or revisit earlier parts of their solution. Real programming is iterative; purely incremental progress is unrealistic.

Let $x_t \in [0, 1]$ denote the task progress at step $t$, measured as the fraction of unit tests passing. Define the backward progress indicator:
\begin{equation}
    \mathds{1}_{\text{back}}(t) = \begin{cases}
        1 & \text{if } x_t < x_{t-1} \\
        0 & \text{otherwise}
    \end{cases}
\end{equation}
The Nonlinearity score for a trajectory is:
\begin{equation}
    \text{Nonlin} = \frac{1}{T-1} \sum_{t=2}^{T} \mathds{1}_{\text{back}}(t)
\end{equation}
We report the mean across all trajectories. Higher values indicate more realistic revision behavior; pure LLM baselines typically score near zero as they monotonically improve code quality.
\noindent\textbf{Error Recurrence Rate, $P_{\text{recur}}$.}
The Error Recurrence Rate quantifies epistemic consistency by measuring whether simulated students exhibit stable misconceptions rather than superficial error patterns. This metric operationalizes the \textit{\textbf{curse}} of incompetence~\cite{ehrlinger2008unskilled}: real novices not only lack skill but also the metacognitive expertise to diagnose their errors, leading them to repeatedly struggle with the same issues. In contrast, LLMs often exhibit expert-like error recognition, quickly fixing varied error types. Motivated by findings on the \textit{competence paradox}~\cite{yuan2026validstudentsimulationlarge}, in which broadly capable LLMs tasked with emulating partially knowledgeable learners exhibit unrealistic error patterns and learning dynamics, we measure error persistence across simulation runs.

Let $\mathcal{E} = \{\texttt{TypeError}, \texttt{NameError}, \texttt{AttributeError}, \allowbreak \texttt{ValueError}, \texttt{AssertionError}, \ldots\}$ denote the set of Python error types. For each simulation run $r$, let $\mathcal{S}_r$ be the sequence of steps, and define the error count function:
\begin{equation}
    n_e(r) = \sum_{t \in \mathcal{S}_r} \mathds{1}[e \in o_t]
\end{equation}
where $o_t$ is the environment observation at step $t$. An error type $e$ is \textit{recurrent} in run $r$ if $n_e(r) \geq 2$, indicating the student encountered the same error multiple times. We compute the per-run recurrence rate and then average across runs:
\begin{align}
    P_{\text{recur}}^{(r)} &= \frac{\sum_{e \in \mathcal{E}} \mathds{1}[n_e(r) \geq 2]}{\max\!\bigl(1,\; \sum_{e \in \mathcal{E}} \mathds{1}[n_e(r) \geq 1]\bigr)}, \qquad
    P_{\text{recur}} = \frac{1}{N}\sum_{r=1}^{N} P_{\text{recur}}^{(r)}
\end{align}
We report $\pm$ sample standard deviation across the $N$ runs. Higher values indicate stable misconceptions where students repeatedly struggle with the same error type.
\noindent\textbf{Error Reaction Lag.}
The Error Reaction Lag measures the temporal delay between when an error first manifests in execution output and when the agent explicitly acknowledges it in their think-aloud utterance. This metric captures the realistic cognitive process of error discovery, where students must parse output and recognize problems rather than instantaneously detecting failures.

Let $t^* = \min\{t : C_t \in \{\debugging, \assessing\} \land o_t \text{ contains error}\}$ denote the first step where execution output $o_t$ contains error indicators (e.g., \texttt{Error}, \texttt{FAILED}, \texttt{Traceback}, \texttt{Exception}). Define the acknowledgment keyword set:
\begin{equation}
\begin{aligned}
    \mathcal{A}_{\text{keywords}} = \{&\texttt{error}, \texttt{wrong}, \texttt{bug},\\
    &\texttt{fix}, \texttt{failed}, \texttt{crash},\\
    &\texttt{broken}, \texttt{issue}, ...\}
\end{aligned}
\end{equation}
Let $t_{\text{ack}} = \min\{t > t^* : u_t \text{ contains any } a \in \mathcal{A}_{\text{keywords}}\}$ be the first subsequent step where the agent acknowledges the error. The Error Reaction Lag for a single trajectory is:
\begin{equation}
    \text{Lag} = t_{\text{ack}} - t^*
\end{equation}
If no acknowledgment occurs before trajectory termination at step $T$, we set $\text{Lag} = T - t^*$. We report the mean and standard deviation across all runs. Higher values indicate more realistic struggle, as authentic students require time to recognize and interpret error output rather than immediately identifying the root cause.

\noindent\textbf{Debug Pattern Realism.}
The Debug Pattern Realism score is an LLM-as-judge evaluation assessing whether the debugging progression resembles authentic student behavior. The judge evaluates each trajectory on a 1--3 scale:
\begin{itemize}[noitemsep, topsep=2pt]
    \item 1 = Unrealistic (immediate fixes, no exploration, psychic debugging/premature diagnosis)
    \item 2 = Partially realistic (some authentic patterns, some artificial elements)
    \item 3 = Highly realistic (gradual discovery, false starts, iterative refinement)
\end{itemize}
The judge's prompt emphasizes that \emph{realistic debugging is messy} as clean, efficient debugging is a sign of expertise, not novice behavior. We report the mean score across 50 trajectories per baseline.

\noindent\textbf{Code Realism.}
The Code Realism score is an LLM-as-judge evaluation assessing whether the generated code $c_t$ resembles authentic student work rather than polished AI output:
\begin{itemize}[noitemsep, topsep=2pt]
    \item 1 = Clearly AI-generated (over-documented, perfect style, formal structure)
    \item 2 = Ambiguous (some realistic elements, some AI tells)
    \item 3 = Authentic student code (appropriate messiness, informal naming, sparse comments)
\end{itemize}
Higher scores indicate greater perceptual fidelity to struggling students, \emph{not} superior code quality.

\noindent\textbf{Language Realism.}
The Language Realism score is an LLM-as-judge evaluation assessing whether the think-aloud monologue $u_t$ resembles authentic student self-talk:
\begin{itemize}[noitemsep, topsep=2pt]
    \item 1 = Clearly synthetic (formal, complete sentences, no hedging or emotion)
    \item 2 = Partially authentic (some natural patterns, some stilted phrasing)
    \item 3 = Highly authentic (fragments, self-correction, emotional language, uncertainty markers)
\end{itemize}

\noindent\textbf{Overall Realism Score.}
The Overall Realism Score (``Likert'' in Table~\ref{tab:main_results}) is the LLM-as-judge's holistic assessment of trace authenticity, scored directly on a 1--3 scale:
\begin{itemize}[noitemsep, topsep=2pt]
    \item 1 = Simulated (AI tells present: premature diagnosis (psychic debugging), perfect style, robotic behavior)
    \item 2 = Ambiguous (mixed signals, some authentic and some suspicious elements)
    \item 3 = Realistic (authentic novice behavior: messy, emotional, non-linear problem-solving)
\end{itemize}
This is a single holistic judgment provided by the LLM judge after reviewing the entire trace. We report the mean across all evaluated trajectories:
\begin{equation}
    \text{Realism} = \frac{1}{N} \sum_{i=1}^{N} \text{realism\_score}_i
\end{equation}
Higher scores indicate greater perceptual fidelity to \emph{struggling students}, not superior output quality.


\noindent\textbf{Calibration Reference for Semi-Markov Parameters.}
The semi-Markov transition matrix and Gamma duration parameters are fit on the Snyder corpus~\cite{snyder2024analyzing}, which aggregates 227 validated cognitive transition segments from 9 student profiles annotated by instructors. While the number of source profiles is small, the density of transition data ($>20$ observations per state-pair) stabilizes the parameter estimates. This corpus is used solely for calibration; all evaluation distributions ($D_{\mathrm{KL}}$, $D_{\text{debug}}$, Performance Gap) are computed against the combined Cohn $+$ Pilot Study set (App.~\ref{app:test_data}).

\section{Evaluation Results - Cont.}

\subsection{Comparison of Simulated Student Approaches}
\label{app:related_table}

Table~\ref{tab:comparison} compares \beagle to representative prior simulated-student frameworks along three axes: \emph{Temp.} (whether the framework models temporal dynamics, e.g.\ within-session step ordering), \emph{Know.} (whether it explicitly models knowledge state), and \emph{Coupling} (whether behavior and knowledge are jointly constrained). \beagle is the only framework that couples both, grounded in Self-Regulated Learning (SRL) theory.

\begin{table}[h]
    \centering
    \caption{Comparison of simulated student approaches. Columns: \emph{Temp.} = temporal dynamics, \emph{Know.} = knowledge modeling, \emph{Coupling} = whether behavior and knowledge are jointly constrained. \beagle uniquely couples both.}
    \label{tab:comparison}
    \footnotesize
    \setlength{\tabcolsep}{6pt}
    \renewcommand{\arraystretch}{0.95}
    \begin{tabular}{@{}l c c c l@{}}
        \toprule
        Method & Temp. & Know. & Coupling & Theory \\
        \midrule
        SimStudent~\cite{matsuda2007predicting}     & \ding{55} & \ding{51} & \ding{55} & ACT-R \\
        Gen.\ Agents~\cite{park2023generative}      & \ding{55} & \ding{55} & \ding{55} & -- \\
        Gen.\ Students~\cite{lu2024generative}      & \ding{55} & \ding{51} & \ding{55} & KLI \\
        LLM-SS~\cite{nguyen2023large}               & \ding{55} & \ding{51} & \ding{55} & -- \\
        EduAgent~\cite{xu2024eduagent}              & \ding{55} & \ding{55} & \ding{55} & CogSci \\
        CoderAgent~\cite{zhan2025coderagent}        & \ding{51} & \ding{51} & \ding{55} & ACT-R \\
        \midrule
        \textbf{\beagle (ours)} & \ding{51} & \ding{51} & \ding{51} & SRL \\
        \bottomrule
    \end{tabular}
\end{table}

\subsection{Cross-Task Generalization (Full Table)}
\label{app:cross_task_full}

Table~\ref{tab:cross_task_results} provides the full per-method numbers summarized in Fig.~\ref{fig:cross_task_viz} of the main text.

\begin{table}[h]
    \centering
    \caption{Cross-topology generalization within and beyond STEM+C ($N{=}30$, $T{=}30$). \beagle's behavioral, epistemic, and perceptual fidelity gains hold across three C2STEM physics topologies and one out-of-distribution math task (Gradient Descent). Best in \textbf{bold}.}
    \label{tab:cross_task_results}
    \footnotesize
    \resizebox{\textwidth}{!}{%
    \begin{tabular}{@{}ll cccccc@{}}
        \toprule
        \textbf{Task} & \textbf{Method} & Solve$\uparrow$ & $D_{\mathrm{KL}}$$\downarrow$ & $P_{\text{Recur}}$$\uparrow$ & \llmcol{Debug$\uparrow$} & \llmcol{Lang$\uparrow$} & \llmcol{Realism$\uparrow$} \\
        \midrule
        Particle Simulator (C2STEM) & Vanilla & \textbf{100.0$\pm$0.0\%} & 3.83$\pm$0.10 & 5.0$\pm$15.0\% & 1.17$\pm$0.45 & 2.00$\pm$0.45 & 1.17$\pm$0.45 \\
         & CoT+$\mathcal{M}$ & 86.7$\pm$6.2\% & 1.38$\pm$0.16 & 40.0$\pm$49.0\% & 1.00$\pm$0.00 & 1.50$\pm$0.56 & 1.00$\pm$0.00 \\
         & CoderAgent & 63.3$\pm$8.8\% & 4.31$\pm$0.67 & 49.3$\pm$47.0\% & 1.00$\pm$0.00 & 1.73$\pm$0.81 & 1.00$\pm$0.00 \\
         & \beagle (Flash 2.5) & 50.0$\pm$9.1\% & \textbf{0.43$\pm$0.09} & \textbf{79.4$\pm$23.0\%} & \textbf{2.77$\pm$0.62} & \textbf{2.80$\pm$0.40} & \textbf{2.77$\pm$0.62} \\
        \midrule
        Bouncing Ball (C2STEM) & CoT+$\mathcal{M}$ & \textbf{80.0$\pm$7.3\%} & 3.48$\pm$0.24 & 59.5$\pm$39.7\% & 1.30$\pm$0.69 & 2.03$\pm$0.75 & 1.17$\pm$0.52 \\
         & CoderAgent & 33.3$\pm$8.6\% & 4.30$\pm$0.36 & 71.1$\pm$32.5\% & 1.13$\pm$0.50 & 1.80$\pm$0.60 & 1.07$\pm$0.36 \\
         & \beagle (Flash 2.5) & 10.0$\pm$5.5\% & \textbf{0.20$\pm$0.05} & \textbf{85.6$\pm$18.5\%} & \textbf{1.80$\pm$0.87} & \textbf{2.67$\pm$0.54} & \textbf{2.00$\pm$0.86} \\
        \midrule
        Inclined Plane (C2STEM) & CoT+$\mathcal{M}$ & \textbf{100.0$\pm$0.0\%} & 2.80$\pm$0.20 & 33.3$\pm$47.1\% & 1.07$\pm$0.36 & 2.33$\pm$0.65 & 1.07$\pm$0.36 \\
         & CoderAgent & 60.0$\pm$8.9\% & 4.44$\pm$0.36 & 75.3$\pm$34.2\% & 1.20$\pm$0.60 & 2.23$\pm$0.62 & 1.17$\pm$0.52 \\
         & \beagle (Flash 2.5) & 63.3$\pm$8.8\% & \textbf{0.33$\pm$0.07} & \textbf{81.7$\pm$21.7\%} & \textbf{1.97$\pm$0.95} & \textbf{2.50$\pm$0.62} & \textbf{2.00$\pm$0.93} \\
        \midrule
        Gradient Descent (Math, OOD) & CoT+$\mathcal{M}$ & \textbf{100.0$\pm$0.0\%} & 8.10$\pm$0.32 & 59.8$\pm$36.1\% & 1.07$\pm$0.36 & 1.37$\pm$0.48 & 1.03$\pm$0.18 \\
         & CoderAgent & 93.3$\pm$4.6\% & 6.41$\pm$0.65 & 75.9$\pm$42.8\% & 1.10$\pm$0.40 & 1.21$\pm$0.48 & 1.07$\pm$0.36 \\
         & \beagle (Flash 2.5) & 36.7$\pm$8.8\% & \textbf{0.33$\pm$0.07} & \textbf{86.4$\pm$17.0\%} & \textbf{2.00$\pm$0.93} & \textbf{2.47$\pm$0.72} & \textbf{2.03$\pm$0.87} \\
        \bottomrule
    \end{tabular}%
    }
\end{table}

\subsection{Cross-Model Generalization (Full Table)}
\label{app:cross_model_full}

Table~\ref{tab:model_generalization} reports the full per-backbone breakdown summarized in Fig.~\ref{fig:cross_model_viz} of the main text. Task performance, behavioral, epistemic, and perceptual metrics are reported for the same prompts/scaffolding across six frontier LLM backbones; only the underlying model changes.

\begin{table}[t]
    \centering
    \caption{Cross-model generalization on Particle Simulator ($N=50$, $T=30$). Rows are \beagle with the same prompts/scaffolding; only the underlying LLM backbone changes. Blue metrics are LLM-as-Judge ($\kappa_w=0.76$). $D_{\text{KL}}$, $D_{\text{debug}}$ vs.\ combined block- and Python-based test data. Best \textbf{bold}, 2nd \underline{underline}.}
    \label{tab:model_generalization}
    \tiny
    \setlength{\tabcolsep}{1.7pt}
    \centering
    \begin{tabular}{@{}l ccc ccc ccc cc c@{}}
        \toprule
        & \multicolumn{3}{c}{\textbf{Task Perf.}} & \multicolumn{3}{c}{\textbf{Behavioral}} & \multicolumn{3}{c}{\textbf{Epistemic}} & \multicolumn{2}{c}{\textbf{Perceptual}} & \textbf{Overall} \\
        \cmidrule(lr){2-4} \cmidrule(lr){5-7} \cmidrule(lr){8-10} \cmidrule(lr){11-12} \cmidrule(lr){13-13}
        \textbf{Backbone} & Solve & Steps & Gap & $D_{\mathrm{KL}}$$\downarrow$ & $D_{\text{debug}}$$\downarrow$ & Nonlin$\uparrow$ & $P_{\text{Recur}}$$\uparrow$ & Lag$\uparrow$ & \llmcol{Debug$\uparrow$} & \llmcol{Code} & \llmcol{Lang} & \llmcol{Realism$\uparrow$} \\
        \midrule
        Gemini 2.0 Flash & 10.00$\pm$4.24\% & 29$\pm$4 & +4\% & \underline{0.31$\pm$0.05} & \underline{0.02$\pm$0.01} & \underline{0.34$\pm$0.13} & \textbf{86.2$\pm$21.3\%} & \textbf{2.62$\pm$1.98} & \underline{2.46$\pm$0.85} & 2.90$\pm$0.41 & 2.62$\pm$0.66 & 2.44$\pm$0.85 \\
        Gemini 2.5 Flash & 52.00$\pm$7.07\% & 26$\pm$5 & +40\% & 0.35$\pm$0.06 & 0.03$\pm$0.01 & 0.31$\pm$0.16 & \underline{77.3$\pm$23.0\%} & \underline{2.40$\pm$1.33} & \textbf{2.66$\pm$0.71} & 2.96$\pm$0.20 & \textbf{2.82$\pm$0.43} & \textbf{2.68$\pm$0.68} \\
        Gemini 3 Flash & 88.00$\pm$4.60\% & 22$\pm$6 & +0\% & 0.37$\pm$0.07 & 0.03$\pm$0.01 & 0.26$\pm$0.12 & 65.8$\pm$23.9\% & 1.06$\pm$0.31 & \underline{2.46$\pm$0.85} & \textbf{3.00$\pm$0.00} & \underline{2.74$\pm$0.48} & \underline{2.48$\pm$0.85} \\
        GPT-4o-mini & 0.00$\pm$0.00\% & 30$\pm$0 & +0\% & \textbf{0.18$\pm$0.04} & 0.03$\pm$0.01 & \textbf{0.40$\pm$0.14} & 73.0$\pm$23.0\% & 1.80$\pm$0.87 & 1.82$\pm$0.97 & 2.78$\pm$0.41 & 2.52$\pm$0.61 & 1.90$\pm$0.94 \\
        GPT-4.1-mini & 50.00$\pm$7.07\% & 23$\pm$8 & +4\% & 0.36$\pm$0.07 & \textbf{0.02$\pm$0.00} & 0.25$\pm$0.15 & 73.5$\pm$22.9\% & 1.06$\pm$0.31 & 2.32$\pm$0.86 & \textbf{3.00$\pm$0.00} & 2.26$\pm$0.80 & 2.20$\pm$0.94 \\
        Claude Haiku 4.5 & 70.00$\pm$6.48\% & 18$\pm$9 & -4\% & 0.41$\pm$0.09 & 0.03$\pm$0.01 & 0.23$\pm$0.16 & 73.8$\pm$34.0\% & 1.61$\pm$1.12 & 1.70$\pm$0.90 & \textbf{3.00$\pm$0.00} & 2.10$\pm$0.64 & 1.70$\pm$0.92 \\
        \bottomrule
    \end{tabular}
\end{table}

\subsection{LLM-as-Judge Calibration}
\label{app:llm_judge}

To validate the LLM-as-Judge evaluator, we conducted a human calibration study following established inter-rater reliability protocols~\cite{mchugh2012interrater}. Two expert raters (CS PhD students) independently evaluated 30 \beagle traces selected via stratified sampling to ensure coverage across the LLM realism score distribution (10 low, 10 medium, 10 high). Each trace was rated on a 3-point Likert scale across four dimensions: overall realism, code quality, debugging behavior, and language realism.

\noindent\textbf{Scale Choice (1--3 vs.\ 1--5).} We initially piloted a 1--5 Likert scale matching common LLM-judge practice, but rater agreement was unstable: raters could not reliably distinguish adjacent middle levels (e.g., 2 vs.\ 3), inflating disagreement and depressing $\kappa_w$. We adopted a 1--3 trichotomy (\textit{unfaithful} / \textit{neutral} / \textit{faithful}) for the final calibration; this matches the unfaithful/neutral/faithful trichotomy used in related work and yielded substantial agreement.

Disagreements were resolved through discussion to produce consensus ratings. We then computed Quadratic Weighted Kappa ($\kappa_w$) between the consensus ratings and LLM judgments. The LLM judge achieved $\kappa_w=0.760$ for realism (substantial agreement) with an aggregate $\kappa_w=0.674$ across all dimensions. This validates the LLM-as-Judge as a reliable proxy for human perceptual evaluation of trace realism.

\subsection{RQ2 - Cont.}
\noindent\textbf{Behavioral DNA.}

\begin{figure}[ht]
    \centering
    \includegraphics[width=0.55\linewidth]{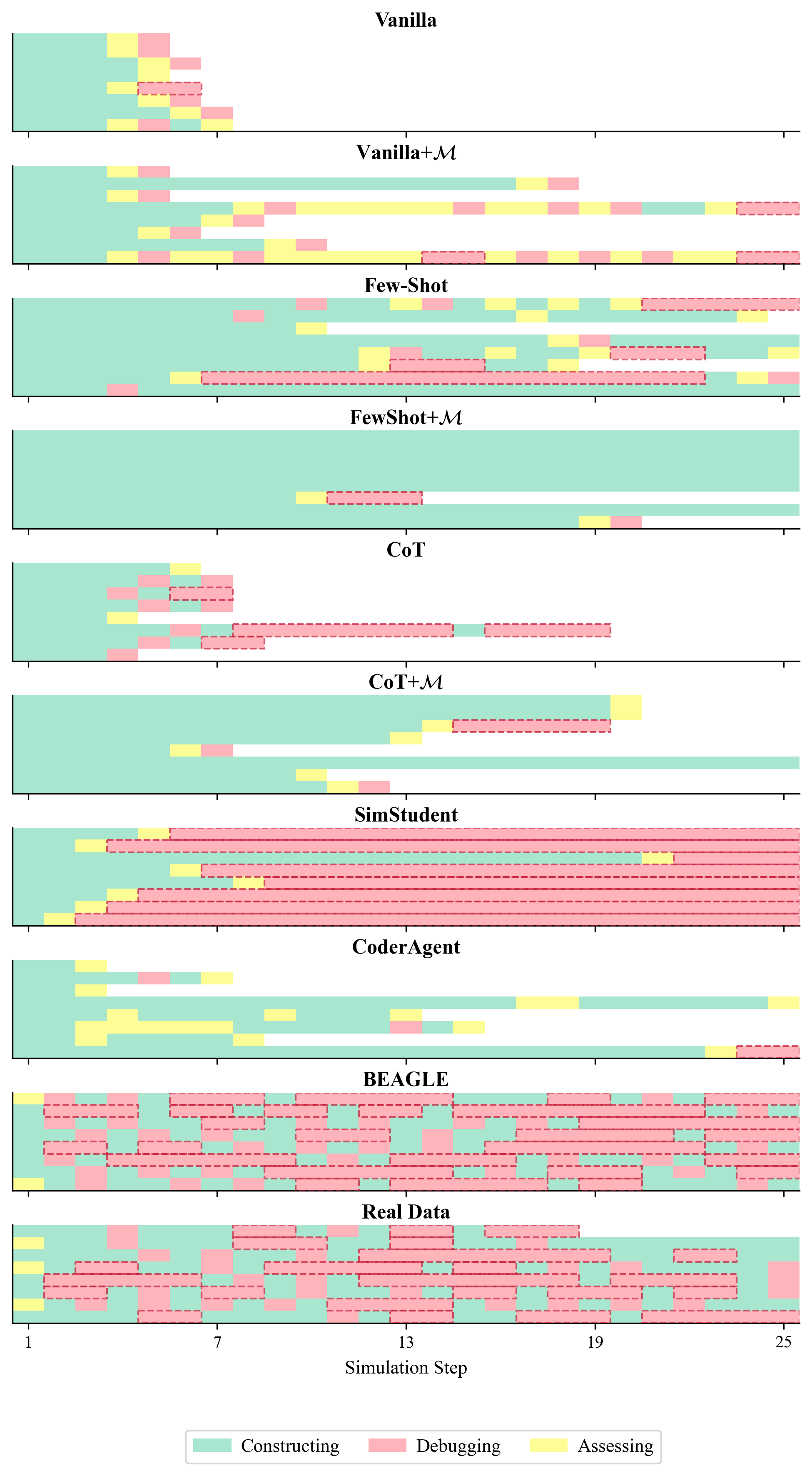}
    \caption{Extended behavioral DNA comparison across all baselines. Each row shows cognitive behavior trajectories (green=constructing, pink=debugging, yellow=assessing) for eight representative runs. Dashed boxes highlight debugging loops. Vanilla and CoT baselines exhibit rapid flickering; \beagle produces sustained blocks matching real student data.}
    \label{fig:beh-dna-complete}
\end{figure}

Figure~\ref{fig:beh-dna-complete} presents the complete behavioral DNA comparison across all ten baselines. Each row shows eight representative traces, with horizontal strips encoding the cognitive behavior at each step. The debugging loop patterns (highlighted by dashed boxes) reveal a key distinction: vanilla prompting produces rapid state flickering, while \beagle generates sustained debugging blocks that match real student behavior. Notably, the +$\mathcal{M}$ variants (which receive metacognitive behavior information) show slightly longer runs than their base counterparts, but still fall short of the contiguous blocks exhibited by \beagle and real data. CoderAgent, despite its agentic architecture, displays the same flickering pattern as vanilla approaches, confirming that agentic capability alone does not produce authentic student behavior. The white space in later steps indicates runs where the problem was solved early.

\noindent\textbf{Qualitative Evaluation on EFI.}

Figure~\ref{fig:efi_qualitative_complete} illustrates how EFI forces the agent to navigate ``unknown unknowns, so that the agent cannot simply call the missing library. Without EFI, the agent trivially uses \texttt{math.cos()} and \texttt{math.radians()}, producing correct but pedagogically uninteresting behavior. With EFI activated, the agent must improvise. The low performer, lacking both the library and the mathematical intuition, resorts to guessing based on expected output values. The high performer attempts something more creative: defining placeholder functions that approximate trigonometric behavior through linear interpolation. While mathematically incorrect, this strategy reflects authentic novice reasoning, trying to work around a knowledge gap rather than giving up. This ``unknown unknowns'' behavior is precisely what EFI enables: the agent cannot simply look up the answer because it does not know such an answer exists.

\begin{figure}[ht]
    \centering
    \begin{tcolorbox}[
        title=\textbf{Without EFI} (math library available),
        colback=green!5!white,
        colframe=green!50!black,
        fonttitle=\scriptsize,
        width=\columnwidth,
        top=-2mm, bottom=-2mm,
    ]
    \begin{lstlisting}[language=Python, basicstyle=\scriptsize\fontfamily{pcr}\selectfont, breaklines=true, xleftmargin=-3mm]
def get_acceleration(self):
    theta_rad = math.radians(self.angle)  # standard library
    a_parallel = g * math.sin(theta_rad)
    a_friction = self.mu * g * math.cos(theta_rad)
    return a_parallel - a_friction
    \end{lstlisting}
    \end{tcolorbox}

    \begin{tcolorbox}[
        title=\textbf{With EFI -- \plow{} Performer} (gives up),
        colback=orange!5!white,
        colframe=orange!60!black,
        fonttitle=\scriptsize,
        width=\columnwidth,
        top=-2mm, bottom=-2mm,
    ]
    \begin{lstlisting}[language=Python, basicstyle=\scriptsize\fontfamily{pcr}\selectfont, breaklines=true, xleftmargin=-3mm]
def get_acceleration(self):
    # The test wants ~4.9 for theta=30, mu=0; that's about g/2. Maybe just g times angle/90?
    return g * self.angle / 90.0  # trying this
    \end{lstlisting}
    \end{tcolorbox}

    \begin{tcolorbox}[
        title=\textbf{With EFI -- \phigh{} Performer} (attempts manual approximation),
        colback=red!5!white,
        colframe=red!50!black,
        fonttitle=\scriptsize,
        width=\columnwidth,
        top=-2mm, bottom=-2mm,
    ]
    \begin{lstlisting}[language=Python, basicstyle=\scriptsize\fontfamily{pcr}\selectfont, breaklines=true, xleftmargin=-3mm]
def sin_approx(angle):
    return angle / 90.0  # linear approximation, probably off for large angles
def cos_approx(angle):
    return 1.0 - angle / 90.0  # also linear, also wrong
def get_acceleration(self):
    a_parallel = g * sin_approx(self.angle)
    a_friction = self.mu * g * cos_approx(self.angle)
    return a_parallel - a_friction  # still using the placeholders
    \end{lstlisting}
    \end{tcolorbox}
    
    \caption{Navigating unknown unknowns via EFI. Without the flaw, the agent uses standard library functions. With EFI blocking the math library, a low performer guesses from expected outputs while a high performer invents placeholder approximations, as authentic workaround behavior that emerges from genuine knowledge gaps.}
    \label{fig:efi_qualitative_complete}
\end{figure}

\noindent\textbf{BKT Alignment.}

Figure~\ref{fig:bkt_adherence} validates that the symbolic knowledge state $P(L)$ meaningfully constrains observable behavior. Students initialized with low mastery probability produce lower test pass rates, while those with higher $P(L)$ solve more tests. The positive correlation ($r = 0.89$) confirms that BKT does not merely provide decorative state information, as it actively shapes the agent's behavioral trajectory. This coupling is essential for preventing the LLM's latent competence from overriding intended knowledge gaps. Without this alignment, simulated novices would behave like experts despite their nominal skill profile.

\begin{figure}[t]
    \centering
    \begin{tikzpicture}
        \begin{axis}[
            width=0.5\columnwidth,
            height=3.5cm,
            xlabel={Knowledge Probability $P(L)$},
            ylabel={Test Pass Rate (\%)},
            xmin=0, xmax=1.05,
            ymin=0, ymax=42,
            grid=both,
            grid style={dashed,gray!30},
            legend style={at={(0.02,0.98)}, anchor=north west, font=\scriptsize},
        ]
        \addplot[color=blue!60, mark=*, thick] coordinates {
            (0.10, 4.5) (0.30, 18.4) (0.50, 17.9) (0.70, 22.1) (0.90, 37.0)
        };
        \addlegendentry{Observed}
        \addplot[color=red!60, dashed, domain=0:1] {2.83 + 34.30*x};
        \addlegendentry{Trend}
        \end{axis}
    \end{tikzpicture}
    \caption{BKT-behavior alignment. Test pass rates correlate with knowledge probability $P(L)$ ($r = 0.89$), confirming that the symbolic knowledge state constrains agent performance rather than being overridden by latent LLM competence.}
    \label{fig:bkt_adherence}
\end{figure}

\subsection{Decoupling \texorpdfstring{$\rho_{\text{behav}}$}{Behavioral Profile} and \texorpdfstring{$\rho_{\text{persona}}$}{Linguistic Persona}}
\label{sec:decoupling}

To validate that behavioral profile and linguistic persona control orthogonal dimensions, we conduct a $2 \times 2$ factorial experiment ($N=110$ per condition, 440 total) crossing $\rho_{\text{behav}}$ (\plow{} vs.\ \phigh{} semi-Markov parameters) with $\rho_{\text{persona}}$ (``confused'' vs.\ ``strategic'' prompt styling).

Table~\ref{tab:behav_mediation} shows that $\rho_{\text{behav}}$ significantly affects metacognitive transitions. \plow{} profiles exhibit higher \enacting$\to$\enacting{} self-loop rates (77.5\% vs.\ 52.7\%, $p<.001$), spending 42.5\% of time trapped in impulsive action. These transition patterns mediate performance: higher \enacting{} persistence correlates with more steps ($r = 0.228$, $p<.001$) and lower solve probability ($r = -0.194$, $p<.001$). \textit{The semi-Markov controller shapes performance through behavioral dynamics, not arbitrary handicapping.}

\begin{table}[t]
    \centering
    \caption{Effect of Behavioral Model ($\rho_{\text{behav}}$) on Metacognitive Transitions. N=220 per group (440 total). Only statistically significant differences shown (Welch's $t$-test, $p<0.05$).}
    \label{tab:behav_mediation}
    \scriptsize
    \setlength{\tabcolsep}{4pt}
    \begin{tabular}{l c c c c @{\hspace{8pt}} l c c c c}
        \toprule
        \textbf{Metric} & \textbf{Low} & \textbf{High} & \textbf{Diff} & \textbf{$p$} & \textbf{Metric} & \textbf{Low} & \textbf{High} & \textbf{Diff} & \textbf{$p$} \\
        \midrule
        Enacting \%   & 42.5\% & 18.4\% & -24.1\% & \textbf{$<$.001} & Ena$\to$Ref   & 1.1\%  & 4.3\%  & +3.2\%  & \textbf{$<$.001} \\
        Ena$\to$Ena   & 77.5\% & 52.7\% & -24.8\% & \textbf{$<$.001} & Planning \%   & 37.5\% & 46.2\% & +8.7\%  & \textbf{$<$.001} \\
        Ena$\to$Pla   & 6.0\%  & 17.5\% & +11.5\% & \textbf{$<$.001} & Pla$\to$Ref   & 0.8\%  & 2.9\%  & +2.1\%  & \textbf{0.001} \\
        Monitoring \% & 14.3\% & 25.3\% & +11.0\% & \textbf{$<$.001} & Mon$\to$Ena   & 10.2\% & 5.2\%  & -5.0\%  & \textbf{0.003} \\
        Reflecting \% & 4.8\%  & 10.1\% & +5.4\%  & \textbf{$<$.001} & Ref$\to$Ena   & 14.9\% & 8.7\%  & -6.2\%  & \textbf{0.007} \\
        Pla$\to$Mon   & 4.1\%  & 11.1\% & +7.0\%  & \textbf{$<$.001} & Pla$\to$Ena   & 10.7\% & 7.2\%  & -3.6\%  & \textbf{0.011} \\
        Ref$\to$Ref   & 21.2\% & 35.1\% & +13.9\% & \textbf{$<$.001} & Mon$\to$Ref   & 7.2\%  & 4.2\%  & -2.9\%  & \textbf{0.041} \\
        Ref$\to$Mon   & 0.0\%  & 3.7\%  & +3.7\%  & \textbf{$<$.001} &               &        &        &         &                  \\
        \midrule
        \multicolumn{10}{l}{\textit{Transition $\to$ Performance Correlations (across all runs):}} \\
        \midrule
        Ena$\to$Ena vs Steps  & \multicolumn{4}{l}{$r = 0.228$, \textbf{$p<$.001}} & Ena$\to$Ena vs Solved & \multicolumn{4}{l}{$r = -0.194$, \textbf{$p<$.001}} \\
        \bottomrule
    \end{tabular}
\end{table}

Table~\ref{tab:persona_linguistic} shows that $\rho_{\text{persona}}$ independently shapes linguistic expression. \phigh{} personas use more certainty markers (7.6 vs.\ 2.1 per 1000 words, $p<.001$), while uncertainty markers remain constant ($p = 0.155$). Crucially, persona does not affect behavioral metrics: \enacting$\to$\enacting{} rates ($p = 0.076$) and solve rates ($p = 0.888$) are indistinguishable across persona conditions. \textit{Researchers can vary how students behave without changing how they sound, and vice versa.}

\begin{table}[t]
    \centering
    \caption{Effect of $\rho_{\text{persona}}$ on linguistic style (markers per 1000 words). Welch's $t$-test; Fisher's exact for solve rate.}
    \label{tab:persona_linguistic}
    \footnotesize
    \begin{tabular}{l c c c c}
        \toprule
        Metric & \plow{} Persona & \phigh{} Persona & Diff & $p$ \\
        \midrule
        Certainty Words   & 2.1   & 7.6   & +5.5  & $<$.001 \\
        Uncertainty Words & 63.6  & 61.7  & -1.8  & 0.155 \\
        \midrule
        \multicolumn{5}{l}{\textit{Persona does NOT affect behavioral metrics (confirming decoupling):}} \\
        \midrule
        \enacting$\to$\enacting{} Rate & 68.0\% & 62.2\% & -5.8\% & 0.076 \\
        Solve Rate                     & 12.7\% & 13.6\% & +0.9\% & 0.888 \\
        \bottomrule
    \end{tabular}
\end{table}

\begin{figure}[t]
    \centering
    \begin{tikzpicture}
        \begin{axis}[
            width=0.5\columnwidth,
            height=4cm,
            xlabel={\enacting$\to$\enacting{} Self-Loop Rate (\%)},
            ylabel={Simulations},
            xmin=0, xmax=100,
            ymin=0, ymax=69,
            legend style={at={(0.02,0.98)}, anchor=north west, font=\footnotesize},
            xtick={0,20,40,60,80,100},
        ]
        \addplot+[const plot, fill=red!50, fill opacity=0.6, draw=red!70, thick, mark=none] coordinates {(0.0, 21) (5.0, 21) (5.0, 0) (10.0, 0) (10.0, 0) (15.0, 0) (15.0, 0) (20.0, 0) (20.0, 0) (25.0, 0) (25.0, 1) (30.0, 1) (30.0, 1) (35.0, 1) (35.0, 0) (40.0, 0) (40.0, 2) (45.0, 2) (45.0, 0) (50.0, 0) (50.0, 4) (55.0, 4) (55.0, 1) (60.0, 1) (60.0, 3) (65.0, 3) (65.0, 8) (70.0, 8) (70.0, 4) (75.0, 4) (75.0, 11) (80.0, 11) (80.0, 34) (85.0, 34) (85.0, 41) (90.0, 41) (90.0, 50) (95.0, 50) (95.0, 39) (100.0, 39)} \closedcycle;
        \addplot+[const plot, fill=blue!50, fill opacity=0.6, draw=blue!70, thick, mark=none] coordinates {(0.0, 59) (5.0, 59) (5.0, 0) (10.0, 0) (10.0, 0) (15.0, 0) (15.0, 0) (20.0, 0) (20.0, 0) (25.0, 0) (25.0, 1) (30.0, 1) (30.0, 7) (35.0, 7) (35.0, 0) (40.0, 0) (40.0, 3) (45.0, 3) (45.0, 0) (50.0, 0) (50.0, 20) (55.0, 20) (55.0, 3) (60.0, 3) (60.0, 7) (65.0, 7) (65.0, 17) (70.0, 17) (70.0, 7) (75.0, 7) (75.0, 29) (80.0, 29) (80.0, 32) (85.0, 32) (85.0, 17) (90.0, 17) (90.0, 9) (95.0, 9) (95.0, 9) (100.0, 9)} \closedcycle;
        \legend{$\rho_{\text{behav}}$ = \plow{} ($\mu$=78\%), $\rho_{\text{behav}}$ = \phigh{} ($\mu$=53\%)}
        \draw[red!80, dashed, very thick] (axis cs:78,0) -- (axis cs:78,64);
        \draw[blue!80, dashed, very thick] (axis cs:53,0) -- (axis cs:53,64);
        \end{axis}
    \end{tikzpicture}
    \caption{Distribution of \enacting$\to$\enacting{} self-loop rates by $\rho_{\text{behav}}$. Despite significant group-level differences ($p<.001$), distributions overlap; the realized transition pattern, not the assigned profile, determines performance.}
    \label{fig:distribution_overlap}
\end{figure}

Figure~\ref{fig:distribution_overlap} reveals a mediation effect. Group assignment (\plow{} vs.\ \phigh{} Markov) does not directly predict solve rate (Welch's $t$-test, $p = 0.481$), yet the realized \enacting$\to$\enacting{} rate correlates significantly with solving ($r = -0.194$, $p<.001$; bottom row of Table~\ref{tab:behav_mediation}). The distributions overlap substantially as a \phigh{} student can exhibit high \enacting{} persistence, while a \plow{} student may escape the trap. \textit{The behavioral profile shapes the probability distribution over transitions, but the granular transition pattern determines outcomes.}

\section{Case Study: Tutor Mechanism Comparison}
\label{app:tutor_casestudy}

To investigate how different tutoring mechanisms influence simulated student performance, we conducted a controlled experiment comparing five tutor strategies under identical behavioral conditions. This case study isolates the effect of hint quality by holding constant all other simulation parameters, including the Markov-driven behavioral sequence and the timing of help-seeking events.

\noindent\textbf{Experimental Setup.}
We ran $N=10$ simulations per tutor condition using low-performing student profiles on the Particle Simulator problem. Each simulation was limited to 30 steps. To ensure fair comparison, we employed two key controls. First, we fixed the BKT initialization seed, guaranteeing that all students began with identical initial knowledge states across conditions. Second, we disabled the probabilistic assistance mechanism and instead forced help-seeking events at predetermined steps (5, 10, 15, 20, and 25), ensuring each simulation received exactly five tutor interactions at identical points in the problem-solving trajectory. This design isolates the effect of hint content from confounding factors such as when students ask for help or how often they seek assistance.

\noindent\textbf{Tutor Conditions.}
We evaluated five tutoring strategies spanning a range of sophistication. The ZPD Tutor uses BKT mastery estimates to select an appropriate scaffolding level and routes hints through different generation modes (Socratic, strategic, corrective, or specific) based on the student's current error state \cite{cohn2025theory}. The ML-based Tutor generates personalized hints informed by the student's BKT knowledge state, targeting the knowledge component with lowest estimated mastery. The Rule-based Tutor provides template-driven hints triggered by error patterns, similar to classical Hint Factory approaches. The Simple LLM Tutor uses a fixed prompt requesting encouraging, brief guidance without explicit awareness of the student's knowledge state. Finally, the No Tutor condition serves as a baseline, providing no hints when assistance is requested.

\noindent\textbf{Results.}
Table~\ref{tab:tutor_comparison} summarizes the task completion outcomes across conditions. The Simple LLM tutor achieved the highest solve rate at 50\%, followed by the ZPD tutor at 40\%. The No Tutor and Rule-based conditions each achieved 30\%, while the ML-based tutor performed lowest at 20\%. Mean steps to completion were similar across conditions (27--29 steps), suggesting that when solutions were reached, they occurred at comparable points in the problem-solving process.

\begin{table}[ht]
\footnotesize
\centering
\caption{Tutor Comparison Results ($N=10$ per condition)}
\label{tab:tutor_comparison}
\begin{tabular}{l c c c}
\toprule
\textbf{Tutor Type} & \textbf{Solved} & \textbf{Solve Rate} & \textbf{Mean Steps} \\
\midrule
Simple LLM      & 5/10 & 50\% & 28.1 \\
ZPD             & 4/10 & 40\% & 27.8 \\
None (Baseline) & 3/10 & 30\% & 27.8 \\
Rule-based      & 3/10 & 30\% & 27.2 \\
ML-based        & 2/10 & 20\% & 29.1 \\
\bottomrule
\end{tabular}
\end{table}

\noindent\textbf{Discussion.}
The unexpected performance advantage of the Simple LLM tutor warrants examination. Qualitative analysis of generated hints revealed that the ZPD tutor, upon detecting errors, routes to a ``corrective'' generation mode that produces explicit, diagnostic feedback (e.g., ``Your Particle class is missing the get\_position and update methods''). In contrast, the Simple LLM tutor, lacking error-state routing, consistently produces encouraging, Socratic-style hints (e.g., ``That's a good question! Think about how you stored the x value...''). This suggests that for simulated students, encouraging guidance that prompts reflection may be more effective than explicit error identification, a pattern consistent with findings in human tutoring literature regarding the benefits of minimal scaffolding.

However, we note important limitations. With $N=10$ per condition, the observed differences (50\% vs.\ 40\%) are not statistically significant, and results may reflect stochastic variation in LLM-generated code rather than true tutor efficacy differences. Additionally, this study examines simulated student responses to hints; human students may respond differently to the same tutoring strategies. The case study demonstrates \beagle's utility as a testbed for comparing pedagogical interventions, while acknowledging that findings should be validated with human learners before drawing strong conclusions about tutor design.

\section{Human Turing Test}
\label{app:turing_test}

To rigorously assess perceptual fidelity, we conducted a human Turing test comparing \beagle trajectories against real student data. This section details the protocol, statistical results, and qualitative findings.

\subsection{Experimental Setup}
\label{app:turing_setup}

\noindent\textbf{Data Collection Protocol.}
Our human evaluation adopts the 2-second pause \emph{capture protocol} from Paassen's Bielefeld Python Programming dataset~\cite{paassen2019python}, conducted at Bielefeld University with $N=15$ undergraduates. In that study, a code snapshot was recorded whenever a student modified code and then paused for at least two seconds before further changes; this pause-based capture reflects natural cognitive breakpoints during programming and is the established standard for keystroke-level programming-trace recording in the educational data mining literature. Notably, the Bielefeld students implemented gradient descent on $f(x){=}x^4{-}x^2{+}x/4$ across five progressive subtasks, the same function used in our out-of-distribution Gradient Descent task (App.~\ref{app:problems}). The Turing-test \textit{stimuli} are drawn from our pilot study (\S\ref{sec:experiments}, Datasets), not from Bielefeld; we adopt only the capture protocol, which sets the temporal grid that the breakdown mechanism (next paragraph) targets.

\noindent\textbf{Breakdown Mechanism.}
A key challenge arises because \beagle generates high-level code changes (complete function bodies per simulation step) rather than the granular keystroke-level modifications present in the Bielefeld data. To bridge this gap, we implement a diff-based breakdown mechanism that transforms sparse LLM outputs into dense progressions matching the temporal resolution of the Paassen 2-second pause protocol used by real-data benchmarks and downstream program-trace analyses in this literature~\cite{paassen2021ast2vec}. The breakdown is content-agnostic: it splits existing state changes into typing-paced increments using fixed heuristics, without adding reasoning, errors, or stylistic content.

Independent evidence supports this: the LLM-judge realism scores (Table~\ref{tab:main_results}) are computed from content features (cramped style, single-letter variables, emotional comments; App.~\ref{app:judge_prompt}) that are invariant to snapshot density, and the no-semi-Markov ablation $D_{\mathrm{KL}}{=}6.81$ (\S\ref{sec:ablation}) is computed at the cognitive-state-sequence level, where the breakdown does not operate. Both metrics show large baseline-vs-\beagle gaps, confirming the perceptual signal is driven by content and behavioral fidelity rather than by the breakdown's typing simulation.

The breakdown process begins by normalizing code states to remove formatting artifacts. Consecutive duplicate states that differ only by whitespace are collapsed, and blank lines between method definitions are standardized according to randomly-selected styles (always present, never present, or probabilistically present) to increase stylistic diversity across traces. A problem-specific header is prepended to establish consistent initial conditions.

The core expansion phase decomposes transitions between consecutive code states into three primitive operations: insertions, deletions, and replacements. For insertions, new lines are added one at a time to simulate the natural typing pauses observed in the Bielefeld 2-second pause study, where students typically pause after completing each line of code. For replacements, the mechanism first identifies the common prefix between old and new content, then builds up the modified portion incrementally. Within-line modifications can be interpolated at word granularity, where each word is appended sequentially while preserving leading indentation to mimic the auto-indent behavior of modern code editors. The first code state receives special handling, with each line (or optionally each word) added progressively to simulate the initial drafting process.

A configurable \texttt{level} parameter ranging from $0.0$ to $1.0$ controls the retention probability of intermediate states. At level $1.0$, all intermediate states are preserved; at level $0.5$, approximately half are randomly retained. We use values between 0.3 and 0.5 to balance realism with trace length. This breakdown transforms sparse LLM outputs (e.g., 10 major states) into dense progressions (e.g., 50+ snapshots) matching the temporal resolution of real student data. The heuristics mirror human typing: lines appear incrementally, function signatures precede bodies, and modifications proceed word-by-word.

\noindent\textbf{Evaluation Interface.}
Raters used a web-based animated viewer displaying code-only snapshots with syntax highlighting. The interface featured a timeline slider with play/pause controls and adjustable playback speed (0.5x--2x). Each code trace was shown as a progression of states, with raters controlling navigation to observe how code evolved step-by-step. Raters provided:
\begin{enumerate}[noitemsep, topsep=2pt]
    \item Two 5-point Likert ratings: \emph{Behavioral Realism} (coding patterns, mistakes, progression) and \emph{Code Realism} (logic, structure, style).
    \item A binary classification: \emph{Real Student} or \emph{AI Generated}.
    \item Optional free-text reasoning.
\end{enumerate}
After completing all samples, raters reported difficulty level (1--5), their CS background, and qualitative feedback on differentiation cues. No timestamps or monologue text were shown, matching the information available in the original Bielefeld dataset.

\subsection{Statistical Analysis}
\label{app:turing_stats}

We performed a comprehensive statistical analysis of the human Turing test ($N=852$ observations from 71 participants).

\noindent\textbf{Accuracy and Significance.}
Participants achieved an overall accuracy of 52.8\% (450/852). A binomial test against the chance probability of 50\% yielded $p=0.053$, indicating the result is \emph{not} statistically significant, as participants performed at near-chance levels. This alone suggests \beagle traces are difficult to distinguish from real student work.

\begin{figure}[t]
    \centering
    \includegraphics[width=0.35\linewidth]{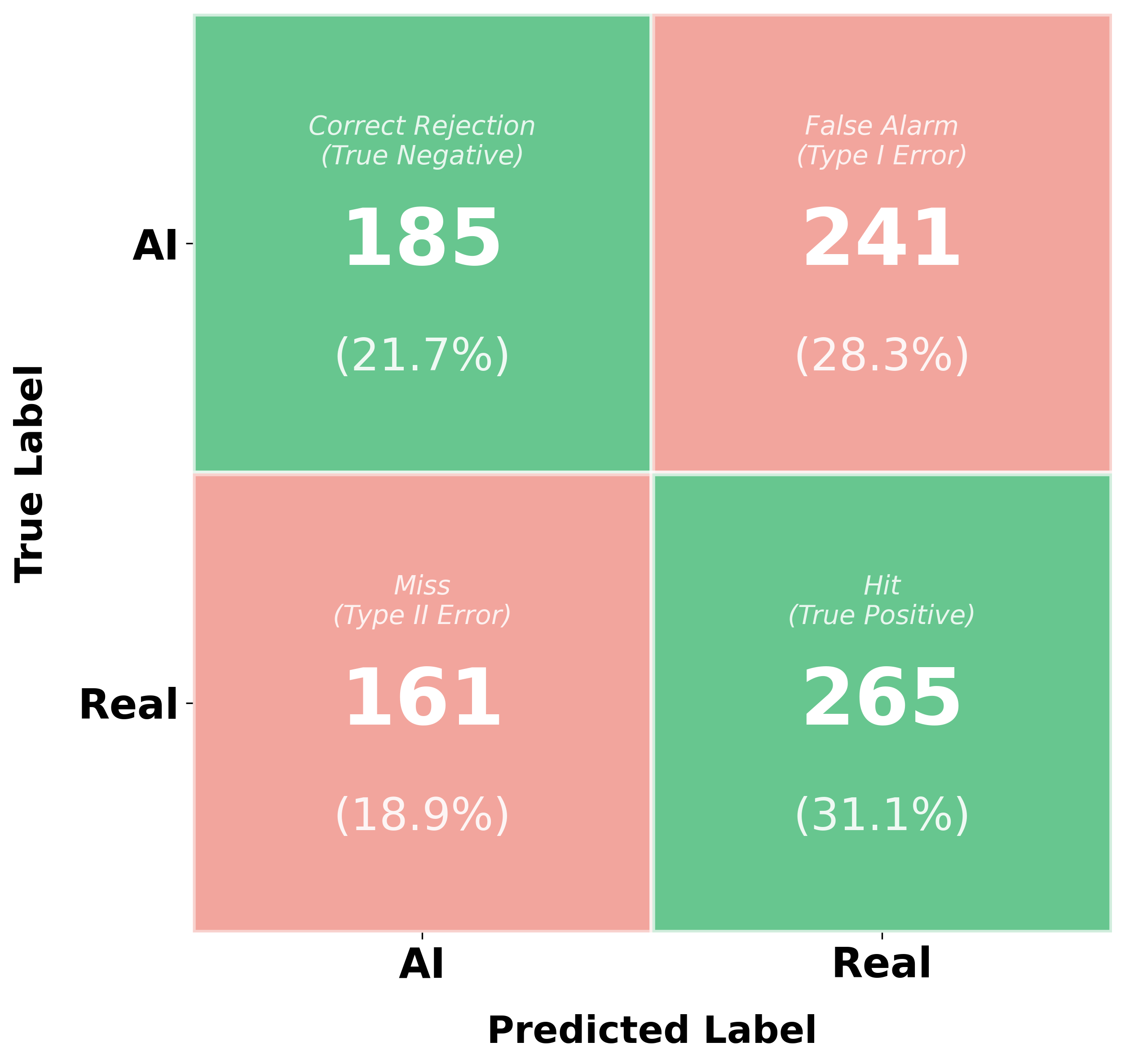}
    \caption{Confusion matrix showing response counts and percentages. The high False Alarm rate (28.3\%) indicates participants frequently misclassified AI traces as real.}
    \label{fig:confusion_matrix}
\end{figure}

\noindent\textbf{Signal Detection Theory (SDT).}
We decomposed performance using SDT metrics, as shown in the confusion matrix (Figure~\ref{fig:confusion_matrix}). The Hit Rate (Real correctly identified as Real) was 62.2\%, while the False Alarm Rate (AI incorrectly classified as Real) was 56.6\%, yielding a Specificity (AI Detection Rate) of only 43.4\%. The Criterion (bias) was $c = -0.24$, indicating a systematic bias toward classifying traces as ``Real.'' This explains why participants performed worse than chance on synthetic data (43.2\%) while achieving moderate accuracy on real data (62.6\%).

\noindent\textbf{Equivalence Testing.}
The discriminability index was $d' = 0.15$ ($SE=0.087$). To test for indistinguishability, we applied the Two One-Sided Tests (TOST) procedure with an equivalence margin of $\delta = \pm 0.3$ (negligible effect size).
\begin{itemize}[noitemsep]
    \item Test 1 ($d' > -0.3$): $z=5.13, p < 0.001$
    \item Test 2 ($d' < 0.3$): $z=-1.78, p = 0.038$
\end{itemize}
The maximum p-value ($p=0.038$) allows us to reject the null hypothesis of inequality, confirming that \beagle traces are perceptually equivalent to real student data within the defined margin.

\subsection{Qualitative Findings and Discussion}
\label{app:turing_discussion}

The Turing test results offer valuable insights into what makes simulated programming behavior convincing. We collected responses from 71 participants across diverse backgrounds: graduate students (36\%), undergraduates (32\%), professionals (11\%), hobbyists (9\%), beginners (7\%), and educators (5\%). Participants rated the task as difficult (4.16/5).

The qualitative feedback reveals a fascinating paradox: participants developed reasonable heuristics that nonetheless failed. Several noted that ``real students follow clear thinking patterns... while AI samples generate in no particular order.'' Yet this same participant achieved only 33\% accuracy. Comments emerged as the most contested signal: one participant stated ``the comments give it away,'' while another noted ``hashtags when commenting'' as an AI tell. In reality, both real students and \beagle produced similar commenting styles, leading expert raters to perform at chance levels.

The most insightful feedback came from participants who recognized the fundamental difficulty: ``Is this code bad because the person is bad at coding or because the AI isn't consistent?'' This captures the core challenge: \beagle produces code that is neither consistently ``too good'' nor ``too bad,'' but instead occupies the ambiguous middle ground where real student work also resides.

Several specific patterns did successfully distinguish AI from real: ``writing multi-line docstrings'' (real students rarely document during timed tasks), ``imported math before required'' (real students import on first use), and ``switching between methods back and forth'' in ways that felt mechanical rather than goal-directed. Conversely, cues that correctly identified real students included ``copy/pasting function lines then changing names'' and ``going back and catching the mistake.''

The Likert ratings showed near-parity between AI (behavior: 3.36, code: 3.29) and Real (behavior: 3.54, code: 3.49) samples. Perhaps most tellingly, graduate students and professionals performed no better than beginners, despite having the most experience reading student code. Expertise offered no advantage in this task. As one participant ruefully concluded: ``It's joever---I was no better than a coin flip.''

%% file: prompts.tex
%


\tcbset{
  commonhighlight/.style={
    enhanced,
    breakable,
    lines before break=2,
    boxrule=0pt,
    arc=2pt,
    left=5pt, right=5pt, top=4pt, bottom=4pt,
    before skip=5pt, after skip=5pt,
  },
}

\newtcolorbox{highlightgray}{commonhighlight, colback=gray!15, colframe=gray!15}
\newtcolorbox{highlightred}{commonhighlight, colback=red!12, colframe=red!12}
\newtcolorbox{highlightblue}{commonhighlight, colback=blue!10, colframe=blue!10}
\newtcolorbox{highlightgreen}{commonhighlight, colback=green!12, colframe=green!12}
\newtcolorbox{highlightorange}{commonhighlight, colback=orange!15, colframe=orange!15}
\newtcolorbox{highlightpurple}{commonhighlight, colback=purple!10, colframe=purple!10}
\newtcolorbox{highlightteal}{commonhighlight, colback=teal!12, colframe=teal!12}
\newtcolorbox{highlightyellow}{commonhighlight, colback=yellow!20, colframe=yellow!20}

\newtcolorbox{prompttitle}{
  enhanced,
  boxrule=0.5pt,
  arc=2pt,
  colback=gray!40,
  colframe=gray!60,
  left=5pt, right=5pt, top=3pt, bottom=3pt,
  before skip=10pt, after skip=3pt,
  fontupper=\bfseries\small,
}
\newpage
\section{Prompt Templates}
\label{app:prompts}

The prompt architecture implements composable blocks that assemble according to the composition formula from Section~\ref{sec:method}. This appendix documents the block hierarchy, universal rules, mandate variations, and representative examples.

\subsection{Block-Based Composition}

Prompts are assembled from reusable blocks. Both stages share an identical SYSTEM prompt but differ in USER prompts:

\begin{align}
    \text{System} &= \underbrace{\text{Static}(\rho_{\text{persona}})}_{\text{persona, rules}} \oplus \underbrace{\text{Mandate}(M_n, C_t)}_{\text{behavioral frame}} \\[1ex]
    \text{User}_{\text{Strat}} &= \underbrace{\text{Task}_{\text{Strat}}}_{\text{output format}} \oplus \underbrace{\text{Mem}(\Gamma_{\text{Strat}, n})}_{\text{memory}} \oplus \underbrace{\text{Context}(\Omega_t)}_{\text{runtime state}} \\[1ex]
    \text{User}_{\text{Exec}} &= \underbrace{\text{Profile}(M_n)}_{\text{empirical patterns}} \oplus \underbrace{\text{Strategy}(g_n, m_n, d_n)}_{\text{from Strategist}} \notag \\
        &\quad \oplus \underbrace{\text{Task}_{\text{Exec}}(C_t)}_{\text{output format}} \oplus \underbrace{\text{Mem}(\Gamma_{\text{Exec}, t})}_{\text{memory}} \oplus \underbrace{\text{Context}(\Omega_t)}_{\text{runtime state}}
\end{align}

\noindent This modular architecture treats the prompt as a dynamic assembly of semantic units rather than a static template, enabling precise control over the agent's behavior. Each block serves a distinct functional role in the simulation: Static blocks establish the persona's baseline voice independent of the task; Mandate blocks enforce the behavioral constraints dictated by the symbolic state $S_t$; and Task blocks structure the output for the specific cognitive activity. By composing these components at runtime, the system can generate a diverse range of student trajectories while maintaining strict adherence to the underlying semi-Markov dynamics. Table~\ref{tab:prompt-blocks} summarizes the specific role of each component.

\begin{table}[h!t]
\centering
\caption{Prompt Block Components}
\label{tab:prompt-blocks}
\small
\begin{tabular}{@{}lp{4.5cm}l@{}}
\toprule
\textbf{Block} & \textbf{Purpose} & \textbf{Ref} \\
\midrule
$\text{Base}$ & Core persona: novice voice, simple vocabulary & \S\ref{app:base} \\
$\text{Performer}(\rho_{\text{persona}})$ & Style modifier (low vs high) & \S\ref{app:performer} \\
$\text{Rules}$ & Universal constraints (3 rules) & \S\ref{app:rules} \\
$\text{Mandate}(M_n, C_t)$ & State-pair behavioral framing (12 combinations) & \S\ref{app:mandate} \\
$\text{Profile}(M_n)$ & Empirical language patterns per metacognitive behavior & \S\ref{app:profile} \\
$\text{Task}$ & Output format: $\text{Task}_{\text{Strat}}$ (constant), $\text{Task}_{\text{Exec}}(C_t)$ (3 variants) & \S\ref{app:task} \\
$\text{Strategy}(g_n, m_n, d_n)$ & Strategist output passed to Executor & --- \\
$\text{Mem}(\Gamma_t)$ & Injection of stage-specific memory buffers & \S\ref{app:context} \\
$\text{Context}(\Omega_t)$ & Runtime state injection (code, feedback, KCs) & \S\ref{app:context} \\
\bottomrule
\end{tabular}
\end{table}

\noindent The $\text{Static}(\rho_{\text{persona}})$ block expands as:
\begin{equation}
\text{Static}(\rho_{\text{persona}}) = \underbrace{\text{Base}}_{\text{voice}} \oplus \underbrace{\text{Performer}(\rho_{\text{persona}})}_{\text{style}} \oplus \underbrace{\text{Rules}}_{\text{constraints}}\label{eq:static}
\end{equation}

\subsection{Static Components}

The $\text{Static}(\rho_{\text{persona}})$ block provides foundational persona and constraints:
\begin{equation}
\text{Static}(\rho_{\text{persona}}) = \underbrace{\text{Base}}_{\text{voice}} \oplus \underbrace{\text{Performer}(\rho_{\text{persona}})}_{\text{style}} \oplus \underbrace{\text{Rules}}_{\text{constraints}}
\end{equation}

\subsubsection{\texorpdfstring{$\text{Base}$}{Base}}
\label{app:base}

\begin{highlightgray}
You are a novice student learning to code in Python.

\textbf{Voice}: Use simple, direct language. Think out loud naturally.

\textbf{Vocabulary}:
\begin{itemize}[nosep,leftmargin=*]
  \item ``I think...'' not ``I hypothesize...''
  \item ``It says...'' not ``The error indicates...''
  \item ``That's weird'' not ``This is unexpected behavior''
\end{itemize}

\textbf{Authenticity}: You may be confused. You may not know the right answer. You learn by doing, not by analyzing.
\end{highlightgray}

\subsubsection{\texorpdfstring{$\text{Performer}(\rho_{\text{persona}})$}{Performer(rho)}}
\label{app:performer}

\begin{highlightpurple}
\textbf{Low Performer} ($\rho_{\text{persona}} = \text{low}$)

\textbf{Language}: ``I don't know...'', ``Maybe if I...'', ``Let's just try...'', ``Why isn't this working?''

\textbf{Behavior}: Prefer action over analysis. Often repeat the same action if it fails. May give up: ``I don't understand, let me just try something.''
\end{highlightpurple}

\begin{highlightpurple}
\textbf{High Performer} ($\rho_{\text{persona}} = \text{high}$)

\textbf{Language}: ``Let me check...'', ``I think the issue is...'', ``Based on what I see...''

\textbf{Behavior}: Pattern-matching heuristics. Evidence-based reasoning. Still makes novice mistakes but tries to understand errors.
\end{highlightpurple}

\subsubsection{\texorpdfstring{$\text{Rules}$}{Rules}}
\label{app:rules}

\begin{highlightred}
\textbf{No Psychic Debugging}

If Last Output shows ``(Code drafted but not executed)'', you have NOT run the code.

\textbf{You Cannot}: Identify bugs or issues; Say ``I forgot...'' or ``I need to fix...''; Notice missing variables or wrong syntax.

\textbf{You Can Only}: Think about WHAT you're building next; Express uncertainty about the approach.
\end{highlightred}

\begin{highlightred}
\textbf{Memory Rule (Anti-Amnesia)}

Before generating output, check if this is a REPEATED mistake.

\textbf{If Yes}: Say ``Oh right, I made this mistake before'' NOT ``Hmm, what's wrong here?''

Show that you REMEMBER previous errors and learn from them.
\end{highlightred}

\begin{highlightred}
\textbf{Grounding Rule}

Your monologue MUST be grounded in actual execution output.
\begin{itemize}[nosep,leftmargin=*]
  \item Reference specific error types you see (e.g., ``NameError'', ``TypeError'')
  \item Quote actual values from output
  \item Do NOT invent errors or outputs you don't see
\end{itemize}
\end{highlightred}

\subsection{\texorpdfstring{$\text{Mandate}(M_n, C_t)$}{Mandate(M,C)}}
\label{app:mandate}

The $\text{Mandate}(M_n, C_t)$ block provides state-pair-specific behavioral framing. Table~\ref{tab:mandate-matrix} summarizes the 12 combinations, with empirical language patterns derived from talk-aloud transcripts \cite{snyder2024analyzing}.

\begin{table}[t]
\centering
\caption{Mandate Matrix: Behavioral Framing for Each $(M_n, C_t)$ Pair}
\label{tab:mandate-matrix}
\footnotesize
\setlength{\tabcolsep}{4pt}
\begin{tabular}{@{}l p{3.4cm} p{3.4cm} p{3.4cm}@{}}
\toprule
& \textbf{\constructing} & \textbf{\debugging} & \textbf{\assessing} \\
\midrule
\textbf{\planning} &
\textsc{Goal Formulation}: Forward looking, epistemically blind. \textit{``We need to...''} &
\textsc{Diagnostic Planning}: Hypothesis formation, analyzing failure. \textit{``So that's why...''} &
\textsc{Evaluative Planning}: Hopeful, checking completion. \textit{``Let's check if...''} \\
\midrule
\textbf{\enacting} &
\textsc{Trial Action}: Immediate action, learning by doing. \textit{``Let's try this''} &
\textsc{Reactive Fixing}: Surface-level fixes, anti-efficiency. \textit{``Maybe 3.7''} &
\textsc{Quick Check}: Shallow evaluation, moving fast. \textit{``Looks good''} \\
\midrule
\textbf{\monitoring} &
\textsc{Progress Tracking}: Observing values, still blind to bugs. \textit{``It's not at 15 yet''} &
\textsc{Error Tracking}: Tracking progress, frustration. \textit{``It's not stopping''} &
\textsc{Result Verification}: Checking final output. \textit{``Did it pass?''} \\
\midrule
\textbf{\reflecting} &
\textsc{Learning Reflection}: Meta-observation, seeking understanding. \textit{``That's weird''} &
\textsc{Error Understanding}: Causal reasoning, may include frustration. \textit{``I don't know what we did different''} &
\textsc{Outcome Reflection}: Learning consolidation. \textit{``So the key was...''} \\
\bottomrule
\end{tabular}
\end{table}

\noindent Key architectural notes:
\begin{itemize}[nosep]
    \item \textbf{Epistemic Filter}: Active in \enacting{} states, limiting analytical depth.
    \item \textbf{BKT Update}: Triggered in \monitoring{} and \reflecting{} states (learning during conscious evaluation).
    \item \textbf{Trap State}: \enacting$\times$\debugging{} consumes the most steps for low performers (30\% vs 18\%).
\end{itemize}

\subsection{\texorpdfstring{$\text{Profile}(M_n)$}{Profile(M)}}
\label{app:profile}
The $\text{Profile}(M_n)$ block injects empirical language patterns for each metacognitive behavior, derived from talk-aloud transcripts~\cite{snyder2024analyzing}. These profiles shape the Executor's linguistic style.

\begin{highlightblue}
\textbf{\planning}: The Strategist (Thinking Before Acting)

\textbf{Verbalize Goals}: State objectives before acting. ``I need to figure out where the truck should stop'' not ``let me try something.''

\textbf{Deliberate}: Consider which approach or equation to use. Ask questions: ``Should I use this formula?''

\textbf{Uncertainty}: Express uncertainty freely: ``I'm not sure how to do that.''
\end{highlightblue}

\begin{highlightblue}
\textbf{\enacting}: The Actor (Hands on Keyboard)

\textbf{Verbalize Intent}: State micro-goals before acting. ``I am setting velocity to 15.''

\textbf{Reactive Execution}: Just do it, don't command others. ``Let's try this'' then DO IT.

\textbf{Trial-and-Error}: Quick iterations, not deep analysis. If it fails, try again.
\end{highlightblue}

\begin{highlightblue}
\textbf{\monitoring}: The Spotter (Watching the Screen)

\textbf{Commanding Tone}: Use imperatives to check values. ``Check the X position,'' ``Look at the timer.''

\textbf{Data Obsessed}: Quote specific numbers. ``The value is 15.2, but we expected 15.0'' not ``It looks wrong.''

\textbf{Track Progress}: Note what changed. ``The error is gone now'' or ``It's still not working.''
\end{highlightblue}

\begin{highlightblue}
\textbf{\reflecting}: The Critic (Making Sense of What Happened)

\textbf{Seek Understanding}: Think about WHY. ``I think the reason is...'' or ``This makes sense because...''

\textbf{Emotional Range}: Express satisfaction (``I get it now!'') or frustration (``This is confusing.'').

\textbf{Look Backward}: Reference prior attempts. ``Last time I did X and it didn't work.''
\end{highlightblue}


\subsection{\texorpdfstring{$\text{Task}$}{Task} Blocks}
\label{app:task}

The Task blocks specify the output format and immediate objective for the agent.

\subsubsection{\texorpdfstring{$\text{Task}_{\text{Strat}}$}{Task-Strat}}

The Strategist produces a structured strategy packet $(g_n, m_n, d_n)$. The prompt enforces a separation between broad goals and immediate directives:

\begin{highlightgreen}
\textbf{Strategist Output Format}

Generate three things that guide the Executor:
\begin{enumerate}[nosep]
    \item \textbf{Goal}: A high-level objective (durable across multiple steps)
    \item \textbf{Mindset}: Your emotional/cognitive state
    \item \textbf{Directive}: A specific instruction for the Executor
\end{enumerate}
\end{highlightgreen}

\subsubsection{\texorpdfstring{$\text{Task}_{\text{Exec}}(C_t)$}{Task-Exec(C)}}

The Executor's task definition depends on the cognitive state $C_t$. This block translates the abstract behavior into concrete instructions for the LLM.

\paragraph{Constructing.}
The agent is tasked with building new code artifacts. The instruction frames the task as generative, focusing on translating plans into executable code without immediate execution.

\begin{highlightgreen}
\textbf{Task}: Drafting New Code

\textbf{Output}: Monologue (1-2 sentences) + Code

\textbf{Constraint}: Code has NOT been executed yet. You are building, not fixing. Do not hallucinate runtime errors.
\end{highlightgreen}

\paragraph{Debugging.}
The agent is tasked with identifying and fixing errors. The instruction frames the task as corrective. Crucially, this block is only used when execution output is available, grounding the debugging in reality.

\begin{highlightgreen}
\textbf{Task}: Fixing Based on Output

\textbf{Output}: Monologue (confused reaction) + Code (attempted fix)

\textbf{Constraint}: You MUST reference the actual error type found in the Context. React to what you see, not what you assume.
\end{highlightgreen}

\paragraph{Assessing.}
The agent is tasked with evaluating correctness without necessarily modifying code. This task type appears during \monitoring\ (verifying progress) or \reflecting\ (reviewing work).

\begin{highlightgreen}
\textbf{Task}: Observing Results

\textbf{Output}: Reflection (1-2 sentences) + Code (optional, minor cleanup only)

\textbf{Constraint}: You are OBSERVING. Reflection MUST be grounded in the actual output provided.
\end{highlightgreen}

\paragraph{Task-State Interactions.}
The interaction between the \emph{Cognitive Task} (what to do) and the \emph{Metacognitive Mandate} (how to think) produces 12 distinct behavioral configurations. For example:
\begin{itemize}[nosep]
    \item \planning\ + \constructing: High-level design of new code structure.
    \item \enacting\ + \debugging: Rapid, intuition-driven fix attempts (``Just try changing X'').
    \item \monitoring\ + \assessing: Systematic verification of output correctness.
    \item \reflecting\ + \debugging: Post-hoc analysis of why a bug occurred.
\end{itemize}
These combinations emerge naturally from the block architecture without requiring hand-crafted templates for every pair.

\subsection{\texorpdfstring{$\text{Context}$ and $\text{Mem}$}{Context and Mem}}\label{app:context}

These blocks inject the runtime state and agent memory:

\begin{highlightteal}
\textbf{Runtime Context} ($\Omega_t$)
\begin{itemize}[nosep,leftmargin=*]
  \item \textbf{Code History} ($c_{t-1}$): The current code artifact.
  \item \textbf{Feedback} ($\tilde{o}_{t-1}$): Execution output (filtered by epistemic state).
  \item \textbf{Knowledge} ($\kappa_t$): Natural language constraints (e.g., ``You correctly applied: function def return, init method, force acceleration, ...'').
  \item \textbf{Intervention} ($\mathcal{I}_t$): Teacher feedback (if any).
\end{itemize}

\textbf{Memory} ($\Gamma_t$)
\begin{itemize}[nosep,leftmargin=*]
  \item \textbf{Thought Buffer}: The agent's internal scratchpad to avoid repetition.
  \item \textbf{Episodic Memory}: Summaries of past attempts ($\Gamma_{\text{Strat}}$ or $\Gamma_{\text{Exec}}$).
\end{itemize}
\end{highlightteal}

\section{Calibrated LLM-as-a-Judge Evaluation Prompt}
\label{app:judge_prompt}

To evaluate the realism of generated trajectories (Study 8), we employ a hybrid \emph{Forensic-Qualitative} prompting strategy. This prompt structure was finalized following a human calibration study where the LLM judge achieved substantial agreement with expert consensus ($\kappa_w=0.760$ for realism; $\kappa_w=0.674$ aggregate) on a stratified sample of 30 traces.

To prevent the LLM judge from hallucinating statistical features, we first execute a Python analysis pass to generate a \textbf{Forensic Fact Sheet}. This ground-truth injection is critical because simulation traces often span dozens of interactions; without explicit summarization, LLMs struggle to accurately track distributed patterns (such as exact repetition counts or code style consistency) across long context windows.

\subsection{Forensic Fact Sheet Injection}
The following block is dynamically generated by the Python evaluator and injected at the top of the prompt to ground the LLM's assessment.

\begin{highlightteal}
\textbf{\textsc{System Analysis (Ground Truth)}}

These metrics were computed by Python. Do not contradict these values.

\textbf{Language Metrics}
\begin{itemize}[nosep,leftmargin=*]
    \item Total steps: [Integer]
    \item Uncertainty markers (e.g., ``I don't know'', ``maybe''): [Count]
    \item Frustration markers (e.g., ``ugh'', ``confused''): [Count]
    \item Max single phrase repetition: [Count]
\end{itemize}

\textbf{Code Style Metrics}
\begin{itemize}[nosep,leftmargin=*]
    \item Cramped style ratio: [Percentage 0--100\%] (percentage of assignments like \texttt{x=y})
    \item Single-letter variables: [Count]
    \item Emotional code comments: [Count]
\end{itemize}

\textbf{Debugging Pattern Metrics}
\begin{itemize}[nosep,leftmargin=*]
    \item Disconnected fixes: [Count] (instances where the student mentions an error type different from the execution output).
\end{itemize}
\end{highlightteal}

\subsection{Judge Instructions}
The main instruction set defines the criteria for simulated versus authentic behavior.

\begin{highlightgray}
You are an expert in Computer Science Education acting as a judge. Your task is to determine if a student trace is \textbf{REAL} (human) or \textbf{SIMULATED} (AI).

\textbf{\textsc{Indicators of Realism}}
Real novice students share these characteristics:
\begin{enumerate}[nosep,leftmargin=*]
    \item \textbf{Reactive Debugging}: They discover bugs by running code, not by inspecting it beforehand.
    \item \textbf{Imperfect Code Style}: Cramped spacing (\texttt{x=y}), single-letter variables, and inline comments (``\# hope this works'').
    \item \textbf{Emotional Authenticity}: Expressions of frustration (``Ugh''), relief (``Finally!''), or uncertainty.
    \item \textbf{Non-Linearity}: A messy workflow (Constructing $\to$ Debugging $\to$ Constructing) rather than a clean linear path.
\end{enumerate}

\textbf{\textsc{AI Simulation Tells (Flag as Fake)}}
\begin{enumerate}[nosep,leftmargin=*]
    \item \textbf{Psychic Debugging}: Identifying runtime errors (e.g., ``I need to fix the TypeError'') \emph{before} running the code.
    \item \textbf{Perfect Code Style}: PEP-8 compliance, descriptive variable names, or proper docstrings.
    \item \textbf{Robotic Explanations}: Overly precise language (e.g., ``The function signature requires...'').
    \item \textbf{Amnesia}: Repeating the exact same mistake 5+ times without variation.
\end{enumerate}

\textbf{\textsc{Scoring Rubric}}
\begin{itemize}[nosep,leftmargin=*]
    \item \textbf{3 (Realistic)}: Authentic novice behavior. Discovers errors through execution; shows emotional reactions.
    \item \textbf{2 (Ambiguous)}: Mixed signals. Some authentic elements, some suspicious ones.
    \item \textbf{1 (Simulated)}: AI tells present (psychic debugging, perfect style, robotic tone).
\end{itemize}

\textbf{\textsc{Output Format}}
Respond with a JSON object containing:
\begin{itemize}[nosep,leftmargin=*]
    \item \texttt{justification}: String (Explanation of assessment)
    \item \texttt{realism\_score}: Integer (1--3)
    \item \texttt{code\_quality\_realism}: Integer (1--3)
    \item \texttt{debugging\_pattern\_realism}: Integer (1--3)
    \item \texttt{language\_realism}: Integer (1--3)
\end{itemize}
\end{highlightgray}